%% file: main.tex
\documentclass[letterpaper, 10 pt, conference]{ieeeconf}  

\IEEEoverridecommandlockouts                              

\usepackage{graphicx}
\usepackage{lineno,hyperref}

\modulolinenumbers[5]
\usepackage{graphics} 
\usepackage{graphicx}
\usepackage{caption}
\usepackage{subcaption}
\usepackage[ruled,vlined,linesnumbered,noresetcount]{algorithm2e}
\usepackage[]{units}
\usepackage{pgfplots}
\usepackage{tikz}
\usetikzlibrary{positioning}
\usetikzlibrary{shapes.geometric}
\usetikzlibrary{shapes.misc}
\usetikzlibrary{external}
\usepackage{paralist}
\usepackage{mathtools}

\usepackage{xcolor}
\usepackage{siunitx}
\usepackage{multirow}
\usepackage{epstopdf}
\usepackage{caption}
\usepackage{epstopdf}
\usepackage{tikz}
\usepackage{pgfplots}
\usetikzlibrary{shapes,arrows,patterns}
\usetikzlibrary{arrows.meta}
\usetikzlibrary{plotmarks}
\usetikzlibrary{calc,decorations.markings,math}
\usetikzlibrary{external}
\pgfplotsset{compat=newest}
\usepackage{grffile}
\usepackage{bm}
  \usepgfplotslibrary{patchplots}
\usepackage{epsfig}
\usepackage{float}
\usepackage{caption}
\usepackage{amsmath} 
\usepackage{amssymb}  
\usepackage{mathtools}
\graphicspath{ {Pictures/} }
\DeclareGraphicsExtensions{.pdf,.eps}
\usepackage[acronym,nomain,nonumberlist]{glossaries}
\input{abbreviations}
\usepackage{url}
\usepackage{threeparttable}
\newlength\fwidth

\title{\LARGE \bf
Geometry Aware NMPC Scheme for Morphing Quadrotor Navigation in Restricted Entrances
\thanks{This work has been partially funded by the European Unions Horizon 2020 Research and Innovation Programme under the Grant Agreement No. 869379 illuMINEation. Corresponding author's e-mail: \textit{andpap@ltu.se} } }

\author{Andreas Papadimitriou, Sina Sharif Mansouri, Christoforos Kanellakis and George Nikolakopoulos
\thanks{The authors are with the Robotics \& AI Team, Department of Computer, Electrical and Space Engineering, Lule\r{a} University of Technology, Lule\r{a} SE-97187, Sweden}
}

\begin{document}

\maketitle
\thispagestyle{empty}
\pagestyle{empty}

\begin{abstract}

Geometry-morphing \glspl{mav} are gaining more and more attention lately, since their ability to modify their geometric morphology in-flight increases their versatility, while expanding their application range. In this novel research field, most of the works focus on the platform design and on the low-level control part for maintaining stability after the deformation. Nevertheless, another aspect of geometry morphing \glspl{mav} is the association of the deformation with respect to the shape and structure of the environment. In this article, we propose a novel \gls{nmpc} structure that modifies the morphology of a quadrotor based on the environmental entrances geometrical shape. The proposed method considers restricted entrances as a constraint in the \gls{nmpc} and modifies the arm configuration of the \gls{mav} to provide a collision free path from the initial position to the desired goal, while passing through the entrance. To the authors' best knowledge, this work is the first to connect the in-flight morphology with the characteristics of environmental shapes. Multiple simulation results depict the performance and efficiency of the proposed scheme in scenarios where the quadrotor is commanded to pass through restricted areas.


 
\end{abstract}
\glsresetall

\section{INTRODUCTION}
\glspl{mav} are agile platforms with a variety of navigation capabilities including hovering over a target or aggressive maneuverability, allowing them to access remote and distance places, navigate through complex and harsh environments~\cite{mansouri2020deploying}. They are currently penetrating into the global economy with fast pace, invading markets from infrastructure to public security~\cite{mansouri2018cooperative,adaldo2017cooperative}, entertainment, search and rescue~\cite{sampedro2018fully}, mining~\cite{mansouri2020nmpcmine} and forest industry~\cite{steich2016tree}. A side effect of the vast interest to invest in \glspl{mav} is to constantly create new needs and market requirements, which eventually push the robotics community to look towards unexplored aspects of the technology. Currently, conventional \glspl{mav} with fixed frames lack the ability to adapt their shape based on the task requirements and environmental structure. For example, the common approach for a conventional platform, in cases where the path traverses through narrow gaps, would have been to miniaturize its size, sacrificing onboard payload. To this end, the concept of geometry-morphing aerial platforms is becoming relevant, trying to address the limitation of flight time and payload in cases where larger platforms with the desired payload can adaptively restructure their morphology in-flight to pass through restricted entrances.

Towards this direction, this article proposes a novel \gls{nmpc} framework for collision-free navigation of a morphing \gls{mav}, through narrow entrances. The platform alters the position of the arms, based on the shape of the entrance, formulated as constraints in the control scheme. Another aspect of the proposed approach is that in case where the size of the entrance is below the allowed feasible, relative to the size of the platform, it does not allow the MAV to pass through that opening. Finally, the framework is able to provide different configurations, while maximizing the symmetry between arms when reforming for passing through the holes, a concept which assist the low-level controller to compensate for the arm configurations~\cite{falanga2018foldable}. 

\subsection{Background \& Motivation}
The majority of the works in the literature of geometry-morphing aerial platforms focus on the design of the \gls{mav} and the low-level control scheme to maintain it's stability when shape reformation occurs during the flight. The development of a foldable quadrotor with the ability to maintain stable flight, after changing its formation by rotating motion of each arm individually, has been presented in~\cite{falanga2018foldable}, while in~\cite{fabris2020geometry} the authors study the effects of the overlapping among the propellers and the body frame. A foldable quadrotor design was presented in~\cite{riviere2018agile}, where the platform was able to decrease its wide-span, by changing the orientation of its propellers based on an actuated elastic mechanism. Furthermore, in~\cite{tuna2020design} a self-foldable quadrotor has been presented with a gear-based mechanism to control the contraction and expansion of the four arms simultaneously. This approach allowed for two possible configurations namely either fully expanded when the drone is deployed or fully contracted when the drone is on the ground. A passive foldable quadrotor has been presented in~\cite{bucki2019design} that utilizes springs for altering its formation. The maneuverability of this design, while the drone is in its reduced form, was limited and the quadrotor can traverse only for a short time through narrow gaps. A sliding arm quadrotor has been presented in~\cite{kumar2020flight} from a modeling and control point of view, while few other works investigate \glspl{mav} with morphing~\cite{bai2019evaluation,derrouaoui2020design,sakaguchi2019novel,papadimitriou2020switching,meiri2019flying,shu2019quadrotor}. Besides transformable quadrotor platforms, other novel aerial vehicles that can alter their structure have been presented in the last years. \textit{DRAGON} is a dual rotor multilink aerial robot that alters its formation with the use of multiple servos while flying~\cite{zhao2018design} and being able to traverse through gaps. 
\newline
Based on the related literature, there is a lack of scientific works that address the coupling of the morphing with the geometrical structure of the environment. 

\subsection{Contributions}
Based on the aforementioned state of the art, the main contribution of this article is threefold. The first contribution stems from incorporating the kinematic model of the \gls{mav}'s arm configuration in a novel \gls{nmpc} framework. This formulation enables to reconfigure the in-flight geometric shape of the \gls{mav}, resulting in different arm configurations, in contrary to the related works which consider mainly four discrete types of configurations (X,Y,O,H).


The second contribution stems from the integration of the geometric constraints, which link the morphology of the platform and the restricted entrance in the \gls{nmpc} formulation. The limited entrances are considered as nonlinear constraints, while based on the arm kinematics, the \gls{nmpc} provides arm configuration to pass through the entrances. The overall framework provides a collision free path by reconfiguring the \gls{mav} arms, while considering \gls{mav} nonlinear dynamics and the geometric shape for a priori known restricted entrance, and only the desired set point is provided for the \gls{nmpc} framework.

Finally, the efficiency of the proposed method is evaluated on multiple simulations, while it is demonstrated that without changing the arm configuration, the collision is unavoidable. In all cases the suggested novel framework is able to provide a collision free path.


\subsection{Outline}
The rest of this article is structured as follows. In Section \ref{sec:Problem_Statement} the research problem is defined, while highlighting the challenges and limitations. Section \ref{sec:NMPC} provides the non-linear model of a \gls{mav}, the kinematics of the folding mechanism, the theoretical control framework and the formulation of the optimization problem. The simulation specifics and results are provided in Section \ref{sec:Results}, while concluding remarks and future work are discussed in \ref{sec:Conclusions}.
\section{Problem Statement and Open Challenges}\label{sec:Problem_Statement}
\begin{figure}[htb]
\begin{center}
\includegraphics[width=\columnwidth]{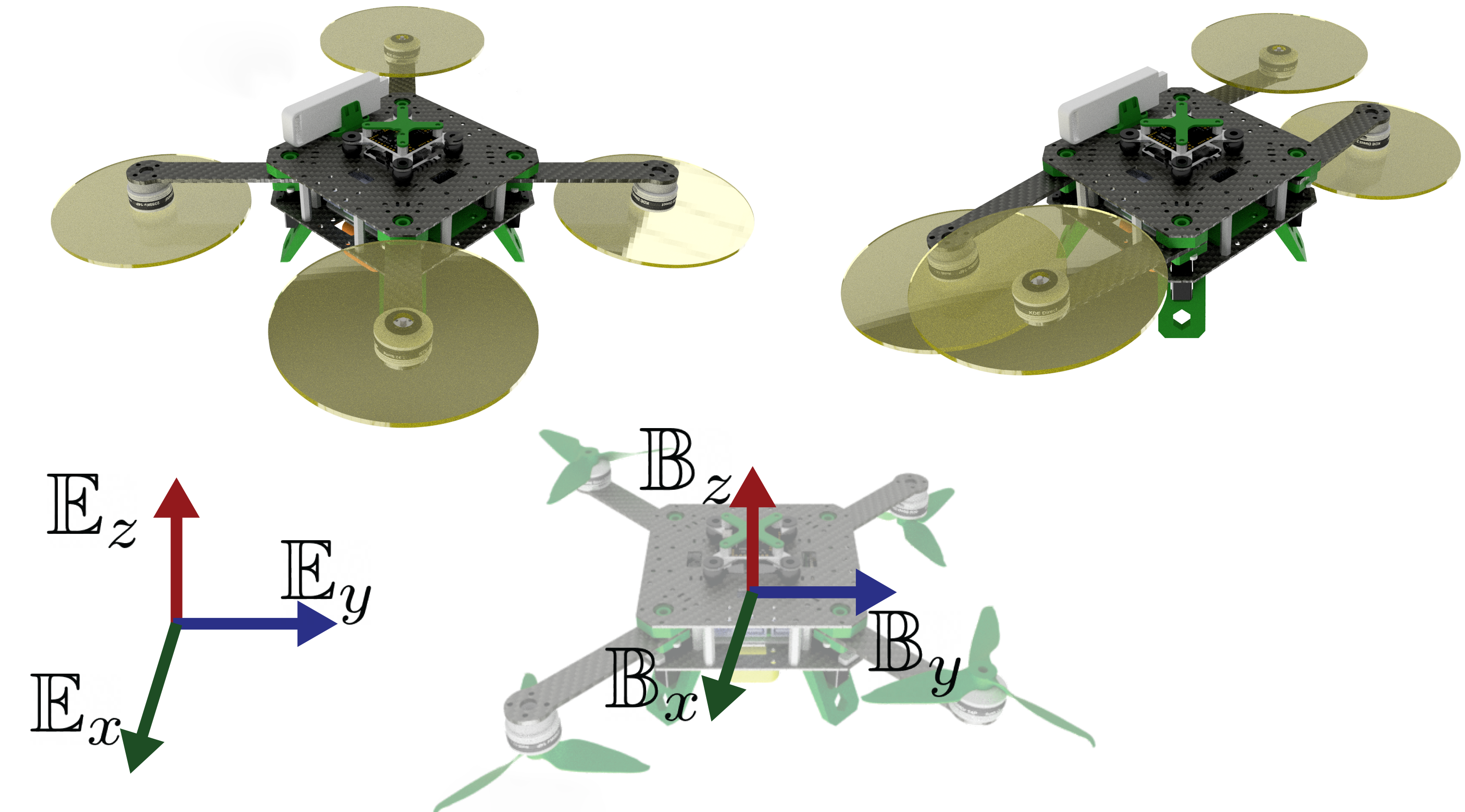}
\caption{Isometric view of the reconfigurable \gls{mav} conceptual design in X and H morphology and a side view with the attached body fixed frame $\mathbb{B}$ and inertial frame $\mathbb{E}$.}
\label{fig:conceptual}
\end{center}
\end{figure}

In Fig.~\ref{fig:conceptual} the conceptual design of a reconfigurable \gls{mav} is illustrated in isometric view for different formations. The arms of the quadrotor are connected on servos and they are able to rotate around the $z$-axis. The geometry varies only related to the $x$ and $y$-axis resulting in a planar-varying geometry as it is described in the sequel in Section \ref{sec:NMPC}. To overcome possible collisions between the propellers at the extreme angles, the motors have been placed alternately upside down (Fig.~\ref{fig:conceptual}). 

\begin{figure}[htb]
\begin{center}
\includegraphics[width=\columnwidth]{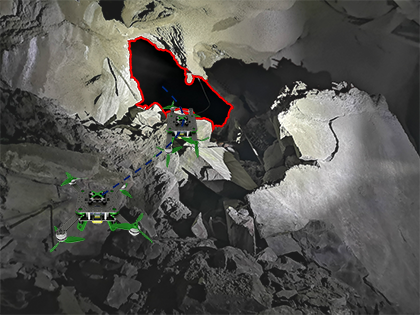}
\caption{Concept image of the proposed method in the narrow entrance in the underground mine. The proposed \gls{nmpc} provides arm configuration to pass through the entrance.}
\label{fig:concept}
\end{center}
\end{figure}

The \glspl{mav} can be considered as a scouting platform to gather information from challenging environments, such as underground mines~\cite{mansouri2020deploying} or search and rescue mission after natural catastrophes such as earthquakes. In most of these applications, the \gls{mav} should pass through narrow passages, as depicted in Figure~\ref{fig:concept} from underground mines. This article focus on the development of the \gls{nmpc} framework that optimizes the size of a quadrotor by adapting its configuration based on the size of constrained entrances. The entrances' location and size assumed to be known, thus the controller of the \gls{mav} provides the thrust, roll, pitch commands, and the arm configuration to navigate through the hole.


The ability of a reconfigurable quadrotor to alter its structure to different morphology can enhance specific characteristics, such us the size reduction and the sensor exposure of the platform at a cost of some other characteristics, such as nominal flight time and maneuverability. Out of all the different formations the specific design can take, the X morphology is the most balanced one thus is considered being the nominal configuration~\cite{falanga2018foldable}.

\begin{figure*}[htbp!]
\begin{center}
\includegraphics[width=0.8\textwidth]{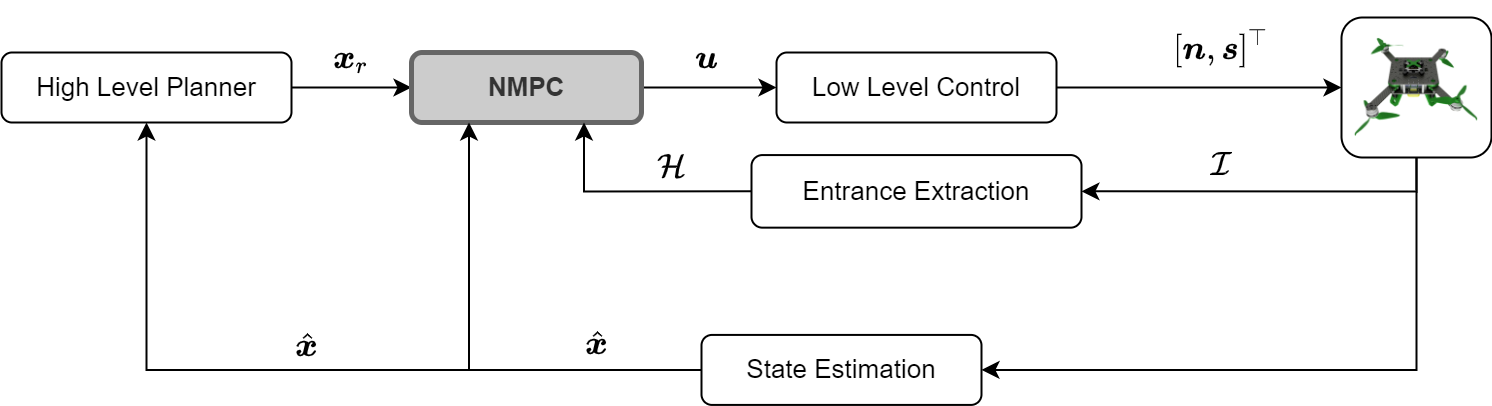}
\caption{The proposed control architecture for the reconfigurable quadrotor, while the proposed \gls{nmpc} module is depicted with a gray block.}
\label{fig:Controllerscheme}
\end{center}
\end{figure*}

Figure~\ref{fig:Controllerscheme} depicts the block diagram of the proposed structure, while the proposed framework of the \gls{nmpc} is depicted with the gray highlighted block. The \gls{nmpc} framework, based on the estimated states $\hat{\bm{x}}$, the entrance information $\mathcal{H}$, and the reference waypoint $\bm{x}_r$ provides thrust, roll, pitch and angles of the arms $\bm{u}$ for the low level controller. The low level controller generates motor and servo commands for the \gls{mav} $\bm{n}$ and $\bm{s}$ respectively to track the desired attitude and morphology. This work focuses on the \gls{nmpc} formulation, assuming that other state of the art entrance extraction methods (e.g.~\cite{sanket2018gapflyt}) will provide $\mathcal{H}$.


\section{Nonlinear Model Predictive Control}\label{sec:NMPC}

\subsection{MAV Dynamics }
 The states of the nonlinear system of the six \gls{dof} \gls{mav} based on~\cite{lindqvist2020non} are $\bm{x}=[p_x, p_y, p_z, v_x, v_y, v_z, \phi,\theta]^\top$ and modelled by \eqref{eq:modeluav}: 
\begin{subequations} \label{eq:modeluav}
\begin{align}
        \dot{\bm{p}}(t) &= \bm{v}(t) \\ 
        \dot{\bm{v}}(t) &= \bm{R}_{x,y}(\theta,\phi) 
        \begin{bmatrix} 0 \\ 0 \\ T \end{bmatrix} + 
        \begin{bmatrix} 0 \\ 0 \\ -g \end{bmatrix} - 
        \begin{bmatrix} A_x & 0 & 0 \\ 0 &  A_y & 0 \\ 0 & 0 & A_z \end{bmatrix} \bm{v}(t),   \\
        \dot{\phi}(t) & = \nicefrac{1}{\tau_\phi} (K_\phi\phi_d(t)-\phi(t)),  \\
        \dot{\theta}(t) & = \nicefrac{1}{\tau_\theta} (K_\theta\theta_d(t)-\theta(t)),
\end{align}
\end{subequations}
where $\bm{p}=[p_x,p_y,p_z]^\top \in \mathbb{R}^3$ is the position vector and $\bm{v} = [v_x, v_y, v_z]^\top \in \mathbb{R}^3$ is the vector of linear velocities, 
$\phi,\theta \in \mathbb{R} \cap [-\pi,\pi]$ are the roll and pitch angles respectively, $\bm{R}_{x,y}$ is the rotation matrix about 
the $x$ and $y$ axis, $T \in [0,1] \cap \mathbb{R}$ is the mass-normalized thrust, $g$ is the gravitational acceleration, 
$A_x, A_y$ and $A_z \in \mathbb{R}$ are the normalized mass drag coefficients. The low-level control system is approximated by the first-order dynamics, driven by the reference pitch and roll angles $\phi_d$ and $\theta_d$ with gains of $K_\phi, K_\theta \in \mathbb{R}^+$ and time constants of $\tau_\phi \in \mathbb{R}^+$ and $\tau_\theta \in \mathbb{R}^+$.

\begin{figure}[htbp!]
\begin{center}
\includegraphics[width=0.8\columnwidth]{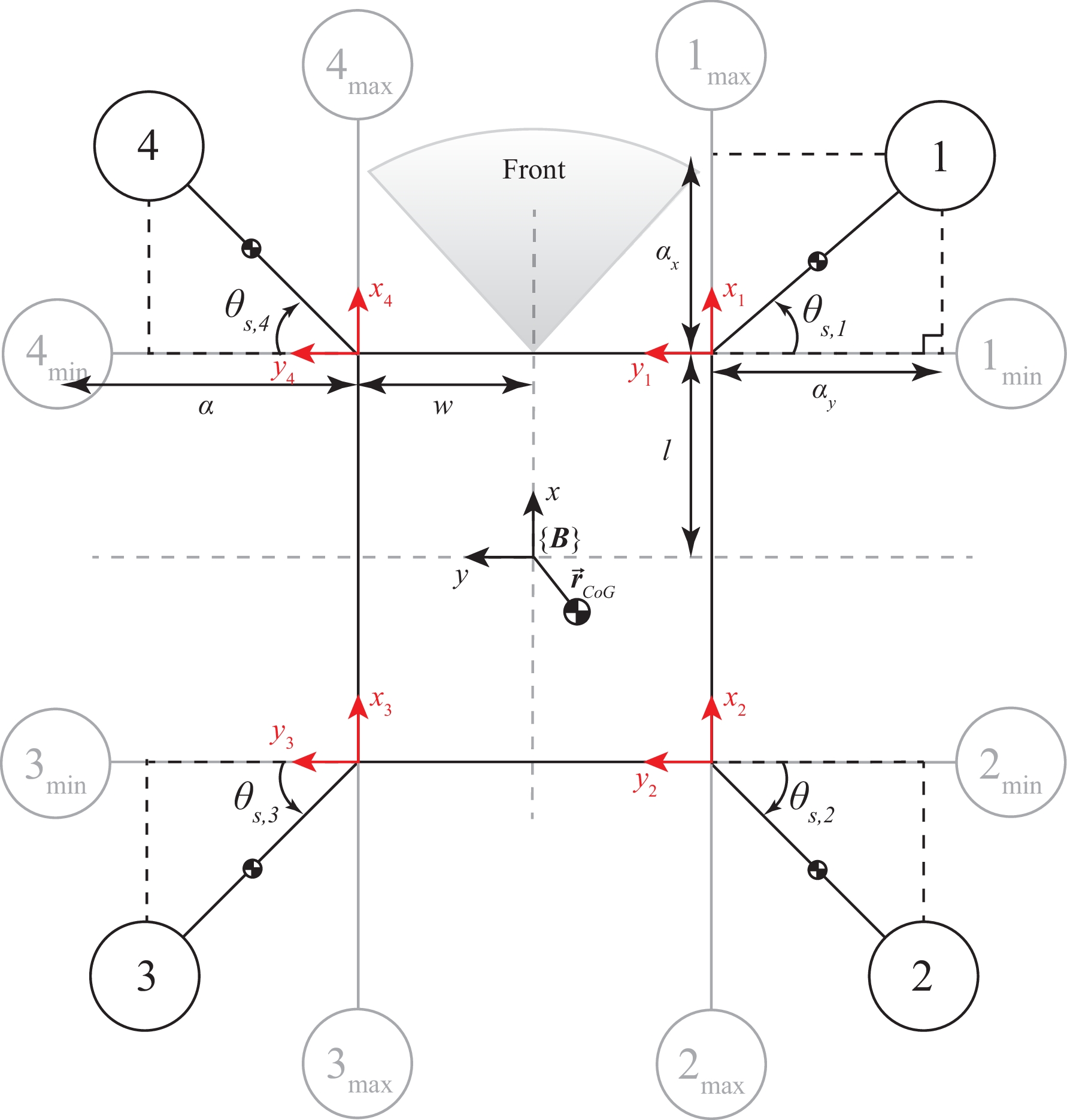}
\caption{2D representation of the reconfigurable quadrotor with the main geometrical properties highlighted}
\label{fig:2Dgeom}
\end{center}
\end{figure}

The notation in the 2D representation of the \gls{mav} in Fig.~\ref{fig:2Dgeom} is followed for the kinematics of the reconfigurable arms. Since the arms of the platform rotate only around the $z$-axis, the kinematics of the arms varies only on the $x-y$ axis. The dimensions of the main body are $(2w \times 2l)$ denoting the width and length of the platform. %
The arm length is denoted as $\alpha$ and the rotation angle of the servo as $\theta^s_i$ for $i=1,2,..,4$. The position of the arms' final point varies based on the angle $\theta^s_i$. The kinematics for the $x-y$ position vector $p_{arm,i}$ of the arms are given in~\eqref{eq:arm_kinematics}.
\begin{subequations}
\label{eq:arm_kinematics}
\begin{align}
&{\bm{p}_{arm,1} =  \left[\arraycolsep=2.2pt\def\arraystretch{1.3}\begin{array}{ccc} - w - \alpha \cos{\theta^{s}_{1}} , & l + \alpha \sin{\theta^{s}_{1}} \end{array}\right]^\top} \\ 
&{\bm{p}_{arm,2} =  \left[\arraycolsep=2.2pt\def\arraystretch{1.3}\begin{array}{ccc} - w - \alpha \cos{\theta^{s}_{2}} , & - l - \alpha \sin{\theta^{s}_{2}} \end{array}\right]^\top} \\ 
&{\bm{p}_{arm,3} =  \left[\arraycolsep=2.2pt\def\arraystretch{1.3}\begin{array}{ccc} w + \alpha \cos{\theta^{s}_{3}} , & - l - \alpha \sin{\theta^{s}_{3}} \end{array}\right]^\top} \\ 
&{\bm{p}_{arm,4} =  \left[\arraycolsep=2.2pt\def\arraystretch{1.3}\begin{array}{ccc} w + \alpha \cos{\theta^{s}_{4}} , &  l + \alpha \sin{\theta^{s}_{4}}  \end{array}\right]^\top}
\end{align}
\end{subequations}
As an example, the X-configuration is achieved when all $\theta^{s}_{1} = \theta^{s}_{2} =\theta^{s}_{3} =\theta^{s}_{4} = \pi/4 $ and the H-configuration which reduces the width of the \gls{mav} to its minimum is $\theta^{s}_{1} = \theta^{s}_{2} =\theta^{s}_{3} =\theta^{s}_{4} = \pi/2 $. The control action based on~\eqref{eq:modeluav} and~\eqref{eq:arm_kinematics} is $[T, \phi_d, \theta_d, \bm{\theta}^s]^\top$, where $\bm{\theta}^s=\{\theta^{s}_{i}, i\in \mathbb{N} \cap [1,4] , \theta_i^s \in\mathbb{R}^4  \cap [0,\pi/2]\}$.
%
\subsection{Objective Function}

The objective of the \gls{nmpc} is to track a reference trajectory from a high-level path planner $\bm{x}_r = [\bm{p}, \bm{v}, \phi, \theta]^\top$ and to generate control actions $\bm{u}=[\phi_d, \theta_d, T, \bm{\theta}^s]^\top$. The controller can guarantee for a safety distance from a priori known obstacles, while generates angle commands for the servos of the arms $\bm{\theta}^s$ to adapt the size of the \gls{mav} for specific type of obstacles. 
A low level controller (Fig.~\ref{fig:Controllerscheme}), depending on: 1) the present structural formation of the platform, 2) the control actions $[\phi_d, \theta_d, T, \bm{\theta}^s]^\top$, and 3) the current heading $\psi$ of the \gls{mav} generates the motor commands $\bm{n} = [n_1, \dots ,n_4]^{\top}$ and the servo commands $\bm{s} = [s_1, \dots, s_4]^{\top}$ for the platform to track the desired attitude and servo angles, a concept that has been also presented in \cite{falanga2018foldable,papadimitriou2020switching}. 

The \gls{nmpc} approach solves a finite-horizon problem at every time instant $k$ with the prediction horizon of $N \in \mathbb{N}^{\ge 2}$. The states and control actions are expressed by $\bm{x}_{k+j|k}$, and $\bm{u}_{k+j\mid k}$ respectively for $k+j,\,\forall j \in \{0,1, \dots, N-1\}$ steps ahead from the current time step $k$. For this purpose, the finite horizon cost function can be written as:
\begin{multline} \label{eq:costfunction} 
J = \sum_{j=0}^{N-1} 
  \underbrace{\|\bm{x}_{k+j+1|k}-\bm{x}_{r}\|_{\bm{Q}_x}^2}_\text{tracking error}      
   + \underbrace{\|\bm{u}_{k+j+1|k}-\bm{u}_{r}\|_{\bm{Q}_u}^2}_\text{actuation cost} \\
   + \underbrace{\|\bm{u}_{k+j|k}-\bm{u}_{k+j-1|k}\|_{\bm{Q}_{\Delta u}}^2 }_\text{smoothness cost} 
   + \underbrace{\|\bm{p}_{k+j+1|k}-\bm{p}_{obs}\|_{\bm{Q}_c}^2}_\text{tracking obstacle's center} .
\end{multline} 
The objective function consists of four terms. The first term ensures the tracking of the desired states $\bm{x}_{r}$ by minimizing the deviation from the current states. The second term, penalizes the deviation from the hover thrust with horizontal roll and pitch, and nominal X-configuration of the reconfigurable \gls{mav}, where $\bm{u}_{r}$ is $[g,\, 0,\, 0,\, \pi/4,\, \pi/4,\, \pi/4,\, \pi/4]^\top$. The third term tracks the aggressiveness of the obtained control actions. This term will guarantee that when the platform is not near an obstacle, the \gls{mav} will keep its optimal formation. The last term imposes on the platform to pass through the middle of the obstacle $\bm{p}_{obs} = [x_{obs},y_{obs},z_{obs}]^{\top}$, with the help of an adaptive weight $\bm{Q}_c$ which is enabled only when the \gls{mav} is in close proximity with an obstacle. In addition, the weights of the objective function's terms are denoted as $\bm{Q}_x \in \mathbb{R}^{8\times 8}$, $\bm{Q}_u\in \mathbb{R}^{7\times 7}$, $\bm{Q}_{\Delta u}\in \mathbb{R}^{7\times 7}$, and $\bm{Q}_{c}\in \mathbb{R}^{3\times 3}$ respectively, which they reflect to the relative importance of each term.
\subsection{Constraints}
\subsubsection{Input Constraint}

To bound the aggressive actuation, the following constraints are introduced for the angles and thrust rate of change $\Delta\phi$, $\Delta\theta$, $\Delta T$ and $\Delta \bm{\theta}^s$:
\begin{subequations} \label{eq:inputdelta_constraints}
\begin{align}
    |\phi_{d,k+j|k} - \phi_{d,k+j+1|k}| 
    {}\leq{}
    \Delta \phi_{\max},
    \\
    |\theta_{d,k+j|k}  - \theta_{d,k+j+1|k}| 
    {}\leq{}
    \Delta \theta_{\max},
    \\
    |\bm{\theta}^s_{k+j|k}  - \bm{\theta}^s_{k+j+1|k}| 
    {}\leq{}
    \Delta \bm{\theta}^s_{\max},
    \\
     |T_{k+j|k}  - T_{k+j+1|k}| 
    {}\leq{}
    \Delta T_{\max},
\end{align}
\end{subequations}
for $j = 0, \ldots, N-1$. Where $\Delta \phi_{\max}$, $\Delta \theta_{\max}$, $T$ and $\Delta \bm{\theta}^s_{\max}$ denote the minimum and maximum rates. Additionally the inputs are constrained within specific boundaries in the following form:
\begin{subequations} \label{eq:input_constraints}
\begin{align}
    [\bm{u}_{min} \leq \bm{u} \leq \bm{u}_{max}],
\end{align}
\end{subequations}
where $\bm{u}_{min} = [\phi_{min}, \theta_{min}, T_{min}, \bm{\theta}^s_{min}]^\top$ and $\bm{u}_{max} = [\phi_{max}, \theta_{max}, T_{max}, \bm{\theta}^s_{max}]^\top$  denote the minimum and maximum values the control actions can take. 

\subsubsection{Restricted Entrances}

There are different types of obstacles in the surrounding environment, nonetheless, all these types of obstacles can be categorized into three types: a) cylinder shape, b) polytope surface, and c) constrained entrance~\cite{lindqvist2020non}. In this work the control framework focus on changing the formation of a morphing quadrotor to pass through limited passages, thus only constrained entrance obstacles are considered. When the \gls{mav} is outside the obstacle, the associated cost is forced to be zero and for this purpose, the function $max(h,0)= [h]_+$ is utilized. With the proper selection of the $h$ expression, the constrained area is negative outside the obstacle and positive inside of it.
\paragraph{Cylindrical Entrances} A restricted entrance at $\bm{p}_{obs} = [x_{obs},y_{obs},z_{obs}]^{\top}$ with radius $r_{obs}$ is defined in \eqref{eq:hentrance}. The equation is defined in such way that is negative inside the radius of the obstacle and positive elsewhere. Thus, in order to guarantee this constraint the \gls{mav} should pass through the opening.
\begin{equation}\label{eq:hentrance}
h_{\text{entrance}} = -(r^{2}_{obs} - (y_{k+j|k} - y_{obs})^2 - (z_{k+j|k} - z_{obs})^2)
\end{equation}
The area of the restricted entrance along the $x$-axis is defined by~\eqref{eq:hentranceX}. While $x_{obs}$ is the center of the obstacle on $x$-axis the $l_1$ and $l_2$ define the extent of the passage for the left and right side of the entrance respectively. Depending on the entrance side either $h_{\text{xmin}}$ or $h_{\text{xmax}}$ will become positive.
\begin{subequations} 
\begin{align}\label{eq:hentranceX}
   & h_{\text{xmin}} = x_{k+j|k} - x_{obs} + l_{1} \\
   & h_{\text{xmax}} = -(x_{k+j|k} - x_{obs}) + l_{2}
\end{align}
\end{subequations}
Finally, the restricted entrance can be realized as a multiplication of the individual constraints~\eqref{eq:entranceconstrained}. The expression results in positive values when the obstacle constraint is violated and zero elsewhere.  
\begin{equation}\label{eq:entranceconstrained}
    C_1 = \sum_{j=0}^{N}[h_{\text{entrance}}]_+[h_{\text{xmin}}]_+[h_{\text{xmax}}]_+
\end{equation}

\paragraph{Cubic Entrances} This type of entrances are defined by their center coordinates $\bm{p}_{obs} = [x_{obs},y_{obs},z_{obs}]^{\top}$ and the length along $x$-axis $l_1+l_2$, width along $y$-axis $w_1$ and height along $z$-axis $h_1$. Thus we can define $y_{min} = y_{obs} - w_{1}/2$ and $y_{max} = y_{obs} + w_{1}/2$ and similar for $z$ axis. The proper h-function can is defined in \eqref{eq:hentranceX_rect} that describes the limits of the rectangular passage. 
 \begin{subequations} 
\begin{align}\label{eq:hentranceX_rect}
   & h_{\text{ymin}} = -(y_{k+j|k} - y_{min}), ~~ h_{\text{ymax}} = y_{k+j|k} - y_{max}\\
   & h_{\text{zmin}} = -(z_{k+j|k} - z_{min}), ~~ h_{\text{zmax}} = z_{k+j|k} - z_{max}
\end{align}
\end{subequations}
Finally, the rectangular restricted entrance constraint is defined in \eqref{eq:entranceconstrained_rect} with the help of \eqref{eq:hentranceX_rect} and \eqref{eq:hentranceX}. 
 
\begin{equation} \label{eq:entranceconstrained_rect}
    C_2 = \sum_{j=0}^{N}[h_{\text{entrance}}]_+ ( [h_{\text{zmin}}]_+ \land [h_{\text{zmax}}]_+ \land [h_{\text{ymin}}]_+ \land [h_{\text{ymax}}]_+   )
\end{equation}
where the $\land$ is the logical AND operator.
\subsubsection{Geometric Constraints}
The $h_d$ in \eqref{eq:mav2hole_cost} defines the limit from which the reconfigurable \gls{mav} need to adapt its size to match the properties of the restricted entrance. More specific $d_\text{safe}$ is the safety distance from an obstacle that the platform does not consider its formation. The second part of the equation is the euclidean distance between the platform and the center of the entrance. Thus, the $[h_{d}]_+$ is positive when the platform is closer than the $d_\text{safe}$ and zero elsewhere.

\begin{equation}\label{eq:mav2hole_cost}
    h_d =d_\text{safe} - \sqrt{(x - x_{obs})^2 + (y - y_{obs})^2 +(z - z_{obs})^2}
\end{equation}
The width of the reconfigurable \gls{mav}---including the variable position of the arms following the definition from \eqref{eq:arm_kinematics}---for the front and rear side of the platform can be defined as:
\begin{subequations}
\label{eq:mav_width}
\begin{align}
&r^2_{\text{front}} = (p^x_{arm,1} - p^x_{arm,4})^2 + (p^y_{arm,1} - p^y_{arm,4})^2 \\
&r^2_{\text{rear}} = (p^x_{arm,2} - p^x_{arm,3})^2 + (p^y_{arm,2} - p^y_{arm,3})^2 
\end{align}
\end{subequations}
Based on~\eqref{eq:mav_width} and with of the restricted entrance $w_{obs}$ the constraint \eqref{eq:arm_constr} is defined and it is positive when the width of the platform is larger than the width of the entrance. The $w_{obs}$ is $2r_{obs}$ and $w_1$ in case of spherical and cubic entrances respectively.
%
\begin{subequations}
\label{eq:arm_constr}
\begin{align}
h_{\text{front}} = r_{\text{front}} - w_{obs} \\
h_{\text{rear}} =  r_{\text{rear}} - w_{obs} 
\end{align}
\end{subequations}
Finally, in \eqref{eq:arm_constr_c}, the full constraint of the arms can be realized as a multiplication of the individual components.
\begin{subequations}
\label{eq:arm_constr_c}
\begin{align}
C_{\text{front}} = [h_d]_+[h_{\text{front}}]_+ \\
C_{\text{rear}} = [h_d]_+[h_{\text{rear}}]_+
\end{align}
\end{subequations}

\subsection{Embedded Optimization}
Based on the previous definitions, the following optimization problem is defined:
\begin{subequations}
\begin{alignat}{2}
&\!\min_{\{u_{k+j\mid k}\}^{N-1}_{j=0}}       & &\sum_{i=1}^{M}   J \quad  \forall i \in \{1, \dots, M\}\label{eq:optProb}\\
&\text{s.t.} &      & \bm{x}_{k+j+1\mid k}=f(\bm{x}_{k+j\mid k},\bm{u}_{k+j\mid k}),\label{eq:constraint1}\\
&                  &      & \bm{u}_{k+j\mid k} \in [\bm{u}_{\min},\bm{u}_{\max}],\label{eq:constraint2} \\
&                  &      & \text{Constraints}~ \eqref{eq:entranceconstrained},  \eqref{eq:inputdelta_constraints},\eqref{eq:input_constraints},\eqref{eq:arm_constr_c}. \label{eq:constraint5}
\end{alignat}
\end{subequations}
The developed \gls{nmpc} with passing through entrance constraints is solved by PANOC~\cite{sathya2018embedded} to assure real-time performance of the control framework. 

\section{Results}\label{sec:Results}

\subsection{Simulation Setup}

Out of the numerous simulations, three representative cases are selected to prove the performance of the proposed control scheme. During the simulations, the eight states of the \gls{mav} considered measurable, while the position of the arms are computed based on the desired angles provided from the \gls{nmpc}.

The parameters of the non-linear \gls{mav} model are the dimensions of the main body $(\unit[0.25\times 0.25\times0.15)]{m}$ denoting the length, width and height respectively and the length of the arms is $\unit[0.30]{m}$, the gravitational acceleration $g = \unit[9.81]{m/s^2}$, the mass normalized drag coefficients $A_x,A_y=0.1$ and $A_z =0.2$, and the time constants $t_{\phi},t_{\theta} = \unit[0.5]{sec}$ with gains $K_{\phi},K_{\theta} = 1$. For visualization purposes we selected longer arms to emphasize the changes in the control scheme.

The prediction horizon $N$ has been set to 40 and the sampling time at $t_s = \unit[0.05]{sec}$. The roll and pitch angles are constrained within $[-0.21 \leq \phi_{d}, \theta_{d} \leq 0.21],\ \unit{rad}$, while the angles change of rate is bounded between $[-0.1 \leq \Delta\phi, \Delta\theta \leq 0.1],\ \unit{rad}$. The rotation of the arms is constrained in $[0 \leq \bm{\theta}^s \leq \pi/2],\ \unit{rad}$. The weights of the states are $\bm{Q}_x = [5, 5, 9, 9, 9, 9, 1, 1]$, while the control action weights are $\bm{Q}_u = [5, 5, 10, 9, 9, 9, 9]$, and the weights for the input rate of change are $\bm{Q}_{\Delta u} = [20, 20 ,20, 10, 10,10,10]$. The weights for passing through the middle of the entrance are $\bm{Q}_{c} = [0, 10, 10]$ if the distance between the \gls{mav} and the center of the obstacle is $d \leq 1\unit{m}$  and  $\bm{Q}_{c} = [0, 0, 0]$ elsewhere. The $\bm{Q}_{c,1} = 0 $ for every case, since the obstacle is parallel with the $x$-axis by definition, thus to navigate through the center of the obstacle the \gls{mav} should pass at $p_{y} \approx p_{obs,y}$ and $p_{z} \approx p_{obs,z}$.
The simulation trials were running on a single core to solve the optimization problem and they performed in a personal computer with an Intel(R) Core(TM) i7-8550U @ $\unit[1.8]{Ghz}$ processor with $\unit[16]{GB}$ of RAM. For all the simulation the mean computation time is $t^{\text{comp.}}_{mean} = 10\unit{ms}$ and the max $t^{\text{comp.}}_{max} = 18\unit{ms}$ 

\subsection{Spherical Entrance Navigation}

\begin{figure}[htbp!] 
\setlength\fwidth{0.8\linewidth}
\centering
\input{Tikz/3Dplot_overlay.tex}
\caption{3D Path response of the reconfigurable quadrotor passing successfully through two restricted cylindrical entrances}
\label{fig:3Dplot}
\end{figure}
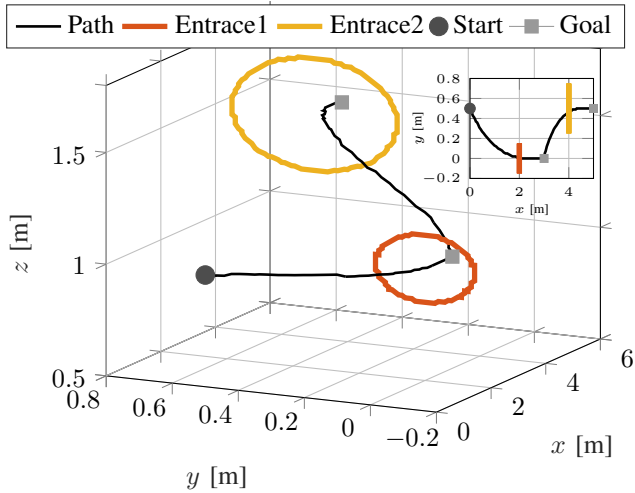

For the first simulation, the reconfigurable quadrotor passes through two narrow passages with different radii. Each time the gls{nmpc} provides the control actions to pass through the obstacle and to regulate the position of the arms to achieve the best balance between size reduction to avoid collision and maintain a morphology as close as possible to the X configuration. Fig.~\ref{fig:3Dplot} illustrates the path that the reconfigurable \gls{mav} follows when entering and exiting from the two cylindrical obstacles.
The first entrance is located at $\unit[(2,0,1)]{m}$ with radius $\unit[0.125]{m}$. The second entrance is located at $\unit[(4,0.5,1.5)]{m}$ with radius $\unit[0.25]{m}$. The initial position of the platform is at $\unit[(0,0.5,1)]{m}$ and the two sequential goal positions are $\unit[(3,0,1)]{m}$ and $\unit[(5,0.5,1.5)]{m}$ respectively.
The heading of the platform is set to look always towards to the obstacle and the safety distance between the obstacle and the platform is $d_{\text{safe}}= 0.3\unit{m}$.

\begin{figure}[htb]
\begin{center}
\includegraphics[width=\columnwidth]{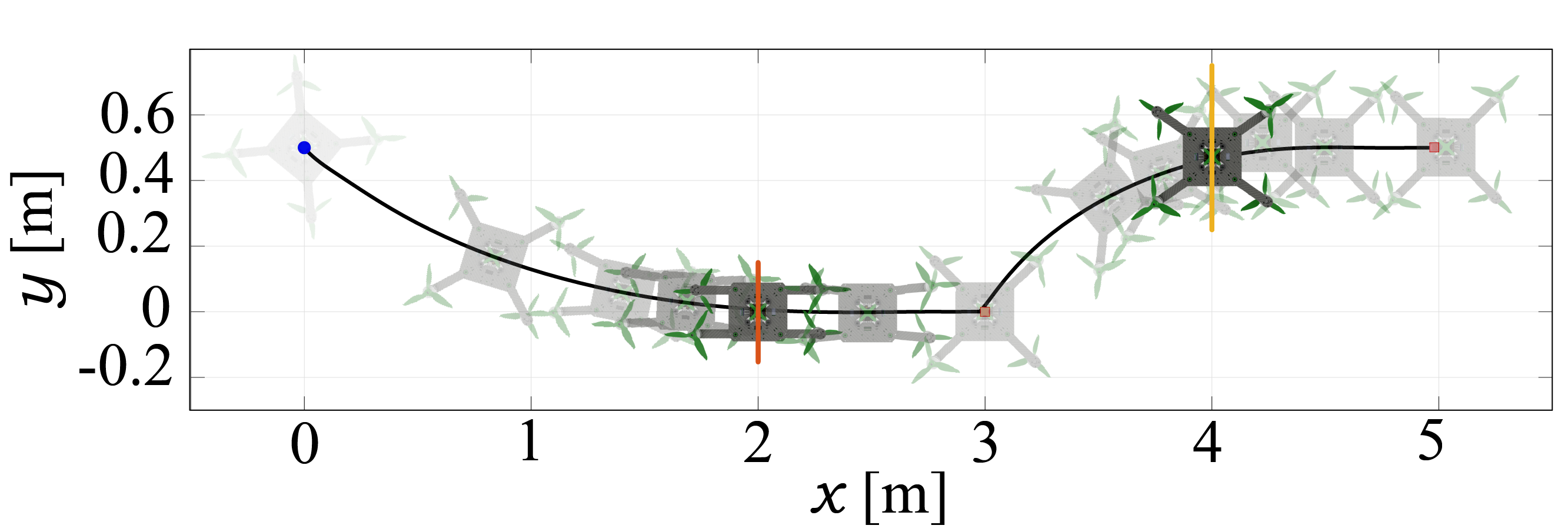}
\caption{Top view of the \gls{mav} navigation in case of multiple spherical entrances.}
\label{fig:2dviewcasde1}
\end{center}
\end{figure}

The top view of the \gls{mav} trajectory and spherical entrances is presented in Fig.~\ref{fig:2dviewcasde1}. As it can be seen, the \gls{mav} reconfigure the arm positions to pass through the entrances.

\begin{figure}[htbp!] 
\setlength\fwidth{0.8\linewidth}
\centering
\input{Tikz/trp.tex}
\caption{The generated $\phi, \theta$ and $T$ for the \gls{mav} navigation in case of multiple spherical entrances.}
\label{fig:RPT1case}
\end{figure}
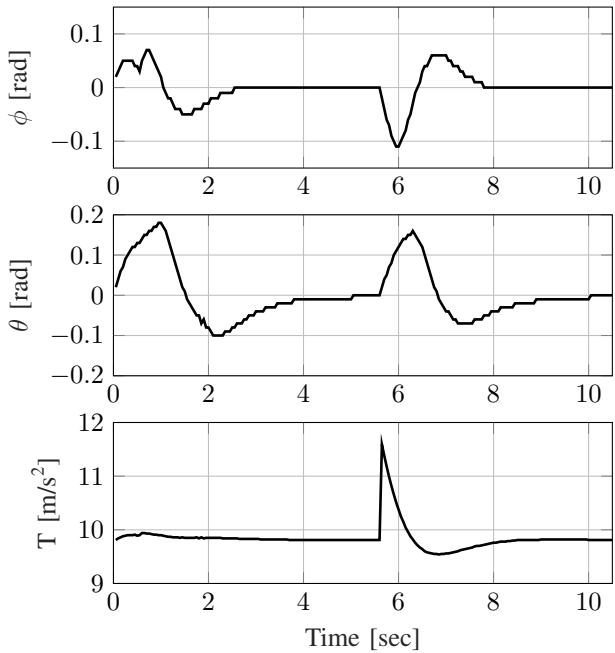

In Fig.~\ref{fig:RPT1case} presents the control actions of $\phi, \theta$ and $T$ the \gls{nmpc} controller provides for the platform to navigate to the final destination. It can be observed that the control actions are smooth and there is no violation of the input constraints.  

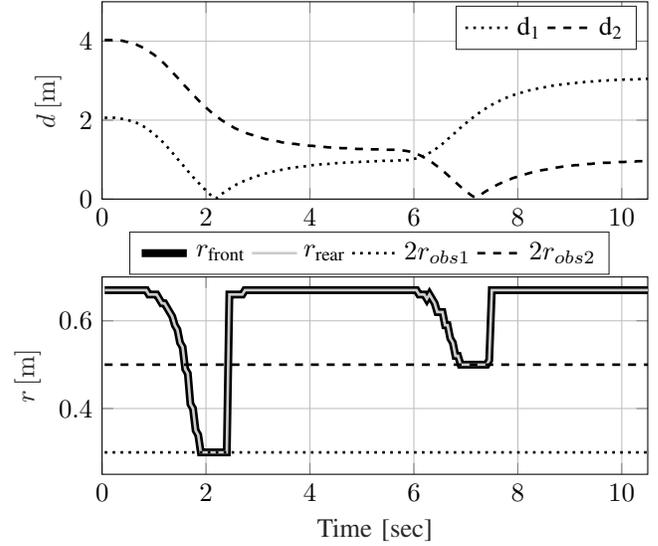
\begin{figure}[htbp!] 
\setlength\fwidth{0.8\linewidth}
\centering
\input{Tikz/config.tex}
\caption{Top: The distance between the \gls{mav} and spherical entrances. Bottom: The front and rear arm distances of the \gls{mav}, while the diameter of the spherical entrances are depicted.}
\label{fig:distances_arms}
\end{figure}

 The top of Fig.~\ref{fig:distances_arms} shows the time response of the euclidean distance between the two obstacles and the platform, while the bottom Fig.~\ref{fig:distances_arms} shows the distance between the two front and two rear arms and in extend the width of the platform, as well as the radius of the obstacles. It can be noticed, when the platform approaches the entrance of the obstacles it reduces its size and as soon as it passes through the obstacles it returns to the initial X-configuration. More specific at the $\unit[2]{sec}$ where the platform reaches the obstacle it has already adapt the size of the radius of the small obstacle. Approximately at the time instant of $\unit[7]{sec}$ the reconfigurable \gls{mav} approaches the second entrance and it adapts its size to pass through the restricted entrance. It should be noted that the \gls{nmpc} provides larger arm distance for the second entrance as it has slightly larger size compare to first entrance.

\subsection{Impassable Entrance}

\begin{figure}[htbp!] 
\setlength\fwidth{0.8\linewidth}
\centering
\input{Tikz/3Dplot_not_pass2.tex}
\caption{The trajectory of the \gls{mav} with three different initial positions, while the entrance is impassable as its diameter violate the minimum dimension of the \gls{mav}. }
\label{fig:3Dplot_not_pass}
\end{figure}
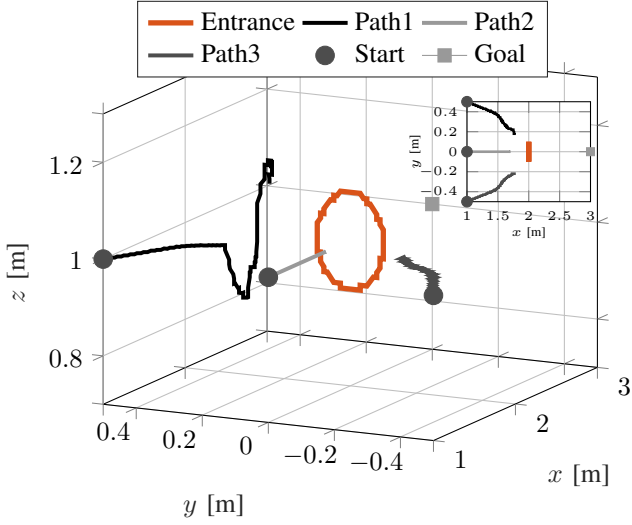

For this case, the reconfigurable \gls{mav} is placed near an extremely narrow passage and it is desired to pass through it. The radius of the obstacle is smaller than the minimum possible size that the platform can adapt i.e. $r_{obs} < r_{front,rear}^{\text{min}}$. In Fig.~\ref{fig:3Dplot_not_pass}, three different starting points have been selected, located at $\unit[(1,0,1)]{m}$, $\unit[(1,0.5,1)]{m}$ and $\unit[(1,-0.5,1)]{m}$ and the goal point is located at $\unit[(0,3,1]){m}$. For all the starting points the \gls{nmpc} succeeds to stop and avoid collision with the obstacle and as a result it cannot reach the final way-point. The safety distance between the obstacle and the platform is $d_{\text{safe}}= 0.3\unit{m}$.

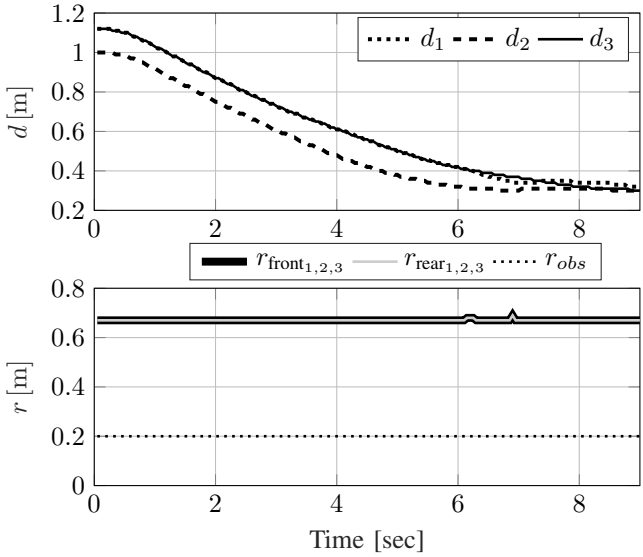
\begin{figure}[htbp!] 
\setlength\fwidth{0.8\linewidth}
\centering
\input{Tikz/config_not_pass2.tex}
\caption{Top: the distance of the \gls{mav} with the spherical entrance. Bottom: the front and rear arm distances when the spherical entrance is impassable.}
\label{fig:distances_arms_not_pass}
\end{figure}

For this scenario, the controller identifies that it cannot pass through the entrance without violating the constraint \eqref{eq:arm_constr}. Since the distance between the obstacle and the \gls{mav} is $d \geq d_{\text{safe}}$ the platform maintains the X-configuration (Fig.~\ref{fig:distances_arms_not_pass}). The same behavior is observed independent of the starting point for all simulations, thus the distance between the arms $r_{\text{front}_{1,2,3}}$ and $r_{\text{rear}_{1,2,3}}$ is the same for all the considered three cases. 

\subsection{Cubic Entrance}

\begin{figure}[htbp!] 
\setlength\fwidth{0.8\linewidth}
\centering
\input{Tikz/3Dplot_rect.tex}
\caption{3D Path response of the reconfigurable \gls{mav} passing successfully through two cubic entrances, while the 2D path is overlayed to highlight the tracking and obstacle avoidance performance.}
\label{fig:3Dplot_rect}
\end{figure}
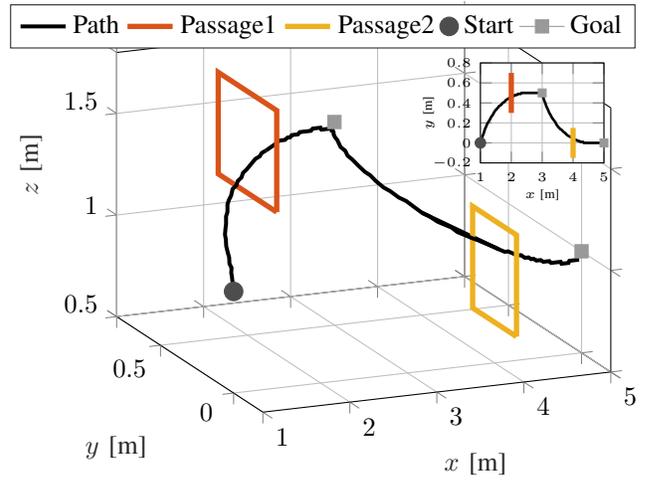

For the last presented case, the reconfigurable \gls{mav} passes through two narrow rectangular shaped entrances with different width. The \gls{nmpc} provides the proper commands for the arms to optimize the size of the \gls{mav} to navigate through the opening. Fig.~\ref{fig:3Dplot_rect} illustrates the path that the \gls{mav} follow when entering and exiting the two cubic restricted entrances.

The center of the first rectangular opening is located at $\unit[(2,0.5,1.5)]{m}$ with dimensions $(\unit[0.1\times 0.4\times0.5)]{m}$. The second entrance is located at $\unit[(4,0,1)]{m}$ with dimensions $(\unit[0.1\times 0.3\times0.5)]{m}$. The initial position of the platform is at $\unit[(1,0,1)]{m}$ and the two sequential goal positions are at $\unit[(3,0.5,1.5)]{m}$ and $\unit[(5,0,1)]{m}$ respectively.
The heading of the platform is set to look always forward to the obstacle and the safety distance between the obstacle and the platform is $d_{\text{safe}}= 0.3\unit{m}$.

\begin{figure}[htbp!] 
\setlength\fwidth{0.8\linewidth}
\centering
\input{Tikz/config_rect.tex}
\caption{Top: the distance of the \gls{mav} to the cubic entrances. Bottom: the front and rear arm distances in case of cubic entrance, while the maximum width of the rectangle is depicted.}
\label{fig:distances_arms_rect}
\end{figure}
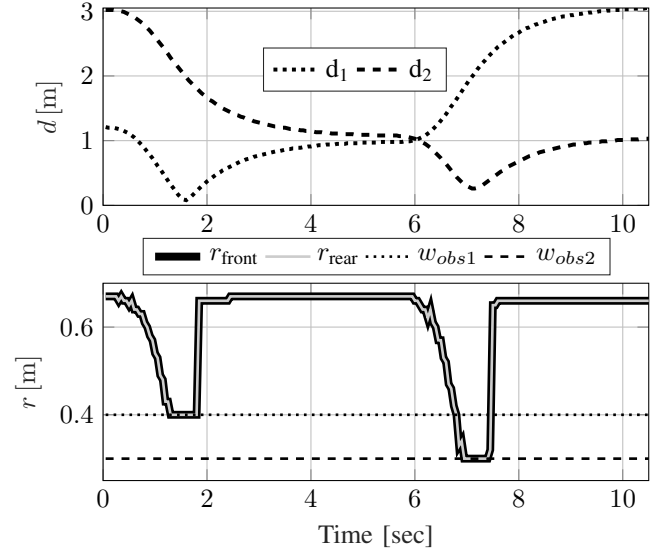  

\section{Conclusion}\label{sec:Conclusions}
This article proposed a novel \gls{nmpc} framework for collision-free navigation of a morphing \gls{mav}, through restricted entrances. The reconfigurable \gls{mav} can alter the position of the arms to adapt its shape based on information about its environment.
More specifically, the \gls{nmpc} was able to provide the necessary control actions to navigate through narrow entrances adapting its size, while flying in two different types of entrances, namely a spherical and a cubic. When the size of the entrance was smaller than the the size of the platform, the \gls{mav} is not attempting to pass through the entrance. In all scenarios the proposed framework provided a collision free path for the \gls{mav}. Future work, will tackle the problem from an experimental point of view as well as to extent our methodology to more complex entrances with non-convex shapes.

\bibliographystyle{IEEEtran}
\bibliography{mybib}

\end{document}

%% file: abbreviations.tex
\newacronym{ahmpc}{AHMPC}{Adaptive Horizon Model Predictive Control}
\newacronym{admm}{ADMM}{Alternating Direction Method of Multipliers}
\newacronym{ahe}{AHE}{Adaptive Histogram Equalization}
\newacronym{ar}{AR}{Augmented Reality}
\newacronym{cdf}{CDF}{Cumulative Distribution Function}
\newacronym{clahe}{CLAHE}{Contrast Limited Adaptive Histogram Equalization}
\newacronym{cnn}{CNN}{Convolutional Neural Network}
\newacronym{dbscan}{DBSCAN}{Density-Based Spatial Clustering of Applications with Noise }
\newacronym{dl}{DL}{Deep Learning}
\newacronym{dof}{DoF}{Degree of Freedom}
\newacronym{ekf}{EKF}{Extended Kalman Filter}
\newacronym{fbe}{FBE}{Forward Backward Envelope}
\newacronym{fov}{FoV}{Field of View}
\newacronym{fps}{fps}{Frame Per Second}
\newacronym{gnss}{GNSS}{Global Navigation Satellite System}
\newacronym{gps}{GPS}{Global Positioning System}
\newacronym{iid}{IID}{Independent Identically Distributed}
\newacronym{imu}{IMU}{Inertial Measurement Unit}
\newacronym{kf}{KF}{Extended Kalman Filter}
\newacronym{llwcf}{LLWCF}{Left Local Wall Continuum Function}
\newacronym{lrcn}{LRCN}{ Long-Term Recurrent Convolutional Network}
\newacronym{lwcf}{LWCF}{Local Wall Continuum Function}
\newacronym{mae}{MAE}{mean absolute error}
\newacronym{mav}{MAV}{Micro Aerial Vehicle}
\newacronym{mhe}{MHE}{Moving Horizon Estimation}
\newacronym{mimo}{MIMO}{Multiple Input Multiple Output}
\newacronym{mlp}{MLP}{MultiLayer Perceptron}
\newacronym{mpc}{MPC}{Model Predictive Control}
\newacronym{msf}{MSF}{Multi Sensor Fusion}
\newacronym{nmhe}{NMHE}{Nonlinear Moving Horizon Estimation}
\newacronym{nmpc}{NMPC}{Nonlinear Model Predictive Control}
\newacronym{nn}{NN}{Nueral Network}
\newacronym{nwu}{NWU}{North-West-Up}
\newacronym{open}{OpEn}{Optimization Engine}
\newacronym{panoc}{PANOC}{Proximal Averaged Newton-type method for Optimal Control}
\newacronym{pdf}{PDF}{Probability Density Function}
\newacronym{pid}{PID}{Proportional Integral Derivative}
\newacronym{psd}{PSD}{ Positive Semi-Definite }
\newacronym{rbf}{RBF}{ Radial Basis Function}
\newacronym{relu}{ReLU}{Rectified Linear Unit}
\newacronym{rl}{RL}{Reinforcement Learning}
\newacronym{rlwcf}{RLWCF}{Right Local Wall Continuum Function}
\newacronym{rmse}{RMSE}{Root Mean Square Error}
\newacronym{ros}{ROS}{Robot Operating System}
\newacronym{sfm}{SfM}{Structure from Motion}
\newacronym{spams}{SPAMS}{SPArse Modeling Software}
\newacronym{sqp}{SQP}{Sequential Quadratic Programming}
\newacronym{slam}{SLAM}{Simultaneous Localization and Mapping}
\newacronym{ugv}{UGV}{Unmanned Ground Vehicle}
\newacronym{ukf}{UKF}{Unscented Kalman Filter}
\newacronym{vi}{VI}{Visual Inertia}
\newacronym{vo}{VO}{Visual Odometry}

%% file: Tikz/3Dplot_overlay.tex
%
%
\definecolor{mycolor1}{rgb}{0.85000,0.32500,0.09800}%
\definecolor{mycolor2}{rgb}{0.92900,0.69400,0.12500}%
\definecolor{mycolor3}{rgb}{0.46600,0.67400,0.18800}%
\definecolor{mycolor4}{rgb}{0.30100,0.74500,0.93300}%
\begin{tikzpicture}

\begin{axis}[%
width=0.951\fwidth,
height=0.77\fwidth,
at={(0\fwidth,1.1\fwidth)},
scale only axis,
xmin=0.00,
xmax=6.00,
tick align=outside,
xlabel style={font=\color{white!15!black}},
xlabel={$x$ [m]},
ymin=-0.20,
ymax=0.80,
ylabel style={font=\color{white!15!black}},
ylabel={$y$ [m]},
zmin=0.50,
zmax=1.80,
zlabel style={font=\color{white!15!black}},
zlabel={$z$ [m]},
view={-63.60}{15.71},
axis background/.style={fill=white},
xmajorgrids,
ymajorgrids,
zmajorgrids,
legend style={at={(1.05,1.02)},legend cell align=left, align=left, legend columns=5, draw=white!15!black}
]
\addplot3 [color=black, line width=1.0pt]
 table[row sep=crcr] {%
0.00	0.50	1.00\\
0.00	0.50	1.00\\
0.00	0.50	1.00\\
0.00	0.50	1.00\\
0.00	0.50	1.00\\
0.01	0.49	1.00\\
0.02	0.49	1.00\\
0.03	0.48	1.00\\
0.04	0.47	1.00\\
0.05	0.47	1.00\\
0.07	0.46	1.00\\
0.09	0.45	1.00\\
0.11	0.43	1.01\\
0.14	0.42	1.01\\
0.17	0.41	1.01\\
0.21	0.39	1.01\\
0.24	0.37	1.01\\
0.29	0.36	1.01\\
0.33	0.34	1.01\\
0.38	0.31	1.01\\
0.44	0.29	1.01\\
0.49	0.27	1.01\\
0.56	0.25	1.01\\
0.62	0.23	1.01\\
0.69	0.21	1.01\\
0.76	0.19	1.01\\
0.83	0.17	1.01\\
0.91	0.15	1.00\\
};
\addplot3 [color=black, line width=1.0pt, forget plot]
 table[row sep=crcr] {%
2.26	-0.00	1.00\\
2.30	-0.00	1.00\\
2.34	-0.00	1.00\\
2.38	-0.00	1.00\\
2.41	-0.00	1.00\\
2.44	-0.00	1.00\\
2.47	-0.00	1.00\\
2.50	-0.00	1.00\\
2.52	-0.00	1.00\\
2.55	-0.00	1.00\\
2.57	-0.00	1.00\\
2.59	-0.00	1.00\\
2.61	-0.00	1.00\\
2.63	-0.00	1.00\\
2.65	-0.00	1.00\\
2.67	-0.00	1.00\\
2.68	-0.00	1.00\\
2.70	0.00	1.00\\
2.71	0.00	1.00\\
2.73	0.00	1.00\\
2.74	0.00	1.00\\
2.75	0.00	1.00\\
2.77	0.00	1.00\\
2.78	0.00	1.00\\
2.79	0.00	1.00\\
2.80	0.00	1.00\\
2.81	0.00	1.00\\
2.82	0.00	1.00\\
2.82	0.00	1.00\\
2.83	0.00	1.00\\
2.84	0.00	1.00\\
2.85	0.00	1.00\\
2.86	0.00	1.00\\
2.86	0.00	1.00\\
2.87	0.00	1.00\\
2.88	0.00	1.00\\
2.88	0.00	1.00\\
2.89	0.00	1.00\\
2.89	0.00	1.00\\
2.90	0.00	1.00\\
2.90	0.00	1.00\\
2.91	0.00	1.00\\
2.91	0.00	1.00\\
2.92	0.00	1.00\\
2.92	0.00	1.00\\
2.93	0.00	1.00\\
2.93	0.00	1.00\\
2.93	0.00	1.00\\
2.94	0.00	1.00\\
2.94	0.00	1.00\\
2.94	0.00	1.00\\
2.95	0.00	1.00\\
2.95	0.00	1.00\\
2.95	0.00	1.00\\
2.95	0.00	1.00\\
2.96	0.00	1.00\\
2.96	0.00	1.00\\
2.96	0.00	1.00\\
2.96	0.00	1.00\\
2.97	0.00	1.00\\
2.97	0.00	1.00\\
2.97	0.00	1.00\\
2.97	0.00	1.00\\
2.97	0.00	1.00\\
2.97	0.00	1.00\\
2.98	0.00	1.00\\
2.98	0.00	1.01\\
2.98	0.00	1.02\\
2.98	0.01	1.04\\
2.99	0.01	1.06\\
3.00	0.02	1.07\\
3.01	0.03	1.09\\
3.02	0.04	1.11\\
3.04	0.06	1.14\\
};
\addlegendentry{Path}

\addplot3 [color=mycolor1, line width=2.0pt]
 table[row sep=crcr] {%
2.00	0.00	1.15\\
2.00	0.01	1.15\\
2.00	0.02	1.15\\
2.00	0.03	1.15\\
2.00	0.04	1.15\\
2.00	0.05	1.14\\
2.00	0.06	1.14\\
2.00	0.06	1.14\\
2.00	0.07	1.13\\
2.00	0.08	1.13\\
2.00	0.09	1.12\\
2.00	0.10	1.11\\
2.00	0.10	1.11\\
2.00	0.11	1.10\\
2.00	0.12	1.09\\
2.00	0.12	1.09\\
2.00	0.13	1.08\\
2.00	0.13	1.07\\
2.00	0.14	1.06\\
2.00	0.14	1.05\\
2.00	0.14	1.04\\
2.00	0.15	1.04\\
2.00	0.15	1.03\\
2.00	0.15	1.02\\
2.00	0.15	1.01\\
2.00	0.15	1.00\\
2.00	0.15	0.99\\
2.00	0.15	0.98\\
2.00	0.15	0.97\\
2.00	0.14	0.96\\
2.00	0.14	0.95\\
2.00	0.14	0.94\\
2.00	0.13	0.93\\
2.00	0.13	0.93\\
2.00	0.12	0.92\\
2.00	0.12	0.91\\
2.00	0.11	0.90\\
2.00	0.11	0.89\\
2.00	0.10	0.89\\
2.00	0.09	0.88\\
2.00	0.09	0.88\\
2.00	0.08	0.87\\
2.00	0.07	0.87\\
2.00	0.06	0.86\\
2.00	0.05	0.86\\
2.00	0.04	0.86\\
2.00	0.03	0.85\\
2.00	0.02	0.85\\
2.00	0.01	0.85\\
2.00	0.00	0.85\\
2.00	-0.00	0.85\\
2.00	-0.01	0.85\\
2.00	-0.02	0.85\\
2.00	-0.03	0.85\\
2.00	-0.04	0.86\\
2.00	-0.05	0.86\\
2.00	-0.06	0.86\\
2.00	-0.07	0.87\\
2.00	-0.08	0.87\\
2.00	-0.09	0.88\\
2.00	-0.09	0.88\\
2.00	-0.10	0.89\\
2.00	-0.11	0.89\\
2.00	-0.11	0.90\\
2.00	-0.12	0.91\\
2.00	-0.12	0.92\\
2.00	-0.13	0.92\\
2.00	-0.13	0.93\\
2.00	-0.14	0.94\\
2.00	-0.14	0.95\\
2.00	-0.14	0.96\\
2.00	-0.15	0.97\\
2.00	-0.15	0.98\\
2.00	-0.15	0.99\\
2.00	-0.15	1.00\\
2.00	-0.15	1.01\\
2.00	-0.15	1.02\\
2.00	-0.15	1.03\\
2.00	-0.15	1.04\\
2.00	-0.14	1.04\\
2.00	-0.14	1.05\\
2.00	-0.14	1.06\\
2.00	-0.13	1.07\\
2.00	-0.13	1.08\\
2.00	-0.12	1.09\\
2.00	-0.12	1.09\\
2.00	-0.11	1.10\\
2.00	-0.10	1.11\\
2.00	-0.10	1.11\\
2.00	-0.09	1.12\\
2.00	-0.08	1.13\\
2.00	-0.07	1.13\\
2.00	-0.06	1.14\\
2.00	-0.06	1.14\\
2.00	-0.05	1.14\\
2.00	-0.04	1.15\\
2.00	-0.03	1.15\\
2.00	-0.02	1.15\\
2.00	-0.01	1.15\\
2.00	-0.00	1.15\\
};
 \addlegendentry{Entrace1}

\addplot3 [color=black, line width=1.0pt, forget plot]
 table[row sep=crcr] {%
0.91	0.15	1.00\\
0.99	0.13	1.00\\
1.06	0.12	1.00\\
1.14	0.10	1.00\\
1.22	0.09	1.00\\
1.29	0.08	1.00\\
1.37	0.06	1.00\\
1.44	0.05	1.00\\
1.51	0.05	1.00\\
1.58	0.04	1.00\\
1.65	0.03	1.00\\
1.72	0.02	1.00\\
1.79	0.02	1.00\\
1.85	0.02	1.00\\
1.91	0.01	1.00\\
1.97	0.01	1.00\\
2.02	0.01	1.00\\
2.08	0.00	1.00\\
2.13	0.00	1.00\\
2.17	0.00	1.00\\
2.22	0.00	1.00\\
2.26	-0.00	1.00\\
};

\addplot3 [color=mycolor2, line width=2.0pt]
 table[row sep=crcr] {%
4.00	0.50	1.75\\
4.00	0.52	1.75\\
4.00	0.53	1.75\\
4.00	0.55	1.75\\
4.00	0.56	1.74\\
4.00	0.58	1.74\\
4.00	0.59	1.73\\
4.00	0.61	1.73\\
4.00	0.62	1.72\\
4.00	0.64	1.71\\
4.00	0.65	1.70\\
4.00	0.66	1.69\\
4.00	0.67	1.68\\
4.00	0.68	1.67\\
4.00	0.69	1.66\\
4.00	0.70	1.65\\
4.00	0.71	1.63\\
4.00	0.72	1.62\\
4.00	0.73	1.60\\
4.00	0.73	1.59\\
4.00	0.74	1.57\\
4.00	0.74	1.56\\
4.00	0.75	1.54\\
4.00	0.75	1.53\\
4.00	0.75	1.51\\
4.00	0.75	1.50\\
4.00	0.75	1.48\\
4.00	0.75	1.46\\
4.00	0.74	1.45\\
4.00	0.74	1.43\\
4.00	0.74	1.42\\
4.00	0.73	1.40\\
4.00	0.72	1.39\\
4.00	0.72	1.38\\
4.00	0.71	1.36\\
4.00	0.70	1.35\\
4.00	0.69	1.34\\
4.00	0.68	1.32\\
4.00	0.67	1.31\\
4.00	0.65	1.30\\
4.00	0.64	1.29\\
4.00	0.63	1.29\\
4.00	0.61	1.28\\
4.00	0.60	1.27\\
4.00	0.59	1.27\\
4.00	0.57	1.26\\
4.00	0.56	1.26\\
4.00	0.54	1.25\\
4.00	0.52	1.25\\
4.00	0.51	1.25\\
4.00	0.49	1.25\\
4.00	0.48	1.25\\
4.00	0.46	1.25\\
4.00	0.44	1.26\\
4.00	0.43	1.26\\
4.00	0.41	1.27\\
4.00	0.40	1.27\\
4.00	0.39	1.28\\
4.00	0.37	1.29\\
4.00	0.36	1.29\\
4.00	0.35	1.30\\
4.00	0.33	1.31\\
4.00	0.32	1.32\\
4.00	0.31	1.34\\
4.00	0.30	1.35\\
4.00	0.29	1.36\\
4.00	0.28	1.38\\
4.00	0.28	1.39\\
4.00	0.27	1.40\\
4.00	0.26	1.42\\
4.00	0.26	1.43\\
4.00	0.26	1.45\\
4.00	0.25	1.46\\
4.00	0.25	1.48\\
4.00	0.25	1.50\\
4.00	0.25	1.51\\
4.00	0.25	1.53\\
4.00	0.25	1.54\\
4.00	0.26	1.56\\
4.00	0.26	1.57\\
4.00	0.27	1.59\\
4.00	0.27	1.60\\
4.00	0.28	1.62\\
4.00	0.29	1.63\\
4.00	0.30	1.65\\
4.00	0.31	1.66\\
4.00	0.32	1.67\\
4.00	0.33	1.68\\
4.00	0.34	1.69\\
4.00	0.35	1.70\\
4.00	0.36	1.71\\
4.00	0.38	1.72\\
4.00	0.39	1.73\\
4.00	0.41	1.73\\
4.00	0.42	1.74\\
4.00	0.44	1.74\\
4.00	0.45	1.75\\
4.00	0.47	1.75\\
4.00	0.48	1.75\\
4.00	0.50	1.75\\
};
 \addlegendentry{Entrace2}

\addplot3 [color=black, line width=1.0pt, forget plot]
 table[row sep=crcr] {%
3.04	0.06	1.14\\
3.06	0.07	1.16\\
3.08	0.09	1.18\\
3.11	0.12	1.20\\
3.14	0.14	1.22\\
3.17	0.16	1.24\\
3.21	0.19	1.26\\
3.25	0.21	1.28\\
3.30	0.24	1.30\\
3.34	0.26	1.32\\
3.39	0.29	1.33\\
3.44	0.31	1.35\\
3.50	0.33	1.36\\
3.55	0.35	1.38\\
3.60	0.37	1.39\\
3.66	0.39	1.40\\
3.71	0.40	1.42\\
3.77	0.42	1.43\\
3.82	0.43	1.44\\
3.88	0.44	1.45\\
3.93	0.45	1.45\\
3.98	0.46	1.46\\
4.03	0.47	1.47\\
4.07	0.48	1.48\\
4.12	0.48	1.48\\
4.16	0.49	1.49\\
4.21	0.49	1.49\\
4.25	0.49	1.49\\
4.28	0.49	1.50\\
4.32	0.50	1.50\\
4.36	0.50	1.50\\
4.39	0.50	1.51\\
4.42	0.50	1.51\\
4.45	0.50	1.51\\
4.48	0.50	1.51\\
4.50	0.50	1.51\\
4.53	0.50	1.51\\
4.55	0.50	1.51\\
4.57	0.50	1.51\\
4.59	0.50	1.51\\
4.61	0.50	1.51\\
4.63	0.50	1.51\\
4.65	0.50	1.51\\
4.67	0.50	1.51\\
4.68	0.50	1.51\\
4.70	0.50	1.51\\
4.71	0.50	1.51\\
4.73	0.50	1.51\\
4.74	0.50	1.51\\
4.75	0.50	1.51\\
4.77	0.50	1.51\\
4.78	0.50	1.51\\
4.79	0.50	1.51\\
4.80	0.50	1.51\\
4.81	0.50	1.51\\
4.82	0.50	1.51\\
4.83	0.50	1.50\\
4.83	0.50	1.50\\
4.84	0.50	1.50\\
4.85	0.50	1.50\\
4.86	0.50	1.50\\
4.86	0.50	1.50\\
4.87	0.50	1.50\\
4.88	0.50	1.50\\
4.88	0.50	1.50\\
4.89	0.50	1.50\\
4.89	0.50	1.50\\
4.90	0.50	1.50\\
4.90	0.50	1.50\\
4.91	0.50	1.50\\
4.91	0.50	1.50\\
4.92	0.50	1.50\\
4.92	0.50	1.50\\
4.92	0.50	1.50\\
4.93	0.50	1.50\\
4.93	0.50	1.50\\
4.93	0.50	1.50\\
4.94	0.50	1.50\\
4.94	0.50	1.50\\
4.94	0.50	1.50\\
4.95	0.50	1.50\\
4.95	0.50	1.50\\
4.95	0.50	1.50\\
4.95	0.50	1.50\\
4.96	0.50	1.50\\
4.96	0.50	1.50\\
4.96	0.50	1.50\\
4.96	0.50	1.50\\
4.96	0.50	1.50\\
};

\addplot3 [color=white!30!black!, only marks, mark size=3.5pt, mark=*, mark options={solid, fill=white!30!black!}]
 table[row sep=crcr] {%
0.00	0.50	1.00\\
};
\addlegendentry{Start}
 \addplot3 [color=white!60!black!, mark size=2.5pt, mark=square*, mark options={solid, fill=white!60!black!}]
 table[row sep=crcr] {%
3.00	0.00	1.00\\
};
\addlegendentry{Goal}
 \addplot3 [color=white!60!black!, mark size=2.5pt, mark=square*, mark options={solid, fill=white!60!black!}]
 table[row sep=crcr] {%
5.00	0.50	1.50\\
};
 \end{axis}

\begin{axis}[%
width=0.2377\fwidth,
height=0.1925\fwidth,
at={(0.70\fwidth,1.55\fwidth)},
scale only axis,
xmin=0.00,
xmax=5.00,
xlabel style={font=\color{white!15!black},font=\tiny,shift={(0.0,0.1)}},
xlabel={$x$ [m]},
ymin=-0.20,
ymax=0.80,
ylabel style={font=\color{white!15!black},font=\tiny,shift={(0.0,-0.2)}},
ticklabel style = {font=\tiny},  
ylabel={$y$ [m]},
axis background/.style={fill=white},
xmajorgrids,
ymajorgrids
]
\addplot [color=black, line width=1.0pt, forget plot]
  table[row sep=crcr]{%
0.00	0.50\\
0.00	0.50\\
0.00	0.50\\
0.00	0.50\\
0.00	0.50\\
0.01	0.49\\
0.02	0.49\\
0.03	0.48\\
0.04	0.47\\
0.05	0.47\\
0.07	0.46\\
0.09	0.45\\
0.11	0.43\\
0.14	0.42\\
0.17	0.41\\
0.21	0.39\\
0.24	0.37\\
0.29	0.36\\
0.33	0.34\\
0.38	0.31\\
0.44	0.29\\
0.49	0.27\\
0.56	0.25\\
0.62	0.23\\
0.69	0.21\\
0.76	0.19\\
0.83	0.17\\
0.91	0.15\\
0.99	0.13\\
1.06	0.12\\
1.14	0.10\\
1.22	0.09\\
1.29	0.08\\
1.37	0.06\\
1.44	0.05\\
1.51	0.05\\
1.58	0.04\\
1.65	0.03\\
1.72	0.02\\
1.79	0.02\\
1.85	0.02\\
1.91	0.01\\
1.97	0.01\\
2.02	0.01\\
2.08	0.00\\
2.13	0.00\\
2.17	0.00\\
2.22	0.00\\
2.26	-0.00\\
2.30	-0.00\\
2.34	-0.00\\
2.38	-0.00\\
2.41	-0.00\\
2.44	-0.00\\
2.47	-0.00\\
2.50	-0.00\\
2.52	-0.00\\
2.55	-0.00\\
2.57	-0.00\\
2.59	-0.00\\
2.61	-0.00\\
2.63	-0.00\\
2.65	-0.00\\
2.67	-0.00\\
2.68	-0.00\\
2.70	0.00\\
2.71	0.00\\
2.73	0.00\\
2.74	0.00\\
2.75	0.00\\
2.77	0.00\\
2.78	0.00\\
2.79	0.00\\
2.80	0.00\\
2.81	0.00\\
2.82	0.00\\
2.82	0.00\\
2.83	0.00\\
2.84	0.00\\
2.85	0.00\\
2.86	0.00\\
2.86	0.00\\
2.87	0.00\\
2.88	0.00\\
2.88	0.00\\
2.89	0.00\\
2.89	0.00\\
2.90	0.00\\
2.90	0.00\\
2.91	0.00\\
2.91	0.00\\
2.92	0.00\\
2.92	0.00\\
2.93	0.00\\
2.93	0.00\\
2.93	0.00\\
2.94	0.00\\
2.94	0.00\\
2.94	0.00\\
2.95	0.00\\
2.95	0.00\\
2.95	0.00\\
2.95	0.00\\
2.96	0.00\\
2.96	0.00\\
2.96	0.00\\
2.96	0.00\\
2.97	0.00\\
2.97	0.00\\
2.97	0.00\\
2.97	0.00\\
2.97	0.00\\
2.97	0.00\\
2.98	0.00\\
2.98	0.00\\
2.98	0.00\\
2.98	0.01\\
2.99	0.01\\
3.00	0.02\\
3.01	0.03\\
3.02	0.04\\
3.04	0.06\\
3.06	0.07\\
3.08	0.09\\
3.11	0.12\\
3.14	0.14\\
3.17	0.16\\
3.21	0.19\\
3.25	0.21\\
3.30	0.24\\
3.34	0.26\\
3.39	0.29\\
3.44	0.31\\
3.50	0.33\\
3.55	0.35\\
3.60	0.37\\
3.66	0.39\\
3.71	0.40\\
3.77	0.42\\
3.82	0.43\\
3.88	0.44\\
3.93	0.45\\
3.98	0.46\\
4.03	0.47\\
4.07	0.48\\
4.12	0.48\\
4.16	0.49\\
4.21	0.49\\
4.25	0.49\\
4.28	0.49\\
4.32	0.50\\
4.36	0.50\\
4.39	0.50\\
4.42	0.50\\
4.45	0.50\\
4.48	0.50\\
4.50	0.50\\
4.53	0.50\\
4.55	0.50\\
4.57	0.50\\
4.59	0.50\\
4.61	0.50\\
4.63	0.50\\
4.65	0.50\\
4.67	0.50\\
4.68	0.50\\
4.70	0.50\\
4.71	0.50\\
4.73	0.50\\
4.74	0.50\\
4.75	0.50\\
4.77	0.50\\
4.78	0.50\\
4.79	0.50\\
4.80	0.50\\
4.81	0.50\\
4.82	0.50\\
4.83	0.50\\
4.83	0.50\\
4.84	0.50\\
4.85	0.50\\
4.86	0.50\\
4.86	0.50\\
4.87	0.50\\
4.88	0.50\\
4.88	0.50\\
4.89	0.50\\
4.89	0.50\\
4.90	0.50\\
4.90	0.50\\
4.91	0.50\\
4.91	0.50\\
4.92	0.50\\
4.92	0.50\\
4.92	0.50\\
4.93	0.50\\
4.93	0.50\\
4.93	0.50\\
4.94	0.50\\
4.94	0.50\\
4.94	0.50\\
4.95	0.50\\
4.95	0.50\\
4.95	0.50\\
4.95	0.50\\
4.96	0.50\\
4.96	0.50\\
4.96	0.50\\
4.96	0.50\\
4.96	0.50\\
};
\addplot [color=mycolor1, line width=2.0pt, forget plot]
  table[row sep=crcr]{%
2.00	0.00\\
2.00	0.01\\
2.00	0.02\\
2.00	0.03\\
2.00	0.04\\
2.00	0.05\\
2.00	0.06\\
2.00	0.06\\
2.00	0.07\\
2.00	0.08\\
2.00	0.09\\
2.00	0.10\\
2.00	0.10\\
2.00	0.11\\
2.00	0.12\\
2.00	0.12\\
2.00	0.13\\
2.00	0.13\\
2.00	0.14\\
2.00	0.14\\
2.00	0.14\\
2.00	0.15\\
2.00	0.15\\
2.00	0.15\\
2.00	0.15\\
2.00	0.15\\
2.00	0.15\\
2.00	0.15\\
2.00	0.15\\
2.00	0.14\\
2.00	0.14\\
2.00	0.14\\
2.00	0.13\\
2.00	0.13\\
2.00	0.12\\
2.00	0.12\\
2.00	0.11\\
2.00	0.11\\
2.00	0.10\\
2.00	0.09\\
2.00	0.09\\
2.00	0.08\\
2.00	0.07\\
2.00	0.06\\
2.00	0.05\\
2.00	0.04\\
2.00	0.03\\
2.00	0.02\\
2.00	0.01\\
2.00	0.00\\
2.00	-0.00\\
2.00	-0.01\\
2.00	-0.02\\
2.00	-0.03\\
2.00	-0.04\\
2.00	-0.05\\
2.00	-0.06\\
2.00	-0.07\\
2.00	-0.08\\
2.00	-0.09\\
2.00	-0.09\\
2.00	-0.10\\
2.00	-0.11\\
2.00	-0.11\\
2.00	-0.12\\
2.00	-0.12\\
2.00	-0.13\\
2.00	-0.13\\
2.00	-0.14\\
2.00	-0.14\\
2.00	-0.14\\
2.00	-0.15\\
2.00	-0.15\\
2.00	-0.15\\
2.00	-0.15\\
2.00	-0.15\\
2.00	-0.15\\
2.00	-0.15\\
2.00	-0.15\\
2.00	-0.14\\
2.00	-0.14\\
2.00	-0.14\\
2.00	-0.13\\
2.00	-0.13\\
2.00	-0.12\\
2.00	-0.12\\
2.00	-0.11\\
2.00	-0.10\\
2.00	-0.10\\
2.00	-0.09\\
2.00	-0.08\\
2.00	-0.07\\
2.00	-0.06\\
2.00	-0.06\\
2.00	-0.05\\
2.00	-0.04\\
2.00	-0.03\\
2.00	-0.02\\
2.00	-0.01\\
2.00	-0.00\\
};
\addplot [color=mycolor2, line width=2.0pt, forget plot]
  table[row sep=crcr]{%
4.00	0.50\\
4.00	0.52\\
4.00	0.53\\
4.00	0.55\\
4.00	0.56\\
4.00	0.58\\
4.00	0.59\\
4.00	0.61\\
4.00	0.62\\
4.00	0.64\\
4.00	0.65\\
4.00	0.66\\
4.00	0.67\\
4.00	0.68\\
4.00	0.69\\
4.00	0.70\\
4.00	0.71\\
4.00	0.72\\
4.00	0.73\\
4.00	0.73\\
4.00	0.74\\
4.00	0.74\\
4.00	0.75\\
4.00	0.75\\
4.00	0.75\\
4.00	0.75\\
4.00	0.75\\
4.00	0.75\\
4.00	0.74\\
4.00	0.74\\
4.00	0.74\\
4.00	0.73\\
4.00	0.72\\
4.00	0.72\\
4.00	0.71\\
4.00	0.70\\
4.00	0.69\\
4.00	0.68\\
4.00	0.67\\
4.00	0.65\\
4.00	0.64\\
4.00	0.63\\
4.00	0.61\\
4.00	0.60\\
4.00	0.59\\
4.00	0.57\\
4.00	0.56\\
4.00	0.54\\
4.00	0.52\\
4.00	0.51\\
4.00	0.49\\
4.00	0.48\\
4.00	0.46\\
4.00	0.44\\
4.00	0.43\\
4.00	0.41\\
4.00	0.40\\
4.00	0.39\\
4.00	0.37\\
4.00	0.36\\
4.00	0.35\\
4.00	0.33\\
4.00	0.32\\
4.00	0.31\\
4.00	0.30\\
4.00	0.29\\
4.00	0.28\\
4.00	0.28\\
4.00	0.27\\
4.00	0.26\\
4.00	0.26\\
4.00	0.26\\
4.00	0.25\\
4.00	0.25\\
4.00	0.25\\
4.00	0.25\\
4.00	0.25\\
4.00	0.25\\
4.00	0.26\\
4.00	0.26\\
4.00	0.27\\
4.00	0.27\\
4.00	0.28\\
4.00	0.29\\
4.00	0.30\\
4.00	0.31\\
4.00	0.32\\
4.00	0.33\\
4.00	0.34\\
4.00	0.35\\
4.00	0.36\\
4.00	0.38\\
4.00	0.39\\
4.00	0.41\\
4.00	0.42\\
4.00	0.44\\
4.00	0.45\\
4.00	0.47\\
4.00	0.48\\
4.00	0.50\\
};
\addplot [color=white!30!black!, only marks, mark size=2.0pt, mark=*, mark options={solid, fill=white!30!black!}, forget plot]
  table[row sep=crcr]{%
0.00	0.50\\
};
\addplot [color=white!60!black!, mark size=1.5pt, mark=square*, mark options={solid, fill=white!60!black!}, forget plot]
  table[row sep=crcr]{%
3.00	0.00\\
};
\addplot [color=white!60!black!, mark size=1.5pt, mark=square*, mark options={solid, fill=white!60!black!}, forget plot]
  table[row sep=crcr]{%
5.00	0.50\\
};
\end{axis}

\end{tikzpicture}%

%% file: Tikz/trp.tex
%
%
\begin{tikzpicture}

\begin{axis}[%
width=0.96\fwidth,
height=0.31\fwidth,
at={(0\fwidth,1.1\fwidth)},
scale only axis,
xmin=0.00,
xmax=10.50,
ymin=-0.15,
ymax=0.15,
ylabel style={font=\color{white!15!black}},
ylabel={$\phi\text{ [rad]}$},
axis background/.style={fill=white},
xmajorgrids,
ymajorgrids
]
\addplot [color=black, line width=1.0pt, forget plot]
  table[row sep=crcr]{%
0.05	0.02\\
0.10	0.03\\
0.15	0.04\\
0.20	0.05\\
0.25	0.05\\
0.30	0.05\\
0.35	0.05\\
0.40	0.05\\
0.45	0.04\\
0.50	0.04\\
0.55	0.03\\
0.60	0.05\\
0.65	0.06\\
0.70	0.07\\
0.75	0.07\\
0.80	0.06\\
0.85	0.05\\
0.90	0.04\\
0.95	0.03\\
1.00	0.02\\
1.05	0.00\\
1.10	-0.01\\
1.15	-0.02\\
1.20	-0.02\\
1.25	-0.03\\
1.30	-0.04\\
1.35	-0.04\\
1.40	-0.04\\
1.45	-0.05\\
1.50	-0.05\\
1.55	-0.05\\
1.60	-0.05\\
1.65	-0.05\\
1.70	-0.04\\
1.75	-0.04\\
1.80	-0.04\\
1.85	-0.04\\
1.90	-0.03\\
1.95	-0.03\\
2.00	-0.03\\
2.05	-0.02\\
2.10	-0.02\\
2.15	-0.02\\
2.20	-0.02\\
2.25	-0.01\\
2.30	-0.01\\
2.35	-0.01\\
2.40	-0.01\\
2.45	-0.01\\
2.50	-0.01\\
2.55	-0.00\\
2.60	-0.00\\
2.65	-0.00\\
2.70	-0.00\\
2.75	-0.00\\
2.80	-0.00\\
2.85	-0.00\\
2.90	0.00\\
2.95	0.00\\
3.00	0.00\\
3.05	0.00\\
3.10	0.00\\
3.15	0.00\\
3.20	0.00\\
3.25	0.00\\
3.30	0.00\\
3.35	0.00\\
3.40	0.00\\
3.45	0.00\\
3.50	0.00\\
3.55	0.00\\
3.60	0.00\\
3.65	0.00\\
3.70	0.00\\
3.75	-0.00\\
3.80	-0.00\\
3.85	-0.00\\
3.90	-0.00\\
3.95	-0.00\\
4.00	-0.00\\
4.05	-0.00\\
4.10	-0.00\\
4.15	-0.00\\
4.20	-0.00\\
4.25	-0.00\\
4.30	-0.00\\
4.35	-0.00\\
4.40	-0.00\\
4.45	-0.00\\
4.50	0.00\\
4.55	0.00\\
4.60	0.00\\
4.65	0.00\\
4.70	0.00\\
4.75	0.00\\
4.80	0.00\\
4.85	0.00\\
4.90	0.00\\
4.95	0.00\\
5.00	0.00\\
5.05	0.00\\
5.10	0.00\\
5.15	0.00\\
5.20	0.00\\
5.25	0.00\\
5.30	0.00\\
5.35	0.00\\
5.40	0.00\\
5.45	0.00\\
5.50	0.00\\
5.55	0.00\\
5.60	0.00\\
5.65	-0.02\\
5.70	-0.04\\
5.75	-0.06\\
5.80	-0.07\\
5.85	-0.09\\
5.90	-0.10\\
5.95	-0.11\\
6.00	-0.11\\
6.05	-0.10\\
6.10	-0.09\\
6.15	-0.08\\
6.20	-0.06\\
6.25	-0.05\\
6.30	-0.03\\
6.35	-0.01\\
6.40	0.00\\
6.45	0.01\\
6.50	0.03\\
6.55	0.04\\
6.60	0.04\\
6.65	0.05\\
6.70	0.06\\
6.75	0.06\\
6.80	0.06\\
6.85	0.06\\
6.90	0.06\\
6.95	0.06\\
7.00	0.06\\
7.05	0.05\\
7.10	0.05\\
7.15	0.04\\
7.20	0.04\\
7.25	0.04\\
7.30	0.03\\
7.35	0.03\\
7.40	0.02\\
7.45	0.02\\
7.50	0.02\\
7.55	0.02\\
7.60	0.01\\
7.65	0.01\\
7.70	0.01\\
7.75	0.01\\
7.80	0.00\\
7.85	0.00\\
7.90	0.00\\
7.95	0.00\\
8.00	0.00\\
8.05	0.00\\
8.10	0.00\\
8.15	-0.00\\
8.20	-0.00\\
8.25	-0.00\\
8.30	-0.00\\
8.35	-0.00\\
8.40	-0.00\\
8.45	-0.00\\
8.50	-0.00\\
8.55	-0.00\\
8.60	-0.00\\
8.65	-0.00\\
8.70	-0.00\\
8.75	-0.00\\
8.80	-0.00\\
8.85	-0.00\\
8.90	-0.00\\
8.95	-0.00\\
9.00	-0.00\\
9.05	0.00\\
9.10	0.00\\
9.15	0.00\\
9.20	0.00\\
9.25	0.00\\
9.30	0.00\\
9.35	0.00\\
9.40	0.00\\
9.45	0.00\\
9.50	0.00\\
9.55	0.00\\
9.60	0.00\\
9.65	0.00\\
9.70	0.00\\
9.75	0.00\\
9.80	0.00\\
9.85	-0.00\\
9.90	-0.00\\
9.95	-0.00\\
10.00	-0.00\\
10.05	-0.00\\
10.10	-0.00\\
10.15	-0.00\\
10.20	-0.00\\
10.25	-0.00\\
10.30	-0.00\\
10.35	-0.00\\
10.40	-0.00\\
10.45	-0.00\\
10.50	-0.00\\
};
\end{axis}

\begin{axis}[%
width=0.96\fwidth,
height=0.31\fwidth,
at={(0\fwidth,0.7\fwidth)},
scale only axis,
xmin=0.00,
xmax=10.50,
ymin=-0.20,
ymax=0.20,
ylabel style={font=\color{white!15!black}},
ylabel={$\theta\text{ [rad]}$},
axis background/.style={fill=white},
xmajorgrids,
ymajorgrids
]
\addplot [color=black, line width=1.0pt, forget plot]
  table[row sep=crcr]{%
0.05	0.02\\
0.10	0.04\\
0.15	0.06\\
0.20	0.07\\
0.25	0.09\\
0.30	0.10\\
0.35	0.11\\
0.40	0.12\\
0.45	0.12\\
0.50	0.13\\
0.55	0.13\\
0.60	0.14\\
0.65	0.15\\
0.70	0.15\\
0.75	0.16\\
0.80	0.16\\
0.85	0.17\\
0.90	0.17\\
0.95	0.18\\
1.00	0.18\\
1.05	0.17\\
1.10	0.16\\
1.15	0.14\\
1.20	0.12\\
1.25	0.10\\
1.30	0.08\\
1.35	0.06\\
1.40	0.04\\
1.45	0.02\\
1.50	0.01\\
1.55	-0.01\\
1.60	-0.02\\
1.65	-0.03\\
1.70	-0.04\\
1.75	-0.05\\
1.80	-0.05\\
1.85	-0.07\\
1.90	-0.06\\
1.95	-0.08\\
2.00	-0.08\\
2.05	-0.09\\
2.10	-0.10\\
2.15	-0.10\\
2.20	-0.10\\
2.25	-0.10\\
2.30	-0.10\\
2.35	-0.09\\
2.40	-0.09\\
2.45	-0.09\\
2.50	-0.08\\
2.55	-0.08\\
2.60	-0.07\\
2.65	-0.07\\
2.70	-0.07\\
2.75	-0.06\\
2.80	-0.06\\
2.85	-0.05\\
2.90	-0.05\\
2.95	-0.05\\
3.00	-0.04\\
3.05	-0.04\\
3.10	-0.04\\
3.15	-0.04\\
3.20	-0.03\\
3.25	-0.03\\
3.30	-0.03\\
3.35	-0.03\\
3.40	-0.03\\
3.45	-0.02\\
3.50	-0.02\\
3.55	-0.02\\
3.60	-0.02\\
3.65	-0.02\\
3.70	-0.02\\
3.75	-0.02\\
3.80	-0.01\\
3.85	-0.01\\
3.90	-0.01\\
3.95	-0.01\\
4.00	-0.01\\
4.05	-0.01\\
4.10	-0.01\\
4.15	-0.01\\
4.20	-0.01\\
4.25	-0.01\\
4.30	-0.01\\
4.35	-0.01\\
4.40	-0.01\\
4.45	-0.01\\
4.50	-0.01\\
4.55	-0.01\\
4.60	-0.01\\
4.65	-0.01\\
4.70	-0.01\\
4.75	-0.01\\
4.80	-0.01\\
4.85	-0.01\\
4.90	-0.01\\
4.95	-0.01\\
5.00	-0.01\\
5.05	-0.00\\
5.10	-0.00\\
5.15	-0.00\\
5.20	-0.00\\
5.25	-0.00\\
5.30	-0.00\\
5.35	-0.00\\
5.40	-0.00\\
5.45	-0.00\\
5.50	-0.00\\
5.55	-0.00\\
5.60	-0.00\\
5.65	0.02\\
5.70	0.04\\
5.75	0.05\\
5.80	0.07\\
5.85	0.08\\
5.90	0.10\\
5.95	0.11\\
6.00	0.12\\
6.05	0.13\\
6.10	0.14\\
6.15	0.14\\
6.20	0.15\\
6.25	0.15\\
6.30	0.16\\
6.35	0.15\\
6.40	0.14\\
6.45	0.13\\
6.50	0.12\\
6.55	0.10\\
6.60	0.08\\
6.65	0.06\\
6.70	0.04\\
6.75	0.02\\
6.80	0.01\\
6.85	-0.01\\
6.90	-0.02\\
6.95	-0.03\\
7.00	-0.04\\
7.05	-0.04\\
7.10	-0.05\\
7.15	-0.06\\
7.20	-0.06\\
7.25	-0.07\\
7.30	-0.07\\
7.35	-0.07\\
7.40	-0.07\\
7.45	-0.07\\
7.50	-0.07\\
7.55	-0.07\\
7.60	-0.06\\
7.65	-0.06\\
7.70	-0.06\\
7.75	-0.06\\
7.80	-0.05\\
7.85	-0.05\\
7.90	-0.05\\
7.95	-0.04\\
8.00	-0.04\\
8.05	-0.04\\
8.10	-0.04\\
8.15	-0.03\\
8.20	-0.03\\
8.25	-0.03\\
8.30	-0.03\\
8.35	-0.03\\
8.40	-0.03\\
8.45	-0.02\\
8.50	-0.02\\
8.55	-0.02\\
8.60	-0.02\\
8.65	-0.02\\
8.70	-0.02\\
8.75	-0.02\\
8.80	-0.02\\
8.85	-0.02\\
8.90	-0.01\\
8.95	-0.01\\
9.00	-0.01\\
9.05	-0.01\\
9.10	-0.01\\
9.15	-0.01\\
9.20	-0.01\\
9.25	-0.01\\
9.30	-0.01\\
9.35	-0.01\\
9.40	-0.01\\
9.45	-0.01\\
9.50	-0.01\\
9.55	-0.01\\
9.60	-0.01\\
9.65	-0.01\\
9.70	-0.01\\
9.75	-0.01\\
9.80	-0.01\\
9.85	-0.01\\
9.90	-0.01\\
9.95	-0.01\\
10.00	-0.01\\
10.05	-0.00\\
10.10	-0.00\\
10.15	-0.00\\
10.20	-0.00\\
10.25	-0.00\\
10.30	-0.00\\
10.35	-0.00\\
10.40	-0.00\\
10.45	-0.00\\
10.50	-0.00\\
};
\end{axis}

\begin{axis}[%
width=0.96\fwidth,
height=0.31\fwidth,
at={(0\fwidth,0.3\fwidth)},
scale only axis,
xmin=0.00,
xmax=10.50,
xlabel style={font=\color{white!15!black}},
xlabel={Time [sec]},
ymin=9.0,
ymax=12.00,
ylabel style={font=\color{white!15!black}},
ylabel={$\text{T [m/s}^\text{2}\text{]}$},
axis background/.style={fill=white},
xmajorgrids,
ymajorgrids
]
\addplot [color=black, line width=1.0pt, forget plot]
  table[row sep=crcr]{%
0.05	9.81\\
0.10	9.84\\
0.15	9.86\\
0.20	9.88\\
0.25	9.89\\
0.30	9.90\\
0.35	9.90\\
0.40	9.90\\
0.45	9.91\\
0.50	9.89\\
0.55	9.90\\
0.60	9.94\\
0.65	9.94\\
0.70	9.93\\
0.75	9.93\\
0.80	9.92\\
0.85	9.91\\
0.90	9.91\\
0.95	9.90\\
1.00	9.90\\
1.05	9.89\\
1.10	9.88\\
1.15	9.87\\
1.20	9.87\\
1.25	9.86\\
1.30	9.86\\
1.35	9.86\\
1.40	9.85\\
1.45	9.86\\
1.50	9.85\\
1.55	9.85\\
1.60	9.85\\
1.65	9.85\\
1.70	9.85\\
1.75	9.86\\
1.80	9.84\\
1.85	9.86\\
1.90	9.84\\
1.95	9.85\\
2.00	9.85\\
2.05	9.85\\
2.10	9.85\\
2.15	9.85\\
2.20	9.85\\
2.25	9.85\\
2.30	9.84\\
2.35	9.84\\
2.40	9.84\\
2.45	9.84\\
2.50	9.84\\
2.55	9.84\\
2.60	9.84\\
2.65	9.83\\
2.70	9.83\\
2.75	9.83\\
2.80	9.83\\
2.85	9.83\\
2.90	9.83\\
2.95	9.83\\
3.00	9.83\\
3.05	9.83\\
3.10	9.82\\
3.15	9.82\\
3.20	9.82\\
3.25	9.82\\
3.30	9.82\\
3.35	9.82\\
3.40	9.82\\
3.45	9.82\\
3.50	9.82\\
3.55	9.82\\
3.60	9.82\\
3.65	9.82\\
3.70	9.81\\
3.75	9.81\\
3.80	9.81\\
3.85	9.81\\
3.90	9.81\\
3.95	9.81\\
4.00	9.81\\
4.05	9.81\\
4.10	9.81\\
4.15	9.81\\
4.20	9.81\\
4.25	9.81\\
4.30	9.81\\
4.35	9.81\\
4.40	9.81\\
4.45	9.81\\
4.50	9.81\\
4.55	9.81\\
4.60	9.81\\
4.65	9.81\\
4.70	9.81\\
4.75	9.81\\
4.80	9.81\\
4.85	9.81\\
4.90	9.81\\
4.95	9.81\\
5.00	9.81\\
5.05	9.81\\
5.10	9.81\\
5.15	9.81\\
5.20	9.81\\
5.25	9.81\\
5.30	9.81\\
5.35	9.81\\
5.40	9.81\\
5.45	9.81\\
5.50	9.81\\
5.55	9.81\\
5.60	9.81\\
5.65	11.59\\
5.70	11.38\\
5.75	11.18\\
5.80	10.99\\
5.85	10.82\\
5.90	10.66\\
5.95	10.51\\
6.00	10.38\\
6.05	10.25\\
6.10	10.15\\
6.15	10.04\\
6.20	9.96\\
6.25	9.88\\
6.30	9.82\\
6.35	9.76\\
6.40	9.71\\
6.45	9.67\\
6.50	9.64\\
6.55	9.60\\
6.60	9.58\\
6.65	9.57\\
6.70	9.56\\
6.75	9.55\\
6.80	9.55\\
6.85	9.54\\
6.90	9.55\\
6.95	9.55\\
7.00	9.56\\
7.05	9.56\\
7.10	9.57\\
7.15	9.58\\
7.20	9.59\\
7.25	9.60\\
7.30	9.62\\
7.35	9.63\\
7.40	9.64\\
7.45	9.65\\
7.50	9.66\\
7.55	9.67\\
7.60	9.69\\
7.65	9.70\\
7.70	9.70\\
7.75	9.71\\
7.80	9.72\\
7.85	9.73\\
7.90	9.74\\
7.95	9.75\\
8.00	9.76\\
8.05	9.76\\
8.10	9.77\\
8.15	9.78\\
8.20	9.78\\
8.25	9.78\\
8.30	9.79\\
8.35	9.79\\
8.40	9.80\\
8.45	9.80\\
8.50	9.81\\
8.55	9.81\\
8.60	9.81\\
8.65	9.81\\
8.70	9.81\\
8.75	9.81\\
8.80	9.81\\
8.85	9.81\\
8.90	9.81\\
8.95	9.81\\
9.00	9.82\\
9.05	9.82\\
9.10	9.82\\
9.15	9.82\\
9.20	9.82\\
9.25	9.82\\
9.30	9.82\\
9.35	9.82\\
9.40	9.82\\
9.45	9.82\\
9.50	9.82\\
9.55	9.82\\
9.60	9.82\\
9.65	9.82\\
9.70	9.82\\
9.75	9.82\\
9.80	9.82\\
9.85	9.82\\
9.90	9.82\\
9.95	9.82\\
10.00	9.81\\
10.05	9.81\\
10.10	9.81\\
10.15	9.81\\
10.20	9.81\\
10.25	9.81\\
10.30	9.81\\
10.35	9.81\\
10.40	9.81\\
10.45	9.81\\
10.50	9.81\\
};
\end{axis}
\end{tikzpicture}%

%% file: Tikz/config.tex
%
%
\begin{tikzpicture}

\begin{axis}[%
width=1.05\fwidth,
height=0.38\fwidth,
at={(0\fwidth,1.1\fwidth)},
scale only axis,
xmin=0.00,
xmax=10.50,
ymin=0.00,
ymax=5.00,
ylabel style={font=\color{white!15!black}},
ylabel={$\unit[d]{[m]}$},
axis background/.style={fill=white},
xmajorgrids,
ymajorgrids,
legend style={legend cell align=left, align=left, legend columns=2, draw=white!15!black}
]
\addplot [color=black, dotted,line width=1.0pt]
  table[row sep=crcr]{%
0.05	2.06\\
0.10	2.06\\
0.15	2.06\\
0.20	2.06\\
0.25	2.06\\
0.30	2.05\\
0.35	2.04\\
0.40	2.03\\
0.45	2.02\\
0.50	2.00\\
0.55	1.98\\
0.60	1.96\\
0.65	1.94\\
0.70	1.91\\
0.75	1.87\\
0.80	1.84\\
0.85	1.80\\
0.90	1.75\\
0.95	1.70\\
1.00	1.65\\
1.05	1.59\\
1.10	1.53\\
1.15	1.47\\
1.20	1.40\\
1.25	1.33\\
1.30	1.25\\
1.35	1.18\\
1.40	1.10\\
1.45	1.02\\
1.50	0.95\\
1.55	0.87\\
1.60	0.79\\
1.65	0.71\\
1.70	0.64\\
1.75	0.56\\
1.80	0.49\\
1.85	0.42\\
1.90	0.35\\
1.95	0.28\\
2.00	0.22\\
2.05	0.15\\
2.10	0.09\\
2.15	0.03\\
2.20	0.02\\
2.25	0.08\\
2.30	0.13\\
2.35	0.17\\
2.40	0.22\\
2.45	0.26\\
2.50	0.30\\
2.55	0.34\\
2.60	0.38\\
2.65	0.41\\
2.70	0.44\\
2.75	0.47\\
2.80	0.50\\
2.85	0.52\\
2.90	0.55\\
2.95	0.57\\
3.00	0.59\\
3.05	0.61\\
3.10	0.63\\
3.15	0.65\\
3.20	0.67\\
3.25	0.68\\
3.30	0.70\\
3.35	0.71\\
3.40	0.73\\
3.45	0.74\\
3.50	0.75\\
3.55	0.77\\
3.60	0.78\\
3.65	0.79\\
3.70	0.80\\
3.75	0.81\\
3.80	0.82\\
3.85	0.82\\
3.90	0.83\\
3.95	0.84\\
4.00	0.85\\
4.05	0.86\\
4.10	0.86\\
4.15	0.87\\
4.20	0.88\\
4.25	0.88\\
4.30	0.89\\
4.35	0.89\\
4.40	0.90\\
4.45	0.90\\
4.50	0.91\\
4.55	0.91\\
4.60	0.92\\
4.65	0.92\\
4.70	0.93\\
4.75	0.93\\
4.80	0.93\\
4.85	0.94\\
4.90	0.94\\
4.95	0.94\\
5.00	0.95\\
5.05	0.95\\
5.10	0.95\\
5.15	0.95\\
5.20	0.96\\
5.25	0.96\\
5.30	0.96\\
5.35	0.96\\
5.40	0.97\\
5.45	0.97\\
5.50	0.97\\
5.55	0.97\\
5.60	0.97\\
5.65	0.97\\
5.70	0.98\\
5.75	0.98\\
5.80	0.98\\
5.85	0.99\\
5.90	0.99\\
5.95	1.00\\
6.00	1.02\\
6.05	1.03\\
6.10	1.05\\
6.15	1.07\\
6.20	1.10\\
6.25	1.13\\
6.30	1.17\\
6.35	1.21\\
6.40	1.25\\
6.45	1.30\\
6.50	1.35\\
6.55	1.40\\
6.60	1.46\\
6.65	1.52\\
6.70	1.57\\
6.75	1.63\\
6.80	1.69\\
6.85	1.75\\
6.90	1.81\\
6.95	1.87\\
7.00	1.92\\
7.05	1.98\\
7.10	2.03\\
7.15	2.08\\
7.20	2.13\\
7.25	2.18\\
7.30	2.23\\
7.35	2.27\\
7.40	2.31\\
7.45	2.35\\
7.50	2.39\\
7.55	2.43\\
7.60	2.46\\
7.65	2.49\\
7.70	2.52\\
7.75	2.55\\
7.80	2.58\\
7.85	2.60\\
7.90	2.63\\
7.95	2.65\\
8.00	2.67\\
8.05	2.69\\
8.10	2.71\\
8.15	2.73\\
8.20	2.75\\
8.25	2.76\\
8.30	2.78\\
8.35	2.79\\
8.40	2.81\\
8.45	2.82\\
8.50	2.83\\
8.55	2.84\\
8.60	2.86\\
8.65	2.87\\
8.70	2.88\\
8.75	2.89\\
8.80	2.90\\
8.85	2.90\\
8.90	2.91\\
8.95	2.92\\
9.00	2.93\\
9.05	2.94\\
9.10	2.94\\
9.15	2.95\\
9.20	2.96\\
9.25	2.96\\
9.30	2.97\\
9.35	2.97\\
9.40	2.98\\
9.45	2.98\\
9.50	2.99\\
9.55	2.99\\
9.60	3.00\\
9.65	3.00\\
9.70	3.00\\
9.75	3.01\\
9.80	3.01\\
9.85	3.02\\
9.90	3.02\\
9.95	3.02\\
10.00	3.02\\
10.05	3.03\\
10.10	3.03\\
10.15	3.03\\
10.20	3.03\\
10.25	3.04\\
10.30	3.04\\
10.35	3.04\\
10.40	3.04\\
10.45	3.04\\
10.50	3.05\\
};
\addlegendentry{$\text{d}_\text{1}$}

\addplot [color=black, dashed, line width=1.0pt]
  table[row sep=crcr]{%
0.05	4.03\\
0.10	4.03\\
0.15	4.03\\
0.20	4.03\\
0.25	4.03\\
0.30	4.02\\
0.35	4.02\\
0.40	4.01\\
0.45	3.99\\
0.50	3.98\\
0.55	3.96\\
0.60	3.94\\
0.65	3.92\\
0.70	3.89\\
0.75	3.86\\
0.80	3.83\\
0.85	3.79\\
0.90	3.75\\
0.95	3.70\\
1.00	3.66\\
1.05	3.60\\
1.10	3.55\\
1.15	3.49\\
1.20	3.43\\
1.25	3.36\\
1.30	3.29\\
1.35	3.22\\
1.40	3.15\\
1.45	3.08\\
1.50	3.00\\
1.55	2.93\\
1.60	2.86\\
1.65	2.79\\
1.70	2.72\\
1.75	2.65\\
1.80	2.58\\
1.85	2.51\\
1.90	2.44\\
1.95	2.38\\
2.00	2.32\\
2.05	2.26\\
2.10	2.20\\
2.15	2.15\\
2.20	2.10\\
2.25	2.05\\
2.30	2.00\\
2.35	1.96\\
2.40	1.92\\
2.45	1.88\\
2.50	1.84\\
2.55	1.80\\
2.60	1.77\\
2.65	1.74\\
2.70	1.71\\
2.75	1.69\\
2.80	1.66\\
2.85	1.64\\
2.90	1.62\\
2.95	1.59\\
3.00	1.57\\
3.05	1.56\\
3.10	1.54\\
3.15	1.52\\
3.20	1.51\\
3.25	1.49\\
3.30	1.48\\
3.35	1.47\\
3.40	1.46\\
3.45	1.44\\
3.50	1.43\\
3.55	1.42\\
3.60	1.41\\
3.65	1.40\\
3.70	1.40\\
3.75	1.39\\
3.80	1.38\\
3.85	1.37\\
3.90	1.36\\
3.95	1.36\\
4.00	1.35\\
4.05	1.35\\
4.10	1.34\\
4.15	1.33\\
4.20	1.33\\
4.25	1.32\\
4.30	1.32\\
4.35	1.31\\
4.40	1.31\\
4.45	1.31\\
4.50	1.30\\
4.55	1.30\\
4.60	1.29\\
4.65	1.29\\
4.70	1.29\\
4.75	1.28\\
4.80	1.28\\
4.85	1.28\\
4.90	1.27\\
4.95	1.27\\
5.00	1.27\\
5.05	1.27\\
5.10	1.26\\
5.15	1.26\\
5.20	1.26\\
5.25	1.26\\
5.30	1.26\\
5.35	1.25\\
5.40	1.25\\
5.45	1.25\\
5.50	1.25\\
5.55	1.25\\
5.60	1.25\\
5.65	1.25\\
5.70	1.24\\
5.75	1.24\\
5.80	1.23\\
5.85	1.22\\
5.90	1.21\\
5.95	1.19\\
6.00	1.17\\
6.05	1.15\\
6.10	1.12\\
6.15	1.09\\
6.20	1.05\\
6.25	1.01\\
6.30	0.97\\
6.35	0.93\\
6.40	0.88\\
6.45	0.83\\
6.50	0.78\\
6.55	0.72\\
6.60	0.67\\
6.65	0.61\\
6.70	0.55\\
6.75	0.49\\
6.80	0.43\\
6.85	0.37\\
6.90	0.31\\
6.95	0.26\\
7.00	0.20\\
7.05	0.15\\
7.10	0.10\\
7.15	0.06\\
7.20	0.05\\
7.25	0.08\\
7.30	0.12\\
7.35	0.17\\
7.40	0.21\\
7.45	0.25\\
7.50	0.28\\
7.55	0.32\\
7.60	0.36\\
7.65	0.39\\
7.70	0.42\\
7.75	0.45\\
7.80	0.48\\
7.85	0.50\\
7.90	0.53\\
7.95	0.55\\
8.00	0.57\\
8.05	0.59\\
8.10	0.61\\
8.15	0.63\\
8.20	0.65\\
8.25	0.67\\
8.30	0.68\\
8.35	0.70\\
8.40	0.71\\
8.45	0.73\\
8.50	0.74\\
8.55	0.75\\
8.60	0.77\\
8.65	0.78\\
8.70	0.79\\
8.75	0.80\\
8.80	0.81\\
8.85	0.82\\
8.90	0.83\\
8.95	0.83\\
9.00	0.84\\
9.05	0.85\\
9.10	0.86\\
9.15	0.86\\
9.20	0.87\\
9.25	0.88\\
9.30	0.88\\
9.35	0.89\\
9.40	0.89\\
9.45	0.90\\
9.50	0.90\\
9.55	0.91\\
9.60	0.91\\
9.65	0.92\\
9.70	0.92\\
9.75	0.92\\
9.80	0.93\\
9.85	0.93\\
9.90	0.93\\
9.95	0.94\\
10.00	0.94\\
10.05	0.94\\
10.10	0.95\\
10.15	0.95\\
10.20	0.95\\
10.25	0.95\\
10.30	0.96\\
10.35	0.96\\
10.40	0.96\\
10.45	0.96\\
10.50	0.96\\
};
\addlegendentry{$\text{d}_\text{2}$}

\end{axis}

\begin{axis}[%
width=1.05\fwidth,
height=0.38\fwidth,
at={(0\fwidth,0.57\fwidth)},
scale only axis,
xmin=0.00,
xmax=10.50,
xlabel style={font=\color{white!15!black}},
xlabel={Time [sec]},
ymin=0.25,
ymax=0.70,
ylabel style={font=\color{white!15!black}},
ylabel={$\unit[r]{[m]}$},
axis background/.style={fill=white},
xmajorgrids,
ymajorgrids,
legend style={at={(0.93,1.23)},legend cell align=left, align=left, legend columns=4, draw=white!15!black}
]
\addplot [color=black, line width=3.0pt]
  table[row sep=crcr]{%
0.05	0.67\\
0.10	0.67\\
0.15	0.67\\
0.20	0.67\\
0.25	0.67\\
0.30	0.67\\
0.35	0.67\\
0.40	0.67\\
0.45	0.67\\
0.50	0.67\\
0.55	0.67\\
0.60	0.67\\
0.65	0.67\\
0.70	0.67\\
0.75	0.67\\
0.80	0.67\\
0.85	0.67\\
0.90	0.66\\
0.95	0.66\\
1.00	0.66\\
1.05	0.66\\
1.10	0.65\\
1.15	0.64\\
1.20	0.64\\
1.25	0.63\\
1.30	0.62\\
1.35	0.61\\
1.40	0.59\\
1.45	0.58\\
1.50	0.55\\
1.55	0.54\\
1.60	0.49\\
1.65	0.48\\
1.70	0.41\\
1.75	0.40\\
1.80	0.35\\
1.85	0.34\\
1.90	0.30\\
1.95	0.30\\
2.00	0.30\\
2.05	0.30\\
2.10	0.30\\
2.15	0.30\\
2.20	0.30\\
2.25	0.30\\
2.30	0.30\\
2.35	0.30\\
2.40	0.30\\
2.45	0.66\\
2.50	0.66\\
2.55	0.66\\
2.60	0.66\\
2.65	0.66\\
2.70	0.66\\
2.75	0.67\\
2.80	0.67\\
2.85	0.67\\
2.90	0.67\\
2.95	0.67\\
3.00	0.67\\
3.05	0.67\\
3.10	0.67\\
3.15	0.67\\
3.20	0.67\\
3.25	0.67\\
3.30	0.67\\
3.35	0.67\\
3.40	0.67\\
3.45	0.67\\
3.50	0.67\\
3.55	0.67\\
3.60	0.67\\
3.65	0.67\\
3.70	0.67\\
3.75	0.67\\
3.80	0.67\\
3.85	0.67\\
3.90	0.67\\
3.95	0.67\\
4.00	0.67\\
4.05	0.67\\
4.10	0.67\\
4.15	0.67\\
4.20	0.67\\
4.25	0.67\\
4.30	0.67\\
4.35	0.67\\
4.40	0.67\\
4.45	0.67\\
4.50	0.67\\
4.55	0.67\\
4.60	0.67\\
4.65	0.67\\
4.70	0.67\\
4.75	0.67\\
4.80	0.67\\
4.85	0.67\\
4.90	0.67\\
4.95	0.67\\
5.00	0.67\\
5.05	0.67\\
5.10	0.67\\
5.15	0.67\\
5.20	0.67\\
5.25	0.67\\
5.30	0.67\\
5.35	0.67\\
5.40	0.67\\
5.45	0.67\\
5.50	0.67\\
5.55	0.67\\
5.60	0.67\\
5.65	0.67\\
5.70	0.67\\
5.75	0.67\\
5.80	0.67\\
5.85	0.67\\
5.90	0.67\\
5.95	0.67\\
6.00	0.67\\
6.05	0.67\\
6.10	0.66\\
6.15	0.66\\
6.20	0.66\\
6.25	0.65\\
6.30	0.66\\
6.35	0.65\\
6.40	0.64\\
6.45	0.62\\
6.50	0.62\\
6.55	0.62\\
6.60	0.59\\
6.65	0.59\\
6.70	0.56\\
6.75	0.55\\
6.80	0.52\\
6.85	0.52\\
6.90	0.50\\
6.95	0.50\\
7.00	0.50\\
7.05	0.50\\
7.10	0.50\\
7.15	0.50\\
7.20	0.50\\
7.25	0.50\\
7.30	0.50\\
7.35	0.50\\
7.40	0.50\\
7.45	0.51\\
7.50	0.67\\
7.55	0.67\\
7.60	0.67\\
7.65	0.67\\
7.70	0.67\\
7.75	0.67\\
7.80	0.67\\
7.85	0.67\\
7.90	0.67\\
7.95	0.67\\
8.00	0.67\\
8.05	0.67\\
8.10	0.67\\
8.15	0.67\\
8.20	0.67\\
8.25	0.67\\
8.30	0.67\\
8.35	0.67\\
8.40	0.67\\
8.45	0.67\\
8.50	0.67\\
8.55	0.67\\
8.60	0.67\\
8.65	0.67\\
8.70	0.67\\
8.75	0.67\\
8.80	0.67\\
8.85	0.67\\
8.90	0.67\\
8.95	0.67\\
9.00	0.67\\
9.05	0.67\\
9.10	0.67\\
9.15	0.67\\
9.20	0.67\\
9.25	0.67\\
9.30	0.67\\
9.35	0.67\\
9.40	0.67\\
9.45	0.67\\
9.50	0.67\\
9.55	0.67\\
9.60	0.67\\
9.65	0.67\\
9.70	0.67\\
9.75	0.67\\
9.80	0.67\\
9.85	0.67\\
9.90	0.67\\
9.95	0.67\\
10.00	0.67\\
10.05	0.67\\
10.10	0.67\\
10.15	0.67\\
10.20	0.67\\
10.25	0.67\\
10.30	0.67\\
10.35	0.67\\
10.40	0.67\\
10.45	0.67\\
10.50	0.67\\
};
\addlegendentry{$r_{\text{front}}$}

\addplot [color=white!80!black!, line width=1.0pt]
  table[row sep=crcr]{%
0.05	0.67\\
0.10	0.67\\
0.15	0.67\\
0.20	0.67\\
0.25	0.67\\
0.30	0.67\\
0.35	0.67\\
0.40	0.67\\
0.45	0.67\\
0.50	0.67\\
0.55	0.67\\
0.60	0.67\\
0.65	0.67\\
0.70	0.67\\
0.75	0.67\\
0.80	0.67\\
0.85	0.67\\
0.90	0.66\\
0.95	0.66\\
1.00	0.66\\
1.05	0.66\\
1.10	0.65\\
1.15	0.64\\
1.20	0.64\\
1.25	0.63\\
1.30	0.62\\
1.35	0.61\\
1.40	0.59\\
1.45	0.58\\
1.50	0.55\\
1.55	0.54\\
1.60	0.49\\
1.65	0.48\\
1.70	0.41\\
1.75	0.40\\
1.80	0.35\\
1.85	0.34\\
1.90	0.30\\
1.95	0.30\\
2.00	0.30\\
2.05	0.30\\
2.10	0.30\\
2.15	0.30\\
2.20	0.30\\
2.25	0.30\\
2.30	0.30\\
2.35	0.30\\
2.40	0.30\\
2.45	0.66\\
2.50	0.66\\
2.55	0.66\\
2.60	0.66\\
2.65	0.66\\
2.70	0.66\\
2.75	0.67\\
2.80	0.67\\
2.85	0.67\\
2.90	0.67\\
2.95	0.67\\
3.00	0.67\\
3.05	0.67\\
3.10	0.67\\
3.15	0.67\\
3.20	0.67\\
3.25	0.67\\
3.30	0.67\\
3.35	0.67\\
3.40	0.67\\
3.45	0.67\\
3.50	0.67\\
3.55	0.67\\
3.60	0.67\\
3.65	0.67\\
3.70	0.67\\
3.75	0.67\\
3.80	0.67\\
3.85	0.67\\
3.90	0.67\\
3.95	0.67\\
4.00	0.67\\
4.05	0.67\\
4.10	0.67\\
4.15	0.67\\
4.20	0.67\\
4.25	0.67\\
4.30	0.67\\
4.35	0.67\\
4.40	0.67\\
4.45	0.67\\
4.50	0.67\\
4.55	0.67\\
4.60	0.67\\
4.65	0.67\\
4.70	0.67\\
4.75	0.67\\
4.80	0.67\\
4.85	0.67\\
4.90	0.67\\
4.95	0.67\\
5.00	0.67\\
5.05	0.67\\
5.10	0.67\\
5.15	0.67\\
5.20	0.67\\
5.25	0.67\\
5.30	0.67\\
5.35	0.67\\
5.40	0.67\\
5.45	0.67\\
5.50	0.67\\
5.55	0.67\\
5.60	0.67\\
5.65	0.67\\
5.70	0.67\\
5.75	0.67\\
5.80	0.67\\
5.85	0.67\\
5.90	0.67\\
5.95	0.67\\
6.00	0.67\\
6.05	0.67\\
6.10	0.66\\
6.15	0.66\\
6.20	0.66\\
6.25	0.65\\
6.30	0.66\\
6.35	0.65\\
6.40	0.64\\
6.45	0.62\\
6.50	0.62\\
6.55	0.62\\
6.60	0.59\\
6.65	0.59\\
6.70	0.56\\
6.75	0.55\\
6.80	0.52\\
6.85	0.52\\
6.90	0.50\\
6.95	0.50\\
7.00	0.50\\
7.05	0.50\\
7.10	0.50\\
7.15	0.50\\
7.20	0.50\\
7.25	0.50\\
7.30	0.50\\
7.35	0.50\\
7.40	0.50\\
7.45	0.51\\
7.50	0.67\\
7.55	0.67\\
7.60	0.67\\
7.65	0.67\\
7.70	0.67\\
7.75	0.67\\
7.80	0.67\\
7.85	0.67\\
7.90	0.67\\
7.95	0.67\\
8.00	0.67\\
8.05	0.67\\
8.10	0.67\\
8.15	0.67\\
8.20	0.67\\
8.25	0.67\\
8.30	0.67\\
8.35	0.67\\
8.40	0.67\\
8.45	0.67\\
8.50	0.67\\
8.55	0.67\\
8.60	0.67\\
8.65	0.67\\
8.70	0.67\\
8.75	0.67\\
8.80	0.67\\
8.85	0.67\\
8.90	0.67\\
8.95	0.67\\
9.00	0.67\\
9.05	0.67\\
9.10	0.67\\
9.15	0.67\\
9.20	0.67\\
9.25	0.67\\
9.30	0.67\\
9.35	0.67\\
9.40	0.67\\
9.45	0.67\\
9.50	0.67\\
9.55	0.67\\
9.60	0.67\\
9.65	0.67\\
9.70	0.67\\
9.75	0.67\\
9.80	0.67\\
9.85	0.67\\
9.90	0.67\\
9.95	0.67\\
10.00	0.67\\
10.05	0.67\\
10.10	0.67\\
10.15	0.67\\
10.20	0.67\\
10.25	0.67\\
10.30	0.67\\
10.35	0.67\\
10.40	0.67\\
10.45	0.67\\
10.50	0.67\\
};
\addlegendentry{$r_{\text{rear}}$}

\addplot [color=black,dotted, line width=1.0pt]
  table[row sep=crcr]{%
0.05	0.30\\
0.10	0.30\\
0.15	0.30\\
0.20	0.30\\
0.25	0.30\\
0.30	0.30\\
0.35	0.30\\
0.40	0.30\\
0.45	0.30\\
0.50	0.30\\
0.55	0.30\\
0.60	0.30\\
0.65	0.30\\
0.70	0.30\\
0.75	0.30\\
0.80	0.30\\
0.85	0.30\\
0.90	0.30\\
0.95	0.30\\
1.00	0.30\\
1.05	0.30\\
1.10	0.30\\
1.15	0.30\\
1.20	0.30\\
1.25	0.30\\
1.30	0.30\\
1.35	0.30\\
1.40	0.30\\
1.45	0.30\\
1.50	0.30\\
1.55	0.30\\
1.60	0.30\\
1.65	0.30\\
1.70	0.30\\
1.75	0.30\\
1.80	0.30\\
1.85	0.30\\
1.90	0.30\\
1.95	0.30\\
2.00	0.30\\
2.05	0.30\\
2.10	0.30\\
2.15	0.30\\
2.20	0.30\\
2.25	0.30\\
2.30	0.30\\
2.35	0.30\\
2.40	0.30\\
2.45	0.30\\
2.50	0.30\\
2.55	0.30\\
2.60	0.30\\
2.65	0.30\\
2.70	0.30\\
2.75	0.30\\
2.80	0.30\\
2.85	0.30\\
2.90	0.30\\
2.95	0.30\\
3.00	0.30\\
3.05	0.30\\
3.10	0.30\\
3.15	0.30\\
3.20	0.30\\
3.25	0.30\\
3.30	0.30\\
3.35	0.30\\
3.40	0.30\\
3.45	0.30\\
3.50	0.30\\
3.55	0.30\\
3.60	0.30\\
3.65	0.30\\
3.70	0.30\\
3.75	0.30\\
3.80	0.30\\
3.85	0.30\\
3.90	0.30\\
3.95	0.30\\
4.00	0.30\\
4.05	0.30\\
4.10	0.30\\
4.15	0.30\\
4.20	0.30\\
4.25	0.30\\
4.30	0.30\\
4.35	0.30\\
4.40	0.30\\
4.45	0.30\\
4.50	0.30\\
4.55	0.30\\
4.60	0.30\\
4.65	0.30\\
4.70	0.30\\
4.75	0.30\\
4.80	0.30\\
4.85	0.30\\
4.90	0.30\\
4.95	0.30\\
5.00	0.30\\
5.05	0.30\\
5.10	0.30\\
5.15	0.30\\
5.20	0.30\\
5.25	0.30\\
5.30	0.30\\
5.35	0.30\\
5.40	0.30\\
5.45	0.30\\
5.50	0.30\\
5.55	0.30\\
5.60	0.30\\
5.65	0.30\\
5.70	0.30\\
5.75	0.30\\
5.80	0.30\\
5.85	0.30\\
5.90	0.30\\
5.95	0.30\\
6.00	0.30\\
6.05	0.30\\
6.10	0.30\\
6.15	0.30\\
6.20	0.30\\
6.25	0.30\\
6.30	0.30\\
6.35	0.30\\
6.40	0.30\\
6.45	0.30\\
6.50	0.30\\
6.55	0.30\\
6.60	0.30\\
6.65	0.30\\
6.70	0.30\\
6.75	0.30\\
6.80	0.30\\
6.85	0.30\\
6.90	0.30\\
6.95	0.30\\
7.00	0.30\\
7.05	0.30\\
7.10	0.30\\
7.15	0.30\\
7.20	0.30\\
7.25	0.30\\
7.30	0.30\\
7.35	0.30\\
7.40	0.30\\
7.45	0.30\\
7.50	0.30\\
7.55	0.30\\
7.60	0.30\\
7.65	0.30\\
7.70	0.30\\
7.75	0.30\\
7.80	0.30\\
7.85	0.30\\
7.90	0.30\\
7.95	0.30\\
8.00	0.30\\
8.05	0.30\\
8.10	0.30\\
8.15	0.30\\
8.20	0.30\\
8.25	0.30\\
8.30	0.30\\
8.35	0.30\\
8.40	0.30\\
8.45	0.30\\
8.50	0.30\\
8.55	0.30\\
8.60	0.30\\
8.65	0.30\\
8.70	0.30\\
8.75	0.30\\
8.80	0.30\\
8.85	0.30\\
8.90	0.30\\
8.95	0.30\\
9.00	0.30\\
9.05	0.30\\
9.10	0.30\\
9.15	0.30\\
9.20	0.30\\
9.25	0.30\\
9.30	0.30\\
9.35	0.30\\
9.40	0.30\\
9.45	0.30\\
9.50	0.30\\
9.55	0.30\\
9.60	0.30\\
9.65	0.30\\
9.70	0.30\\
9.75	0.30\\
9.80	0.30\\
9.85	0.30\\
9.90	0.30\\
9.95	0.30\\
10.00	0.30\\
10.05	0.30\\
10.10	0.30\\
10.15	0.30\\
10.20	0.30\\
10.25	0.30\\
10.30	0.30\\
10.35	0.30\\
10.40	0.30\\
10.45	0.30\\
10.50	0.30\\
};
\addlegendentry{$2r_{obs1}$}

\addplot [color=black, dashed, line width=1.0pt]
  table[row sep=crcr]{%
0.05	0.50\\
0.10	0.50\\
0.15	0.50\\
0.20	0.50\\
0.25	0.50\\
0.30	0.50\\
0.35	0.50\\
0.40	0.50\\
0.45	0.50\\
0.50	0.50\\
0.55	0.50\\
0.60	0.50\\
0.65	0.50\\
0.70	0.50\\
0.75	0.50\\
0.80	0.50\\
0.85	0.50\\
0.90	0.50\\
0.95	0.50\\
1.00	0.50\\
1.05	0.50\\
1.10	0.50\\
1.15	0.50\\
1.20	0.50\\
1.25	0.50\\
1.30	0.50\\
1.35	0.50\\
1.40	0.50\\
1.45	0.50\\
1.50	0.50\\
1.55	0.50\\
1.60	0.50\\
1.65	0.50\\
1.70	0.50\\
1.75	0.50\\
1.80	0.50\\
1.85	0.50\\
1.90	0.50\\
1.95	0.50\\
2.00	0.50\\
2.05	0.50\\
2.10	0.50\\
2.15	0.50\\
2.20	0.50\\
2.25	0.50\\
2.30	0.50\\
2.35	0.50\\
2.40	0.50\\
2.45	0.50\\
2.50	0.50\\
2.55	0.50\\
2.60	0.50\\
2.65	0.50\\
2.70	0.50\\
2.75	0.50\\
2.80	0.50\\
2.85	0.50\\
2.90	0.50\\
2.95	0.50\\
3.00	0.50\\
3.05	0.50\\
3.10	0.50\\
3.15	0.50\\
3.20	0.50\\
3.25	0.50\\
3.30	0.50\\
3.35	0.50\\
3.40	0.50\\
3.45	0.50\\
3.50	0.50\\
3.55	0.50\\
3.60	0.50\\
3.65	0.50\\
3.70	0.50\\
3.75	0.50\\
3.80	0.50\\
3.85	0.50\\
3.90	0.50\\
3.95	0.50\\
4.00	0.50\\
4.05	0.50\\
4.10	0.50\\
4.15	0.50\\
4.20	0.50\\
4.25	0.50\\
4.30	0.50\\
4.35	0.50\\
4.40	0.50\\
4.45	0.50\\
4.50	0.50\\
4.55	0.50\\
4.60	0.50\\
4.65	0.50\\
4.70	0.50\\
4.75	0.50\\
4.80	0.50\\
4.85	0.50\\
4.90	0.50\\
4.95	0.50\\
5.00	0.50\\
5.05	0.50\\
5.10	0.50\\
5.15	0.50\\
5.20	0.50\\
5.25	0.50\\
5.30	0.50\\
5.35	0.50\\
5.40	0.50\\
5.45	0.50\\
5.50	0.50\\
5.55	0.50\\
5.60	0.50\\
5.65	0.50\\
5.70	0.50\\
5.75	0.50\\
5.80	0.50\\
5.85	0.50\\
5.90	0.50\\
5.95	0.50\\
6.00	0.50\\
6.05	0.50\\
6.10	0.50\\
6.15	0.50\\
6.20	0.50\\
6.25	0.50\\
6.30	0.50\\
6.35	0.50\\
6.40	0.50\\
6.45	0.50\\
6.50	0.50\\
6.55	0.50\\
6.60	0.50\\
6.65	0.50\\
6.70	0.50\\
6.75	0.50\\
6.80	0.50\\
6.85	0.50\\
6.90	0.50\\
6.95	0.50\\
7.00	0.50\\
7.05	0.50\\
7.10	0.50\\
7.15	0.50\\
7.20	0.50\\
7.25	0.50\\
7.30	0.50\\
7.35	0.50\\
7.40	0.50\\
7.45	0.50\\
7.50	0.50\\
7.55	0.50\\
7.60	0.50\\
7.65	0.50\\
7.70	0.50\\
7.75	0.50\\
7.80	0.50\\
7.85	0.50\\
7.90	0.50\\
7.95	0.50\\
8.00	0.50\\
8.05	0.50\\
8.10	0.50\\
8.15	0.50\\
8.20	0.50\\
8.25	0.50\\
8.30	0.50\\
8.35	0.50\\
8.40	0.50\\
8.45	0.50\\
8.50	0.50\\
8.55	0.50\\
8.60	0.50\\
8.65	0.50\\
8.70	0.50\\
8.75	0.50\\
8.80	0.50\\
8.85	0.50\\
8.90	0.50\\
8.95	0.50\\
9.00	0.50\\
9.05	0.50\\
9.10	0.50\\
9.15	0.50\\
9.20	0.50\\
9.25	0.50\\
9.30	0.50\\
9.35	0.50\\
9.40	0.50\\
9.45	0.50\\
9.50	0.50\\
9.55	0.50\\
9.60	0.50\\
9.65	0.50\\
9.70	0.50\\
9.75	0.50\\
9.80	0.50\\
9.85	0.50\\
9.90	0.50\\
9.95	0.50\\
10.00	0.50\\
10.05	0.50\\
10.10	0.50\\
10.15	0.50\\
10.20	0.50\\
10.25	0.50\\
10.30	0.50\\
10.35	0.50\\
10.40	0.50\\
10.45	0.50\\
10.50	0.50\\
};
\addlegendentry{$2r_{obs2}$}
\end{axis}

\end{tikzpicture}%

%% file: Tikz/3Dplot_not_pass2.tex
%
%
\definecolor{mycolor1}{rgb}{0.49400,0.18400,0.55600}%
\definecolor{mycolor2}{rgb}{0.30100,0.74500,0.93300}%
\definecolor{mycolor3}{rgb}{0.00000,0.44700,0.74100}%
\definecolor{mycolor4}{rgb}{0.85000,0.32500,0.09800}%

\begin{tikzpicture}

\begin{axis}[%
width=0.951\fwidth,
height=0.77\fwidth,
at={(0\fwidth,1.1\fwidth)},
scale only axis,
xmin=1.00,
xmax=3.00,
tick align=outside,
xlabel style={font=\color{white!15!black}},
xlabel={$x$ [m]},
ymin=-0.50,
ymax=0.50,
ylabel style={font=\color{white!15!black}},
ylabel={$y$ [m]},
zmin=0.70,
zmax=1.30,
zlabel style={font=\color{white!15!black}},
zlabel={$z$ [m]},
view={-63.60}{15.71},
axis background/.style={fill=white},
xmajorgrids,
ymajorgrids,
zmajorgrids,
legend style={at={(0.90,1.1)},legend cell align=left, align=left, legend columns=3, draw=white!15!black}
]
\addplot3 [color=mycolor4, line width=2.0pt]
 table[row sep=crcr] {%
2.00	0.00	1.10\\
2.00	0.01	1.10\\
2.00	0.01	1.10\\
2.00	0.02	1.10\\
2.00	0.03	1.10\\
2.00	0.03	1.10\\
2.00	0.04	1.09\\
2.00	0.04	1.09\\
2.00	0.05	1.09\\
2.00	0.05	1.08\\
2.00	0.06	1.08\\
2.00	0.06	1.08\\
2.00	0.07	1.07\\
2.00	0.07	1.07\\
2.00	0.08	1.06\\
2.00	0.08	1.06\\
2.00	0.08	1.05\\
2.00	0.09	1.05\\
2.00	0.09	1.04\\
2.00	0.09	1.04\\
2.00	0.10	1.03\\
2.00	0.10	1.02\\
2.00	0.10	1.02\\
2.00	0.10	1.01\\
2.00	0.10	1.00\\
2.00	0.10	1.00\\
2.00	0.10	0.99\\
2.00	0.10	0.99\\
2.00	0.10	0.98\\
2.00	0.10	0.97\\
2.00	0.09	0.97\\
2.00	0.09	0.96\\
2.00	0.09	0.96\\
2.00	0.09	0.95\\
2.00	0.08	0.94\\
2.00	0.08	0.94\\
2.00	0.08	0.93\\
2.00	0.07	0.93\\
2.00	0.07	0.93\\
2.00	0.06	0.92\\
2.00	0.06	0.92\\
2.00	0.05	0.91\\
2.00	0.05	0.91\\
2.00	0.04	0.91\\
2.00	0.03	0.91\\
2.00	0.03	0.90\\
2.00	0.02	0.90\\
2.00	0.02	0.90\\
2.00	0.01	0.90\\
2.00	0.00	0.90\\
2.00	-0.00	0.90\\
2.00	-0.01	0.90\\
2.00	-0.02	0.90\\
2.00	-0.02	0.90\\
2.00	-0.03	0.90\\
2.00	-0.03	0.91\\
2.00	-0.04	0.91\\
2.00	-0.05	0.91\\
2.00	-0.05	0.91\\
2.00	-0.06	0.92\\
2.00	-0.06	0.92\\
2.00	-0.07	0.93\\
2.00	-0.07	0.93\\
2.00	-0.08	0.93\\
2.00	-0.08	0.94\\
2.00	-0.08	0.94\\
2.00	-0.09	0.95\\
2.00	-0.09	0.96\\
2.00	-0.09	0.96\\
2.00	-0.09	0.97\\
2.00	-0.10	0.97\\
2.00	-0.10	0.98\\
2.00	-0.10	0.99\\
2.00	-0.10	0.99\\
2.00	-0.10	1.00\\
2.00	-0.10	1.00\\
2.00	-0.10	1.01\\
2.00	-0.10	1.02\\
2.00	-0.10	1.02\\
2.00	-0.10	1.03\\
2.00	-0.09	1.04\\
2.00	-0.09	1.04\\
2.00	-0.09	1.05\\
2.00	-0.08	1.05\\
2.00	-0.08	1.06\\
2.00	-0.08	1.06\\
2.00	-0.07	1.07\\
2.00	-0.07	1.07\\
2.00	-0.06	1.08\\
2.00	-0.06	1.08\\
2.00	-0.05	1.08\\
2.00	-0.05	1.09\\
2.00	-0.04	1.09\\
2.00	-0.04	1.09\\
2.00	-0.03	1.10\\
2.00	-0.03	1.10\\
2.00	-0.02	1.10\\
2.00	-0.01	1.10\\
2.00	-0.01	1.10\\
2.00	-0.00	1.10\\
};
 \addlegendentry{Entrance}
\addplot3 [color=black, line width=1.5pt]
 table[row sep=crcr] {%
1.00	0.50	1.00\\
1.00	0.50	1.00\\
1.00	0.50	1.00\\
1.00	0.50	1.00\\
1.00	0.50	1.00\\
1.00	0.50	1.00\\
1.01	0.50	1.00\\
1.01	0.50	1.00\\
1.01	0.50	1.00\\
1.02	0.50	1.00\\
1.02	0.49	1.00\\
1.03	0.49	1.00\\
1.03	0.49	1.00\\
1.04	0.49	1.00\\
1.05	0.49	1.00\\
1.05	0.49	1.00\\
1.06	0.48	1.00\\
1.07	0.48	1.00\\
1.08	0.48	1.00\\
1.08	0.48	1.00\\
1.09	0.48	1.00\\
1.10	0.47	1.00\\
1.11	0.47	1.00\\
1.12	0.47	1.00\\
1.13	0.47	1.00\\
1.13	0.47	1.00\\
1.14	0.46	1.00\\
1.15	0.46	1.00\\
1.16	0.46	1.00\\
1.17	0.46	1.00\\
1.17	0.46	1.00\\
1.18	0.45	1.00\\
1.19	0.45	1.00\\
1.20	0.45	1.00\\
1.21	0.45	1.00\\
1.21	0.45	1.00\\
1.22	0.44	1.00\\
1.23	0.44	1.00\\
1.24	0.44	1.00\\
1.25	0.44	1.00\\
1.25	0.44	1.00\\
1.26	0.43	1.00\\
1.27	0.43	1.00\\
1.28	0.43	1.00\\
1.28	0.43	1.00\\
1.29	0.43	1.00\\
1.30	0.43	1.00\\
1.31	0.42	1.00\\
1.31	0.42	1.00\\
1.32	0.42	1.00\\
1.33	0.42	1.00\\
1.33	0.42	1.00\\
1.34	0.41	1.00\\
1.35	0.41	1.00\\
1.36	0.41	1.00\\
1.36	0.41	1.00\\
1.37	0.41	1.00\\
1.38	0.41	1.00\\
1.38	0.40	1.00\\
1.39	0.40	1.00\\
1.40	0.40	1.00\\
1.40	0.40	1.00\\
1.41	0.40	1.00\\
1.42	0.39	1.00\\
1.42	0.39	1.00\\
1.43	0.39	1.00\\
1.44	0.39	1.00\\
1.44	0.39	1.00\\
1.45	0.39	1.00\\
1.45	0.38	1.00\\
1.46	0.38	1.00\\
1.46	0.38	1.00\\
1.47	0.38	1.00\\
1.47	0.38	1.00\\
1.48	0.37	1.00\\
1.48	0.37	1.00\\
1.49	0.37	1.00\\
1.49	0.37	1.00\\
1.50	0.36	1.00\\
1.50	0.36	1.00\\
1.51	0.36	1.00\\
1.51	0.35	1.00\\
1.52	0.35	1.00\\
1.52	0.35	1.00\\
1.53	0.34	1.00\\
1.53	0.34	1.00\\
1.54	0.34	1.00\\
1.54	0.33	1.00\\
1.54	0.33	1.00\\
1.55	0.32	1.00\\
1.55	0.32	1.00\\
1.56	0.32	1.00\\
1.56	0.31	1.00\\
1.56	0.31	1.00\\
1.57	0.31	1.00\\
1.57	0.30	1.00\\
1.58	0.30	1.00\\
1.58	0.29	1.00\\
1.58	0.29	1.00\\
1.59	0.29	1.00\\
1.59	0.29	1.00\\
1.60	0.28	1.00\\
1.60	0.28	1.00\\
1.60	0.28	0.99\\
1.61	0.27	0.97\\
1.61	0.27	0.96\\
1.62	0.27	0.95\\
1.62	0.26	0.94\\
1.63	0.26	0.93\\
1.63	0.26	0.92\\
1.64	0.26	0.92\\
1.64	0.25	0.91\\
1.65	0.25	0.91\\
1.65	0.25	0.90\\
1.65	0.24	0.90\\
1.66	0.24	0.90\\
1.66	0.24	0.90\\
1.66	0.24	0.89\\
1.67	0.23	0.89\\
1.67	0.23	0.89\\
1.67	0.23	0.89\\
1.68	0.23	0.89\\
1.68	0.23	0.90\\
1.68	0.23	0.91\\
1.69	0.23	0.92\\
1.69	0.23	0.92\\
1.70	0.23	0.93\\
1.70	0.23	0.93\\
1.70	0.23	0.94\\
1.71	0.23	0.94\\
1.71	0.23	0.95\\
1.72	0.22	0.95\\
1.72	0.22	0.96\\
1.73	0.22	0.96\\
1.73	0.22	0.97\\
1.73	0.22	0.97\\
1.74	0.22	0.97\\
1.74	0.22	0.98\\
1.75	0.22	1.00\\
1.75	0.22	1.03\\
1.75	0.22	1.05\\
1.75	0.22	1.07\\
1.75	0.22	1.08\\
1.75	0.21	1.10\\
1.75	0.21	1.11\\
1.75	0.21	1.12\\
1.76	0.20	1.13\\
1.76	0.20	1.14\\
1.76	0.20	1.15\\
1.76	0.20	1.16\\
1.76	0.19	1.16\\
1.76	0.19	1.16\\
1.76	0.19	1.17\\
1.76	0.19	1.17\\
1.76	0.19	1.17\\
1.76	0.18	1.17\\
1.76	0.18	1.17\\
1.76	0.18	1.17\\
1.76	0.18	1.17\\
1.76	0.18	1.16\\
1.77	0.18	1.16\\
1.77	0.18	1.16\\
1.77	0.18	1.17\\
1.77	0.18	1.17\\
1.77	0.18	1.17\\
1.77	0.18	1.17\\
1.77	0.18	1.17\\
1.77	0.18	1.16\\
1.77	0.18	1.16\\
1.77	0.18	1.16\\
1.77	0.18	1.16\\
1.77	0.18	1.15\\
1.77	0.18	1.15\\
1.77	0.18	1.15\\
1.77	0.18	1.14\\
1.77	0.19	1.14\\
1.77	0.19	1.14\\
1.78	0.19	1.13\\
1.78	0.19	1.13\\
1.78	0.19	1.12\\
};
 \addlegendentry{Path1}

\addplot3 [color=white!60!black!, line width=1.5pt]
 table[row sep=crcr] {%
1.00	0.00	1.00\\
1.00	0.00	1.00\\
1.00	-0.00	1.00\\
1.00	-0.00	1.00\\
1.00	-0.00	1.00\\
1.00	-0.00	1.00\\
1.01	-0.00	1.00\\
1.01	-0.00	1.00\\
1.01	-0.00	1.00\\
1.02	-0.00	1.00\\
1.02	-0.00	1.00\\
1.03	-0.00	1.00\\
1.03	-0.00	1.00\\
1.04	-0.00	1.00\\
1.05	-0.00	1.00\\
1.05	-0.00	1.00\\
1.06	-0.00	1.00\\
1.07	-0.00	1.00\\
1.08	-0.00	1.00\\
1.08	-0.00	1.00\\
1.09	-0.00	1.00\\
1.10	-0.00	1.00\\
1.11	-0.00	1.00\\
1.12	-0.00	1.00\\
1.13	-0.00	1.00\\
1.13	-0.00	1.00\\
1.14	-0.00	1.00\\
1.15	-0.00	1.00\\
1.16	-0.00	1.00\\
1.17	-0.00	1.00\\
1.17	-0.00	1.00\\
1.18	-0.00	1.00\\
1.19	-0.00	1.00\\
1.20	-0.00	1.00\\
1.21	-0.00	1.00\\
1.21	-0.00	1.00\\
1.22	-0.00	1.00\\
1.23	-0.00	1.00\\
1.24	-0.00	1.00\\
1.25	-0.00	1.00\\
1.25	-0.00	1.00\\
1.26	-0.00	1.00\\
1.27	-0.00	1.00\\
1.28	-0.00	1.00\\
1.28	-0.00	1.00\\
1.29	-0.00	1.00\\
1.30	-0.00	1.00\\
1.31	-0.00	1.00\\
1.31	-0.00	1.00\\
1.32	-0.00	1.00\\
1.33	-0.00	1.00\\
1.34	-0.00	1.00\\
1.34	-0.00	1.00\\
1.35	-0.00	1.00\\
1.36	-0.00	1.00\\
1.36	-0.00	1.00\\
1.37	-0.00	1.00\\
1.38	-0.00	1.00\\
1.39	-0.00	1.00\\
1.39	-0.00	1.00\\
1.40	-0.00	1.00\\
1.41	-0.00	1.00\\
1.41	-0.00	1.00\\
1.42	-0.00	1.00\\
1.43	-0.00	1.00\\
1.43	-0.00	1.00\\
1.44	-0.00	1.00\\
1.45	-0.00	1.00\\
1.45	-0.00	1.00\\
1.46	-0.00	1.00\\
1.47	-0.00	1.00\\
1.47	-0.00	1.00\\
1.48	-0.00	1.00\\
1.49	-0.00	1.00\\
1.49	-0.00	1.00\\
1.50	-0.00	1.00\\
1.51	-0.00	1.00\\
1.51	-0.00	1.00\\
1.52	-0.00	1.00\\
1.52	-0.00	1.00\\
1.53	-0.00	1.00\\
1.54	-0.00	1.00\\
1.54	-0.00	1.00\\
1.55	-0.00	1.00\\
1.55	-0.00	1.00\\
1.56	-0.00	1.00\\
1.56	-0.00	1.00\\
1.57	-0.00	1.00\\
1.57	-0.00	1.00\\
1.58	-0.00	1.00\\
1.58	-0.00	1.00\\
1.59	-0.00	1.00\\
1.59	-0.00	1.00\\
1.60	-0.00	1.00\\
1.60	-0.00	1.00\\
1.61	-0.00	1.00\\
1.61	-0.00	1.00\\
1.61	-0.00	1.00\\
1.62	-0.00	1.00\\
1.62	-0.00	1.00\\
1.63	-0.00	1.00\\
1.63	-0.00	1.00\\
1.63	-0.00	1.00\\
1.64	-0.00	1.00\\
1.64	-0.00	1.00\\
1.64	-0.00	1.00\\
1.65	-0.00	1.00\\
1.65	-0.00	1.00\\
1.65	-0.00	1.00\\
1.66	-0.00	1.00\\
1.66	-0.00	1.00\\
1.66	-0.00	1.00\\
1.67	-0.00	1.00\\
1.67	-0.00	1.00\\
1.67	-0.00	1.00\\
1.67	-0.00	1.00\\
1.68	-0.00	1.00\\
1.68	-0.00	1.00\\
1.68	-0.00	1.00\\
1.68	-0.00	1.00\\
1.68	-0.00	1.00\\
1.69	-0.00	1.00\\
1.69	-0.00	1.00\\
1.69	-0.00	1.00\\
1.69	-0.00	1.00\\
1.69	-0.00	1.00\\
1.69	-0.00	1.00\\
1.69	-0.00	1.00\\
1.69	-0.00	1.00\\
1.69	-0.00	1.00\\
1.69	-0.00	1.00\\
1.69	-0.00	1.00\\
1.70	-0.00	1.00\\
1.70	-0.00	1.00\\
1.70	-0.00	1.00\\
1.70	-0.00	1.00\\
1.70	-0.00	1.00\\
1.70	-0.00	1.00\\
1.70	-0.00	1.00\\
1.70	-0.00	1.00\\
1.69	-0.00	1.00\\
1.69	-0.00	1.00\\
1.69	-0.00	1.00\\
1.69	-0.00	1.00\\
1.69	-0.00	1.00\\
1.69	-0.00	1.00\\
1.69	-0.00	1.00\\
1.69	-0.00	1.00\\
1.69	-0.00	1.00\\
1.69	-0.00	1.00\\
1.69	-0.00	1.00\\
1.69	-0.00	1.00\\
1.69	-0.00	1.00\\
1.69	-0.00	1.00\\
1.69	-0.00	1.00\\
1.69	-0.00	1.00\\
1.69	-0.00	1.00\\
1.69	-0.00	1.00\\
1.69	-0.00	1.00\\
1.69	-0.00	1.00\\
1.69	-0.00	1.00\\
1.69	-0.00	1.00\\
1.69	-0.00	1.00\\
1.69	-0.00	1.00\\
1.69	-0.00	1.00\\
1.69	-0.00	1.00\\
1.69	-0.00	1.00\\
1.69	-0.00	1.00\\
1.69	-0.00	1.00\\
1.69	-0.00	1.00\\
1.70	-0.00	1.00\\
1.70	-0.00	1.00\\
1.70	-0.00	1.00\\
1.70	-0.00	1.00\\
1.70	-0.00	1.00\\
1.70	-0.00	1.00\\
1.70	-0.00	1.00\\
1.70	-0.00	1.00\\
1.70	-0.00	1.00\\
1.70	-0.00	1.00\\
};
 \addlegendentry{Path2}

\addplot3 [color=white!30!black!, line width=1.5pt]
 table[row sep=crcr] {%
1.00	-0.50	1.00\\
1.00	-0.50	1.00\\
1.00	-0.50	1.00\\
1.00	-0.50	1.00\\
1.00	-0.50	1.00\\
1.00	-0.50	1.00\\
1.01	-0.50	1.00\\
1.01	-0.50	1.00\\
1.01	-0.50	1.00\\
1.02	-0.50	1.00\\
1.02	-0.49	1.00\\
1.03	-0.49	1.00\\
1.03	-0.49	1.00\\
1.04	-0.49	1.00\\
1.05	-0.49	1.00\\
1.05	-0.49	1.00\\
1.06	-0.48	1.00\\
1.07	-0.48	1.00\\
1.08	-0.48	1.00\\
1.08	-0.48	1.00\\
1.09	-0.48	1.00\\
1.10	-0.47	1.00\\
1.11	-0.47	1.00\\
1.12	-0.47	1.00\\
1.13	-0.47	1.00\\
1.13	-0.47	1.00\\
1.14	-0.46	1.00\\
1.15	-0.46	1.00\\
1.16	-0.46	1.00\\
1.17	-0.46	1.00\\
1.17	-0.46	1.00\\
1.18	-0.45	1.00\\
1.19	-0.45	1.00\\
1.20	-0.45	1.00\\
1.21	-0.45	1.00\\
1.21	-0.45	1.00\\
1.22	-0.44	1.00\\
1.23	-0.44	1.00\\
1.24	-0.44	1.00\\
1.25	-0.44	1.00\\
1.25	-0.44	1.00\\
1.26	-0.43	1.00\\
1.27	-0.43	1.00\\
1.28	-0.43	1.00\\
1.28	-0.43	1.00\\
1.29	-0.43	1.00\\
1.30	-0.43	1.00\\
1.31	-0.42	1.00\\
1.31	-0.42	1.00\\
1.32	-0.42	1.00\\
1.33	-0.42	1.00\\
1.33	-0.42	1.00\\
1.34	-0.41	1.00\\
1.35	-0.41	1.00\\
1.36	-0.41	1.00\\
1.36	-0.41	1.00\\
1.37	-0.41	1.00\\
1.38	-0.41	1.00\\
1.38	-0.40	1.00\\
1.39	-0.40	1.00\\
1.40	-0.40	1.00\\
1.40	-0.40	1.00\\
1.41	-0.40	1.00\\
1.42	-0.39	1.00\\
1.42	-0.39	1.00\\
1.43	-0.39	1.00\\
1.44	-0.39	1.00\\
1.44	-0.39	1.00\\
1.45	-0.39	1.00\\
1.45	-0.38	1.00\\
1.46	-0.38	1.00\\
1.46	-0.38	1.00\\
1.47	-0.38	1.00\\
1.47	-0.38	1.00\\
1.48	-0.37	1.00\\
1.48	-0.37	1.00\\
1.49	-0.37	1.00\\
1.49	-0.37	1.00\\
1.50	-0.36	1.00\\
1.50	-0.36	1.00\\
1.51	-0.36	1.00\\
1.51	-0.35	1.00\\
1.52	-0.35	1.00\\
1.52	-0.35	1.00\\
1.53	-0.34	1.00\\
1.53	-0.34	1.00\\
1.54	-0.34	1.00\\
1.54	-0.33	1.00\\
1.54	-0.33	1.00\\
1.55	-0.32	1.00\\
1.55	-0.32	1.00\\
1.56	-0.32	1.00\\
1.56	-0.31	1.00\\
1.56	-0.31	1.00\\
1.57	-0.31	1.00\\
1.57	-0.30	1.00\\
1.58	-0.30	1.00\\
1.58	-0.30	1.00\\
1.59	-0.29	1.00\\
1.59	-0.29	1.00\\
1.60	-0.29	1.00\\
1.60	-0.28	1.00\\
1.60	-0.28	1.00\\
1.61	-0.28	1.00\\
1.61	-0.28	1.00\\
1.62	-0.27	1.00\\
1.62	-0.27	1.00\\
1.62	-0.27	1.00\\
1.63	-0.27	1.00\\
1.63	-0.27	1.00\\
1.64	-0.26	1.00\\
1.64	-0.26	1.00\\
1.64	-0.26	1.00\\
1.65	-0.26	1.00\\
1.65	-0.26	1.00\\
1.65	-0.26	1.00\\
1.66	-0.25	1.00\\
1.66	-0.25	1.00\\
1.66	-0.25	1.00\\
1.67	-0.25	1.00\\
1.67	-0.25	1.00\\
1.67	-0.25	1.00\\
1.68	-0.25	1.00\\
1.68	-0.24	1.00\\
1.68	-0.24	1.00\\
1.69	-0.24	1.00\\
1.69	-0.24	1.00\\
1.69	-0.24	1.00\\
1.69	-0.24	1.00\\
1.70	-0.24	1.00\\
1.70	-0.24	1.00\\
1.70	-0.24	1.00\\
1.70	-0.24	1.00\\
1.71	-0.24	1.00\\
1.71	-0.24	1.00\\
1.71	-0.24	1.00\\
1.71	-0.24	1.00\\
1.71	-0.24	1.00\\
1.72	-0.23	1.00\\
1.72	-0.23	1.00\\
1.72	-0.23	1.00\\
1.72	-0.23	1.00\\
1.72	-0.23	1.00\\
1.73	-0.23	1.00\\
1.73	-0.23	1.00\\
1.73	-0.23	1.00\\
1.73	-0.22	1.00\\
1.74	-0.22	1.00\\
1.74	-0.22	1.00\\
1.74	-0.22	1.00\\
1.74	-0.22	1.00\\
1.75	-0.22	1.00\\
1.75	-0.22	1.00\\
1.75	-0.22	1.00\\
1.75	-0.22	1.00\\
1.76	-0.22	1.00\\
1.76	-0.22	1.00\\
1.76	-0.22	1.00\\
1.76	-0.22	1.00\\
1.77	-0.22	1.00\\
1.77	-0.22	1.00\\
1.77	-0.22	1.00\\
1.77	-0.22	1.00\\
1.77	-0.22	1.00\\
1.78	-0.22	1.00\\
1.78	-0.22	1.00\\
1.78	-0.22	1.00\\
1.78	-0.22	1.00\\
1.78	-0.22	1.00\\
1.78	-0.22	1.00\\
1.78	-0.22	1.00\\
1.79	-0.22	1.00\\
1.79	-0.22	1.00\\
1.79	-0.22	1.00\\
1.79	-0.22	1.00\\
1.79	-0.22	1.00\\
1.79	-0.22	1.00\\
1.79	-0.22	1.00\\
1.79	-0.22	1.00\\
1.79	-0.22	1.00\\
};
 \addlegendentry{Path3}

 \addplot3 [color=white!30!black!, only marks, mark size=3.5pt, mark=*, mark options={solid, fill=white!30!black!}]
 table[row sep=crcr] {%
1.00	0.50	1.00\\
};
 \addlegendentry{Start}
 \addplot3 [color=white!60!black!, mark size=2.5pt, mark=square*, mark options={solid, fill=white!60!black!}]
 table[row sep=crcr] {%
3.00	0.00	1.00\\
};
\addlegendentry{Goal}
 \addplot3 [color=white!30!black!, only marks, mark size=3.5pt, mark=*, mark options={solid, fill=white!30!black!}]
 table[row sep=crcr] {%
1.00	0.00	1.00\\
};
 \addplot3 [color=white!30!black!, only marks, mark size=3.5pt, mark=*, mark options={solid, fill=white!30!black!}]
 table[row sep=crcr] {%
1.00	-0.50	1.00\\
};
 \end{axis}

\begin{axis}[%
width=0.2377\fwidth,
height=0.1925\fwidth,
at={(0.70\fwidth,1.56\fwidth)},
scale only axis,
xmin=1.00,
xmax=3.00,
xlabel style={font=\color{white!15!black},font=\tiny,shift={(0.0,0.1)}},
xlabel={$x$ [m]},
ymin=-0.50,
ymax=0.50,
ylabel style={font=\color{white!15!black},font=\tiny,shift={(0.0,-0.2)}},
ticklabel style = {font=\tiny},  
ylabel={$y$ [m]},
axis background/.style={fill=white},
xmajorgrids,
ymajorgrids
]
%
%
\addplot [color=black, line width=1.0pt, forget plot]
  table[row sep=crcr]{%
1.00	0.50\\
1.00	0.50\\
1.00	0.50\\
1.00	0.50\\
1.00	0.50\\
1.00	0.50\\
1.01	0.50\\
1.01	0.50\\
1.01	0.50\\
1.02	0.50\\
1.02	0.49\\
1.03	0.49\\
1.03	0.49\\
1.04	0.49\\
1.05	0.49\\
1.05	0.49\\
1.06	0.48\\
1.07	0.48\\
1.08	0.48\\
1.08	0.48\\
1.09	0.48\\
1.10	0.47\\
1.11	0.47\\
1.12	0.47\\
1.13	0.47\\
1.13	0.47\\
1.14	0.46\\
1.15	0.46\\
1.16	0.46\\
1.17	0.46\\
1.17	0.46\\
1.18	0.45\\
1.19	0.45\\
1.20	0.45\\
1.21	0.45\\
1.21	0.45\\
1.22	0.44\\
1.23	0.44\\
1.24	0.44\\
1.25	0.44\\
1.25	0.44\\
1.26	0.43\\
1.27	0.43\\
1.28	0.43\\
1.28	0.43\\
1.29	0.43\\
1.30	0.43\\
1.31	0.42\\
1.31	0.42\\
1.32	0.42\\
1.33	0.42\\
1.33	0.42\\
1.34	0.41\\
1.35	0.41\\
1.36	0.41\\
1.36	0.41\\
1.37	0.41\\
1.38	0.41\\
1.38	0.40\\
1.39	0.40\\
1.40	0.40\\
1.40	0.40\\
1.41	0.40\\
1.42	0.39\\
1.42	0.39\\
1.43	0.39\\
1.44	0.39\\
1.44	0.39\\
1.45	0.39\\
1.45	0.38\\
1.46	0.38\\
1.46	0.38\\
1.47	0.38\\
1.47	0.38\\
1.48	0.37\\
1.48	0.37\\
1.49	0.37\\
1.49	0.37\\
1.50	0.36\\
1.50	0.36\\
1.51	0.36\\
1.51	0.35\\
1.52	0.35\\
1.52	0.35\\
1.53	0.34\\
1.53	0.34\\
1.54	0.34\\
1.54	0.33\\
1.54	0.33\\
1.55	0.32\\
1.55	0.32\\
1.56	0.32\\
1.56	0.31\\
1.56	0.31\\
1.57	0.31\\
1.57	0.30\\
1.58	0.30\\
1.58	0.29\\
1.58	0.29\\
1.59	0.29\\
1.59	0.29\\
1.60	0.28\\
1.60	0.28\\
1.60	0.28\\
1.61	0.27\\
1.61	0.27\\
1.62	0.27\\
1.62	0.26\\
1.63	0.26\\
1.63	0.26\\
1.64	0.26\\
1.64	0.25\\
1.65	0.25\\
1.65	0.25\\
1.65	0.24\\
1.66	0.24\\
1.66	0.24\\
1.66	0.24\\
1.67	0.23\\
1.67	0.23\\
1.67	0.23\\
1.68	0.23\\
1.68	0.23\\
1.68	0.23\\
1.69	0.23\\
1.69	0.23\\
1.70	0.23\\
1.70	0.23\\
1.70	0.23\\
1.71	0.23\\
1.71	0.23\\
1.72	0.22\\
1.72	0.22\\
1.73	0.22\\
1.73	0.22\\
1.73	0.22\\
1.74	0.22\\
1.74	0.22\\
1.75	0.22\\
1.75	0.22\\
1.75	0.22\\
1.75	0.22\\
1.75	0.22\\
1.75	0.21\\
1.75	0.21\\
1.75	0.21\\
1.76	0.20\\
1.76	0.20\\
1.76	0.20\\
1.76	0.20\\
1.76	0.19\\
1.76	0.19\\
1.76	0.19\\
1.76	0.19\\
1.76	0.19\\
1.76	0.18\\
1.76	0.18\\
1.76	0.18\\
1.76	0.18\\
1.76	0.18\\
1.77	0.18\\
1.77	0.18\\
1.77	0.18\\
1.77	0.18\\
1.77	0.18\\
1.77	0.18\\
1.77	0.18\\
1.77	0.18\\
1.77	0.18\\
1.77	0.18\\
1.77	0.18\\
1.77	0.18\\
1.77	0.18\\
1.77	0.18\\
1.77	0.18\\
1.77	0.19\\
1.77	0.19\\
1.78	0.19\\
1.78	0.19\\
1.78	0.19\\
};
\addplot [color=white!60!black!, line width=1.0pt, forget plot]
  table[row sep=crcr]{%
1.00	0.00\\
1.00	0.00\\
1.00	-0.00\\
1.00	-0.00\\
1.00	-0.00\\
1.00	-0.00\\
1.01	-0.00\\
1.01	-0.00\\
1.01	-0.00\\
1.02	-0.00\\
1.02	-0.00\\
1.03	-0.00\\
1.03	-0.00\\
1.04	-0.00\\
1.05	-0.00\\
1.05	-0.00\\
1.06	-0.00\\
1.07	-0.00\\
1.08	-0.00\\
1.08	-0.00\\
1.09	-0.00\\
1.10	-0.00\\
1.11	-0.00\\
1.12	-0.00\\
1.13	-0.00\\
1.13	-0.00\\
1.14	-0.00\\
1.15	-0.00\\
1.16	-0.00\\
1.17	-0.00\\
1.17	-0.00\\
1.18	-0.00\\
1.19	-0.00\\
1.20	-0.00\\
1.21	-0.00\\
1.21	-0.00\\
1.22	-0.00\\
1.23	-0.00\\
1.24	-0.00\\
1.25	-0.00\\
1.25	-0.00\\
1.26	-0.00\\
1.27	-0.00\\
1.28	-0.00\\
1.28	-0.00\\
1.29	-0.00\\
1.30	-0.00\\
1.31	-0.00\\
1.31	-0.00\\
1.32	-0.00\\
1.33	-0.00\\
1.34	-0.00\\
1.34	-0.00\\
1.35	-0.00\\
1.36	-0.00\\
1.36	-0.00\\
1.37	-0.00\\
1.38	-0.00\\
1.39	-0.00\\
1.39	-0.00\\
1.40	-0.00\\
1.41	-0.00\\
1.41	-0.00\\
1.42	-0.00\\
1.43	-0.00\\
1.43	-0.00\\
1.44	-0.00\\
1.45	-0.00\\
1.45	-0.00\\
1.46	-0.00\\
1.47	-0.00\\
1.47	-0.00\\
1.48	-0.00\\
1.49	-0.00\\
1.49	-0.00\\
1.50	-0.00\\
1.51	-0.00\\
1.51	-0.00\\
1.52	-0.00\\
1.52	-0.00\\
1.53	-0.00\\
1.54	-0.00\\
1.54	-0.00\\
1.55	-0.00\\
1.55	-0.00\\
1.56	-0.00\\
1.56	-0.00\\
1.57	-0.00\\
1.57	-0.00\\
1.58	-0.00\\
1.58	-0.00\\
1.59	-0.00\\
1.59	-0.00\\
1.60	-0.00\\
1.60	-0.00\\
1.61	-0.00\\
1.61	-0.00\\
1.61	-0.00\\
1.62	-0.00\\
1.62	-0.00\\
1.63	-0.00\\
1.63	-0.00\\
1.63	-0.00\\
1.64	-0.00\\
1.64	-0.00\\
1.64	-0.00\\
1.65	-0.00\\
1.65	-0.00\\
1.65	-0.00\\
1.66	-0.00\\
1.66	-0.00\\
1.66	-0.00\\
1.67	-0.00\\
1.67	-0.00\\
1.67	-0.00\\
1.67	-0.00\\
1.68	-0.00\\
1.68	-0.00\\
1.68	-0.00\\
1.68	-0.00\\
1.68	-0.00\\
1.69	-0.00\\
1.69	-0.00\\
1.69	-0.00\\
1.69	-0.00\\
1.69	-0.00\\
1.69	-0.00\\
1.69	-0.00\\
1.69	-0.00\\
1.69	-0.00\\
1.69	-0.00\\
1.69	-0.00\\
1.70	-0.00\\
1.70	-0.00\\
1.70	-0.00\\
1.70	-0.00\\
1.70	-0.00\\
1.70	-0.00\\
1.70	-0.00\\
1.70	-0.00\\
1.69	-0.00\\
1.69	-0.00\\
1.69	-0.00\\
1.69	-0.00\\
1.69	-0.00\\
1.69	-0.00\\
1.69	-0.00\\
1.69	-0.00\\
1.69	-0.00\\
1.69	-0.00\\
1.69	-0.00\\
1.69	-0.00\\
1.69	-0.00\\
1.69	-0.00\\
1.69	-0.00\\
1.69	-0.00\\
1.69	-0.00\\
1.69	-0.00\\
1.69	-0.00\\
1.69	-0.00\\
1.69	-0.00\\
1.69	-0.00\\
1.69	-0.00\\
1.69	-0.00\\
1.69	-0.00\\
1.69	-0.00\\
1.69	-0.00\\
1.69	-0.00\\
1.69	-0.00\\
1.69	-0.00\\
1.70	-0.00\\
1.70	-0.00\\
1.70	-0.00\\
1.70	-0.00\\
1.70	-0.00\\
1.70	-0.00\\
1.70	-0.00\\
1.70	-0.00\\
1.70	-0.00\\
1.70	-0.00\\
};
\addplot [color=white!30!black!, line width=1.0pt, forget plot]
  table[row sep=crcr]{%
1.00	-0.50\\
1.00	-0.50\\
1.00	-0.50\\
1.00	-0.50\\
1.00	-0.50\\
1.00	-0.50\\
1.01	-0.50\\
1.01	-0.50\\
1.01	-0.50\\
1.02	-0.50\\
1.02	-0.49\\
1.03	-0.49\\
1.03	-0.49\\
1.04	-0.49\\
1.05	-0.49\\
1.05	-0.49\\
1.06	-0.48\\
1.07	-0.48\\
1.08	-0.48\\
1.08	-0.48\\
1.09	-0.48\\
1.10	-0.47\\
1.11	-0.47\\
1.12	-0.47\\
1.13	-0.47\\
1.13	-0.47\\
1.14	-0.46\\
1.15	-0.46\\
1.16	-0.46\\
1.17	-0.46\\
1.17	-0.46\\
1.18	-0.45\\
1.19	-0.45\\
1.20	-0.45\\
1.21	-0.45\\
1.21	-0.45\\
1.22	-0.44\\
1.23	-0.44\\
1.24	-0.44\\
1.25	-0.44\\
1.25	-0.44\\
1.26	-0.43\\
1.27	-0.43\\
1.28	-0.43\\
1.28	-0.43\\
1.29	-0.43\\
1.30	-0.43\\
1.31	-0.42\\
1.31	-0.42\\
1.32	-0.42\\
1.33	-0.42\\
1.33	-0.42\\
1.34	-0.41\\
1.35	-0.41\\
1.36	-0.41\\
1.36	-0.41\\
1.37	-0.41\\
1.38	-0.41\\
1.38	-0.40\\
1.39	-0.40\\
1.40	-0.40\\
1.40	-0.40\\
1.41	-0.40\\
1.42	-0.39\\
1.42	-0.39\\
1.43	-0.39\\
1.44	-0.39\\
1.44	-0.39\\
1.45	-0.39\\
1.45	-0.38\\
1.46	-0.38\\
1.46	-0.38\\
1.47	-0.38\\
1.47	-0.38\\
1.48	-0.37\\
1.48	-0.37\\
1.49	-0.37\\
1.49	-0.37\\
1.50	-0.36\\
1.50	-0.36\\
1.51	-0.36\\
1.51	-0.35\\
1.52	-0.35\\
1.52	-0.35\\
1.53	-0.34\\
1.53	-0.34\\
1.54	-0.34\\
1.54	-0.33\\
1.54	-0.33\\
1.55	-0.32\\
1.55	-0.32\\
1.56	-0.32\\
1.56	-0.31\\
1.56	-0.31\\
1.57	-0.31\\
1.57	-0.30\\
1.58	-0.30\\
1.58	-0.30\\
1.59	-0.29\\
1.59	-0.29\\
1.60	-0.29\\
1.60	-0.28\\
1.60	-0.28\\
1.61	-0.28\\
1.61	-0.28\\
1.62	-0.27\\
1.62	-0.27\\
1.62	-0.27\\
1.63	-0.27\\
1.63	-0.27\\
1.64	-0.26\\
1.64	-0.26\\
1.64	-0.26\\
1.65	-0.26\\
1.65	-0.26\\
1.65	-0.26\\
1.66	-0.25\\
1.66	-0.25\\
1.66	-0.25\\
1.67	-0.25\\
1.67	-0.25\\
1.67	-0.25\\
1.68	-0.25\\
1.68	-0.24\\
1.68	-0.24\\
1.69	-0.24\\
1.69	-0.24\\
1.69	-0.24\\
1.69	-0.24\\
1.70	-0.24\\
1.70	-0.24\\
1.70	-0.24\\
1.70	-0.24\\
1.71	-0.24\\
1.71	-0.24\\
1.71	-0.24\\
1.71	-0.24\\
1.71	-0.24\\
1.72	-0.23\\
1.72	-0.23\\
1.72	-0.23\\
1.72	-0.23\\
1.72	-0.23\\
1.73	-0.23\\
1.73	-0.23\\
1.73	-0.23\\
1.73	-0.22\\
1.74	-0.22\\
1.74	-0.22\\
1.74	-0.22\\
1.74	-0.22\\
1.75	-0.22\\
1.75	-0.22\\
1.75	-0.22\\
1.75	-0.22\\
1.76	-0.22\\
1.76	-0.22\\
1.76	-0.22\\
1.76	-0.22\\
1.77	-0.22\\
1.77	-0.22\\
1.77	-0.22\\
1.77	-0.22\\
1.77	-0.22\\
1.78	-0.22\\
1.78	-0.22\\
1.78	-0.22\\
1.78	-0.22\\
1.78	-0.22\\
1.78	-0.22\\
1.78	-0.22\\
1.79	-0.22\\
1.79	-0.22\\
1.79	-0.22\\
1.79	-0.22\\
1.79	-0.22\\
1.79	-0.22\\
1.79	-0.22\\
1.79	-0.22\\
1.79	-0.22\\
};
\addplot [color=mycolor4, line width=2.0pt, forget plot]
  table[row sep=crcr]{%
2.00	0.00\\
2.00	0.01\\
2.00	0.01\\
2.00	0.02\\
2.00	0.03\\
2.00	0.03\\
2.00	0.04\\
2.00	0.04\\
2.00	0.05\\
2.00	0.05\\
2.00	0.06\\
2.00	0.06\\
2.00	0.07\\
2.00	0.07\\
2.00	0.08\\
2.00	0.08\\
2.00	0.08\\
2.00	0.09\\
2.00	0.09\\
2.00	0.09\\
2.00	0.10\\
2.00	0.10\\
2.00	0.10\\
2.00	0.10\\
2.00	0.10\\
2.00	0.10\\
2.00	0.10\\
2.00	0.10\\
2.00	0.10\\
2.00	0.10\\
2.00	0.09\\
2.00	0.09\\
2.00	0.09\\
2.00	0.09\\
2.00	0.08\\
2.00	0.08\\
2.00	0.08\\
2.00	0.07\\
2.00	0.07\\
2.00	0.06\\
2.00	0.06\\
2.00	0.05\\
2.00	0.05\\
2.00	0.04\\
2.00	0.03\\
2.00	0.03\\
2.00	0.02\\
2.00	0.02\\
2.00	0.01\\
2.00	0.00\\
2.00	-0.00\\
2.00	-0.01\\
2.00	-0.02\\
2.00	-0.02\\
2.00	-0.03\\
2.00	-0.03\\
2.00	-0.04\\
2.00	-0.05\\
2.00	-0.05\\
2.00	-0.06\\
2.00	-0.06\\
2.00	-0.07\\
2.00	-0.07\\
2.00	-0.08\\
2.00	-0.08\\
2.00	-0.08\\
2.00	-0.09\\
2.00	-0.09\\
2.00	-0.09\\
2.00	-0.09\\
2.00	-0.10\\
2.00	-0.10\\
2.00	-0.10\\
2.00	-0.10\\
2.00	-0.10\\
2.00	-0.10\\
2.00	-0.10\\
2.00	-0.10\\
2.00	-0.10\\
2.00	-0.10\\
2.00	-0.09\\
2.00	-0.09\\
2.00	-0.09\\
2.00	-0.08\\
2.00	-0.08\\
2.00	-0.08\\
2.00	-0.07\\
2.00	-0.07\\
2.00	-0.06\\
2.00	-0.06\\
2.00	-0.05\\
2.00	-0.05\\
2.00	-0.04\\
2.00	-0.04\\
2.00	-0.03\\
2.00	-0.03\\
2.00	-0.02\\
2.00	-0.01\\
2.00	-0.01\\
2.00	-0.00\\
};
\addplot [color=white!30!black!, only marks, mark size=2.0pt, mark=*, mark options={solid, fill=white!30!black!}, forget plot]
  table[row sep=crcr]{%
1.00	0.50\\
};
\addplot [color=white!30!black!, only marks, mark size=2.0pt, mark=*, mark options={solid, fill=white!30!black!}, forget plot]
  table[row sep=crcr]{%
1.00	0.00\\
};
\addplot [color=white!30!black!, only marks, mark size=2.0pt, mark=*, mark options={solid, fill=white!30!black!}, forget plot]
  table[row sep=crcr]{%
1.00	-0.50\\
};
\addplot [color=white!60!black!, mark size=1.5pt, mark=square*, mark options={solid, fill=white!60!black!}, forget plot]
  table[row sep=crcr]{%
3.00	0.00\\
};
\end{axis}

\end{tikzpicture}%

%% file: Tikz/config_not_pass2.tex
%
%
\begin{tikzpicture}

\begin{axis}[%
width=1.05\fwidth,
height=0.38\fwidth,
at={(0\fwidth,1.1\fwidth)},
scale only axis,
xmin=0.00,
xmax=9.00,
ymin=0.20,
ymax=1.20,
ylabel style={font=\color{white!15!black}},
ylabel={$\unit[d]{[m]}$},
axis background/.style={fill=white},
xmajorgrids,
ymajorgrids,
legend style={legend cell align=left, align=left, legend columns=3, draw=white!15!black}
]
\addplot [color=black, dotted,line width=1.5pt]
  table[row sep=crcr]{%
0.05	1.12\\
0.10	1.12\\
0.15	1.12\\
0.20	1.12\\
0.25	1.12\\
0.30	1.11\\
0.35	1.11\\
0.40	1.11\\
0.45	1.10\\
0.50	1.10\\
0.55	1.10\\
0.60	1.09\\
0.65	1.08\\
0.70	1.08\\
0.75	1.07\\
0.80	1.06\\
0.85	1.06\\
0.90	1.05\\
0.95	1.04\\
1.00	1.03\\
1.05	1.02\\
1.10	1.02\\
1.15	1.01\\
1.20	1.00\\
1.25	0.99\\
1.30	0.98\\
1.35	0.98\\
1.40	0.97\\
1.45	0.96\\
1.50	0.95\\
1.55	0.94\\
1.60	0.94\\
1.65	0.93\\
1.70	0.92\\
1.75	0.91\\
1.80	0.90\\
1.85	0.90\\
1.90	0.89\\
1.95	0.88\\
2.00	0.87\\
2.05	0.87\\
2.10	0.86\\
2.15	0.85\\
2.20	0.84\\
2.25	0.84\\
2.30	0.83\\
2.35	0.82\\
2.40	0.81\\
2.45	0.81\\
2.50	0.80\\
2.55	0.79\\
2.60	0.78\\
2.65	0.78\\
2.70	0.77\\
2.75	0.76\\
2.80	0.76\\
2.85	0.75\\
2.90	0.74\\
2.95	0.74\\
3.00	0.73\\
3.05	0.72\\
3.10	0.72\\
3.15	0.71\\
3.20	0.70\\
3.25	0.70\\
3.30	0.69\\
3.35	0.69\\
3.40	0.68\\
3.45	0.67\\
3.50	0.67\\
3.55	0.66\\
3.60	0.66\\
3.65	0.65\\
3.70	0.65\\
3.75	0.64\\
3.80	0.64\\
3.85	0.63\\
3.90	0.62\\
3.95	0.62\\
4.00	0.61\\
4.05	0.61\\
4.10	0.60\\
4.15	0.60\\
4.20	0.59\\
4.25	0.59\\
4.30	0.58\\
4.35	0.57\\
4.40	0.57\\
4.45	0.56\\
4.50	0.56\\
4.55	0.55\\
4.60	0.55\\
4.65	0.54\\
4.70	0.53\\
4.75	0.53\\
4.80	0.52\\
4.85	0.52\\
4.90	0.51\\
4.95	0.51\\
5.00	0.50\\
5.05	0.50\\
5.10	0.49\\
5.15	0.49\\
5.20	0.48\\
5.25	0.48\\
5.30	0.47\\
5.35	0.47\\
5.40	0.46\\
5.45	0.46\\
5.50	0.46\\
5.55	0.45\\
5.60	0.45\\
5.65	0.44\\
5.70	0.44\\
5.75	0.44\\
5.80	0.43\\
5.85	0.43\\
5.90	0.42\\
5.95	0.42\\
6.00	0.42\\
6.05	0.41\\
6.10	0.41\\
6.15	0.41\\
6.20	0.40\\
6.25	0.40\\
6.30	0.39\\
6.35	0.39\\
6.40	0.38\\
6.45	0.38\\
6.50	0.37\\
6.55	0.37\\
6.60	0.36\\
6.65	0.36\\
6.70	0.36\\
6.75	0.35\\
6.80	0.35\\
6.85	0.35\\
6.90	0.34\\
6.95	0.34\\
7.00	0.34\\
7.05	0.34\\
7.10	0.34\\
7.15	0.34\\
7.20	0.34\\
7.25	0.34\\
7.30	0.34\\
7.35	0.35\\
7.40	0.35\\
7.45	0.35\\
7.50	0.35\\
7.55	0.35\\
7.60	0.35\\
7.65	0.35\\
7.70	0.35\\
7.75	0.35\\
7.80	0.35\\
7.85	0.35\\
7.90	0.34\\
7.95	0.34\\
8.00	0.34\\
8.05	0.34\\
8.10	0.34\\
8.15	0.34\\
8.20	0.34\\
8.25	0.34\\
8.30	0.34\\
8.35	0.34\\
8.40	0.34\\
8.45	0.34\\
8.50	0.34\\
8.55	0.33\\
8.60	0.33\\
8.65	0.33\\
8.70	0.33\\
8.75	0.33\\
8.80	0.32\\
8.85	0.32\\
8.90	0.32\\
8.95	0.32\\
9.00	0.32\\
};
\addlegendentry{$d_1$}

\addplot [color=black, dashed, line width=1.5pt]
  table[row sep=crcr]{%
0.05	1.00\\
0.10	1.00\\
0.15	1.00\\
0.20	1.00\\
0.25	1.00\\
0.30	1.00\\
0.35	0.99\\
0.40	0.99\\
0.45	0.99\\
0.50	0.98\\
0.55	0.98\\
0.60	0.97\\
0.65	0.97\\
0.70	0.96\\
0.75	0.95\\
0.80	0.95\\
0.85	0.94\\
0.90	0.93\\
0.95	0.92\\
1.00	0.92\\
1.05	0.91\\
1.10	0.90\\
1.15	0.89\\
1.20	0.88\\
1.25	0.87\\
1.30	0.87\\
1.35	0.86\\
1.40	0.85\\
1.45	0.84\\
1.50	0.83\\
1.55	0.83\\
1.60	0.82\\
1.65	0.81\\
1.70	0.80\\
1.75	0.79\\
1.80	0.79\\
1.85	0.78\\
1.90	0.77\\
1.95	0.76\\
2.00	0.75\\
2.05	0.75\\
2.10	0.74\\
2.15	0.73\\
2.20	0.72\\
2.25	0.72\\
2.30	0.71\\
2.35	0.70\\
2.40	0.69\\
2.45	0.69\\
2.50	0.68\\
2.55	0.67\\
2.60	0.66\\
2.65	0.66\\
2.70	0.65\\
2.75	0.64\\
2.80	0.64\\
2.85	0.63\\
2.90	0.62\\
2.95	0.61\\
3.00	0.61\\
3.05	0.60\\
3.10	0.59\\
3.15	0.59\\
3.20	0.58\\
3.25	0.57\\
3.30	0.57\\
3.35	0.56\\
3.40	0.55\\
3.45	0.55\\
3.50	0.54\\
3.55	0.53\\
3.60	0.53\\
3.65	0.52\\
3.70	0.51\\
3.75	0.51\\
3.80	0.50\\
3.85	0.49\\
3.90	0.49\\
3.95	0.48\\
4.00	0.48\\
4.05	0.47\\
4.10	0.46\\
4.15	0.46\\
4.20	0.45\\
4.25	0.45\\
4.30	0.44\\
4.35	0.44\\
4.40	0.43\\
4.45	0.43\\
4.50	0.42\\
4.55	0.42\\
4.60	0.41\\
4.65	0.41\\
4.70	0.40\\
4.75	0.40\\
4.80	0.39\\
4.85	0.39\\
4.90	0.39\\
4.95	0.38\\
5.00	0.38\\
5.05	0.37\\
5.10	0.37\\
5.15	0.37\\
5.20	0.36\\
5.25	0.36\\
5.30	0.36\\
5.35	0.35\\
5.40	0.35\\
5.45	0.35\\
5.50	0.34\\
5.55	0.34\\
5.60	0.34\\
5.65	0.33\\
5.70	0.33\\
5.75	0.33\\
5.80	0.33\\
5.85	0.32\\
5.90	0.32\\
5.95	0.32\\
6.00	0.32\\
6.05	0.32\\
6.10	0.31\\
6.15	0.31\\
6.20	0.31\\
6.25	0.31\\
6.30	0.31\\
6.35	0.31\\
6.40	0.31\\
6.45	0.31\\
6.50	0.31\\
6.55	0.31\\
6.60	0.31\\
6.65	0.30\\
6.70	0.30\\
6.75	0.30\\
6.80	0.30\\
6.85	0.30\\
6.90	0.30\\
6.95	0.30\\
7.00	0.30\\
7.05	0.31\\
7.10	0.31\\
7.15	0.31\\
7.20	0.31\\
7.25	0.31\\
7.30	0.31\\
7.35	0.31\\
7.40	0.31\\
7.45	0.31\\
7.50	0.31\\
7.55	0.31\\
7.60	0.31\\
7.65	0.31\\
7.70	0.31\\
7.75	0.31\\
7.80	0.31\\
7.85	0.31\\
7.90	0.31\\
7.95	0.31\\
8.00	0.31\\
8.05	0.31\\
8.10	0.31\\
8.15	0.31\\
8.20	0.31\\
8.25	0.31\\
8.30	0.31\\
8.35	0.31\\
8.40	0.31\\
8.45	0.31\\
8.50	0.31\\
8.55	0.30\\
8.60	0.30\\
8.65	0.30\\
8.70	0.30\\
8.75	0.30\\
8.80	0.30\\
8.85	0.30\\
8.90	0.30\\
8.95	0.30\\
9.00	0.30\\
};
\addlegendentry{$d_2$}

\addplot [color=black, line width=1.0pt]
  table[row sep=crcr]{%
0.05	1.12\\
0.10	1.12\\
0.15	1.12\\
0.20	1.12\\
0.25	1.12\\
0.30	1.11\\
0.35	1.11\\
0.40	1.11\\
0.45	1.10\\
0.50	1.10\\
0.55	1.10\\
0.60	1.09\\
0.65	1.08\\
0.70	1.08\\
0.75	1.07\\
0.80	1.06\\
0.85	1.06\\
0.90	1.05\\
0.95	1.04\\
1.00	1.03\\
1.05	1.02\\
1.10	1.02\\
1.15	1.01\\
1.20	1.00\\
1.25	0.99\\
1.30	0.98\\
1.35	0.98\\
1.40	0.97\\
1.45	0.96\\
1.50	0.95\\
1.55	0.94\\
1.60	0.94\\
1.65	0.93\\
1.70	0.92\\
1.75	0.91\\
1.80	0.90\\
1.85	0.90\\
1.90	0.89\\
1.95	0.88\\
2.00	0.87\\
2.05	0.87\\
2.10	0.86\\
2.15	0.85\\
2.20	0.84\\
2.25	0.84\\
2.30	0.83\\
2.35	0.82\\
2.40	0.81\\
2.45	0.81\\
2.50	0.80\\
2.55	0.79\\
2.60	0.78\\
2.65	0.78\\
2.70	0.77\\
2.75	0.76\\
2.80	0.76\\
2.85	0.75\\
2.90	0.74\\
2.95	0.74\\
3.00	0.73\\
3.05	0.72\\
3.10	0.72\\
3.15	0.71\\
3.20	0.70\\
3.25	0.70\\
3.30	0.69\\
3.35	0.69\\
3.40	0.68\\
3.45	0.67\\
3.50	0.67\\
3.55	0.66\\
3.60	0.66\\
3.65	0.65\\
3.70	0.65\\
3.75	0.64\\
3.80	0.64\\
3.85	0.63\\
3.90	0.62\\
3.95	0.62\\
4.00	0.61\\
4.05	0.61\\
4.10	0.60\\
4.15	0.60\\
4.20	0.59\\
4.25	0.58\\
4.30	0.58\\
4.35	0.57\\
4.40	0.57\\
4.45	0.56\\
4.50	0.56\\
4.55	0.55\\
4.60	0.55\\
4.65	0.54\\
4.70	0.53\\
4.75	0.53\\
4.80	0.52\\
4.85	0.52\\
4.90	0.51\\
4.95	0.51\\
5.00	0.50\\
5.05	0.50\\
5.10	0.49\\
5.15	0.49\\
5.20	0.48\\
5.25	0.48\\
5.30	0.47\\
5.35	0.47\\
5.40	0.46\\
5.45	0.46\\
5.50	0.45\\
5.55	0.45\\
5.60	0.45\\
5.65	0.44\\
5.70	0.44\\
5.75	0.43\\
5.80	0.43\\
5.85	0.43\\
5.90	0.42\\
5.95	0.42\\
6.00	0.42\\
6.05	0.41\\
6.10	0.41\\
6.15	0.41\\
6.20	0.40\\
6.25	0.40\\
6.30	0.40\\
6.35	0.40\\
6.40	0.39\\
6.45	0.39\\
6.50	0.39\\
6.55	0.39\\
6.60	0.38\\
6.65	0.38\\
6.70	0.38\\
6.75	0.38\\
6.80	0.37\\
6.85	0.37\\
6.90	0.37\\
6.95	0.37\\
7.00	0.37\\
7.05	0.36\\
7.10	0.36\\
7.15	0.36\\
7.20	0.36\\
7.25	0.35\\
7.30	0.35\\
7.35	0.35\\
7.40	0.35\\
7.45	0.34\\
7.50	0.34\\
7.55	0.34\\
7.60	0.34\\
7.65	0.33\\
7.70	0.33\\
7.75	0.33\\
7.80	0.33\\
7.85	0.33\\
7.90	0.32\\
7.95	0.32\\
8.00	0.32\\
8.05	0.32\\
8.10	0.32\\
8.15	0.32\\
8.20	0.31\\
8.25	0.31\\
8.30	0.31\\
8.35	0.31\\
8.40	0.31\\
8.45	0.31\\
8.50	0.31\\
8.55	0.31\\
8.60	0.31\\
8.65	0.31\\
8.70	0.31\\
8.75	0.31\\
8.80	0.31\\
8.85	0.30\\
8.90	0.30\\
8.95	0.30\\
9.00	0.30\\
};
\addlegendentry{$d_3$}
\end{axis}

\begin{axis}[%
width=1.05\fwidth,
height=0.38\fwidth,
at={(0\fwidth,0.57\fwidth)},
scale only axis,
xmin=0.00,
xmax=9.00,
xlabel style={font=\color{white!15!black}},
xlabel={Time [sec]},
ymin=0.00,
ymax=0.80,
ylabel style={font=\color{white!15!black}},
ylabel={$\unit[r]{[m]}$},
axis background/.style={fill=white},
xmajorgrids,
ymajorgrids,
legend style={at={(0.93,1.23)},legend cell align=left, align=left, legend columns=4, draw=white!15!black}
]
\addplot [color=black, line width=3.0pt]
  table[row sep=crcr]{%
0.05	0.67\\
0.10	0.67\\
0.15	0.67\\
0.20	0.67\\
0.25	0.67\\
0.30	0.67\\
0.35	0.67\\
0.40	0.67\\
0.45	0.67\\
0.50	0.67\\
0.55	0.67\\
0.60	0.67\\
0.65	0.67\\
0.70	0.67\\
0.75	0.67\\
0.80	0.67\\
0.85	0.67\\
0.90	0.67\\
0.95	0.67\\
1.00	0.67\\
1.05	0.67\\
1.10	0.67\\
1.15	0.67\\
1.20	0.67\\
1.25	0.67\\
1.30	0.67\\
1.35	0.67\\
1.40	0.67\\
1.45	0.67\\
1.50	0.67\\
1.55	0.67\\
1.60	0.67\\
1.65	0.67\\
1.70	0.67\\
1.75	0.67\\
1.80	0.67\\
1.85	0.67\\
1.90	0.67\\
1.95	0.67\\
2.00	0.67\\
2.05	0.67\\
2.10	0.67\\
2.15	0.67\\
2.20	0.67\\
2.25	0.67\\
2.30	0.67\\
2.35	0.67\\
2.40	0.67\\
2.45	0.67\\
2.50	0.67\\
2.55	0.67\\
2.60	0.67\\
2.65	0.67\\
2.70	0.67\\
2.75	0.67\\
2.80	0.67\\
2.85	0.67\\
2.90	0.67\\
2.95	0.67\\
3.00	0.67\\
3.05	0.67\\
3.10	0.67\\
3.15	0.67\\
3.20	0.67\\
3.25	0.67\\
3.30	0.67\\
3.35	0.67\\
3.40	0.67\\
3.45	0.67\\
3.50	0.67\\
3.55	0.67\\
3.60	0.67\\
3.65	0.67\\
3.70	0.67\\
3.75	0.67\\
3.80	0.67\\
3.85	0.67\\
3.90	0.67\\
3.95	0.67\\
4.00	0.67\\
4.05	0.67\\
4.10	0.67\\
4.15	0.67\\
4.20	0.67\\
4.25	0.67\\
4.30	0.67\\
4.35	0.67\\
4.40	0.67\\
4.45	0.67\\
4.50	0.67\\
4.55	0.67\\
4.60	0.67\\
4.65	0.67\\
4.70	0.67\\
4.75	0.67\\
4.80	0.67\\
4.85	0.67\\
4.90	0.67\\
4.95	0.67\\
5.00	0.67\\
5.05	0.67\\
5.10	0.67\\
5.15	0.67\\
5.20	0.67\\
5.25	0.67\\
5.30	0.67\\
5.35	0.67\\
5.40	0.67\\
5.45	0.67\\
5.50	0.67\\
5.55	0.67\\
5.60	0.67\\
5.65	0.67\\
5.70	0.67\\
5.75	0.67\\
5.80	0.67\\
5.85	0.67\\
5.90	0.67\\
5.95	0.67\\
6.00	0.67\\
6.05	0.67\\
6.10	0.67\\
6.15	0.68\\
6.20	0.68\\
6.25	0.68\\
6.30	0.67\\
6.35	0.67\\
6.40	0.67\\
6.45	0.67\\
6.50	0.67\\
6.55	0.67\\
6.60	0.67\\
6.65	0.67\\
6.70	0.67\\
6.75	0.67\\
6.80	0.67\\
6.85	0.67\\
6.90	0.69\\
6.95	0.67\\
7.00	0.67\\
7.05	0.67\\
7.10	0.67\\
7.15	0.67\\
7.20	0.67\\
7.25	0.67\\
7.30	0.67\\
7.35	0.67\\
7.40	0.67\\
7.45	0.67\\
7.50	0.67\\
7.55	0.67\\
7.60	0.67\\
7.65	0.67\\
7.70	0.67\\
7.75	0.67\\
7.80	0.67\\
7.85	0.67\\
7.90	0.67\\
7.95	0.67\\
8.00	0.67\\
8.05	0.67\\
8.10	0.67\\
8.15	0.67\\
8.20	0.67\\
8.25	0.67\\
8.30	0.67\\
8.35	0.67\\
8.40	0.67\\
8.45	0.67\\
8.50	0.67\\
8.55	0.67\\
8.60	0.67\\
8.65	0.67\\
8.70	0.67\\
8.75	0.67\\
8.80	0.67\\
8.85	0.67\\
8.90	0.67\\
8.95	0.67\\
9.00	0.67\\
};
\addlegendentry{$r_{\text{front}_{1,2,3}}$}

\addplot [color=white!80!black!, line width=1.0pt]
  table[row sep=crcr]{%
0.05	0.67\\
0.10	0.67\\
0.15	0.67\\
0.20	0.67\\
0.25	0.67\\
0.30	0.67\\
0.35	0.67\\
0.40	0.67\\
0.45	0.67\\
0.50	0.67\\
0.55	0.67\\
0.60	0.67\\
0.65	0.67\\
0.70	0.67\\
0.75	0.67\\
0.80	0.67\\
0.85	0.67\\
0.90	0.67\\
0.95	0.67\\
1.00	0.67\\
1.05	0.67\\
1.10	0.67\\
1.15	0.67\\
1.20	0.67\\
1.25	0.67\\
1.30	0.67\\
1.35	0.67\\
1.40	0.67\\
1.45	0.67\\
1.50	0.67\\
1.55	0.67\\
1.60	0.67\\
1.65	0.67\\
1.70	0.67\\
1.75	0.67\\
1.80	0.67\\
1.85	0.67\\
1.90	0.67\\
1.95	0.67\\
2.00	0.67\\
2.05	0.67\\
2.10	0.67\\
2.15	0.67\\
2.20	0.67\\
2.25	0.67\\
2.30	0.67\\
2.35	0.67\\
2.40	0.67\\
2.45	0.67\\
2.50	0.67\\
2.55	0.67\\
2.60	0.67\\
2.65	0.67\\
2.70	0.67\\
2.75	0.67\\
2.80	0.67\\
2.85	0.67\\
2.90	0.67\\
2.95	0.67\\
3.00	0.67\\
3.05	0.67\\
3.10	0.67\\
3.15	0.67\\
3.20	0.67\\
3.25	0.67\\
3.30	0.67\\
3.35	0.67\\
3.40	0.67\\
3.45	0.67\\
3.50	0.67\\
3.55	0.67\\
3.60	0.67\\
3.65	0.67\\
3.70	0.67\\
3.75	0.67\\
3.80	0.67\\
3.85	0.67\\
3.90	0.67\\
3.95	0.67\\
4.00	0.67\\
4.05	0.67\\
4.10	0.67\\
4.15	0.67\\
4.20	0.67\\
4.25	0.67\\
4.30	0.67\\
4.35	0.67\\
4.40	0.67\\
4.45	0.67\\
4.50	0.67\\
4.55	0.67\\
4.60	0.67\\
4.65	0.67\\
4.70	0.67\\
4.75	0.67\\
4.80	0.67\\
4.85	0.67\\
4.90	0.67\\
4.95	0.67\\
5.00	0.67\\
5.05	0.67\\
5.10	0.67\\
5.15	0.67\\
5.20	0.67\\
5.25	0.67\\
5.30	0.67\\
5.35	0.67\\
5.40	0.67\\
5.45	0.67\\
5.50	0.67\\
5.55	0.67\\
5.60	0.67\\
5.65	0.67\\
5.70	0.67\\
5.75	0.67\\
5.80	0.67\\
5.85	0.67\\
5.90	0.67\\
5.95	0.67\\
6.00	0.67\\
6.05	0.67\\
6.10	0.67\\
6.15	0.68\\
6.20	0.68\\
6.25	0.68\\
6.30	0.67\\
6.35	0.67\\
6.40	0.67\\
6.45	0.67\\
6.50	0.67\\
6.55	0.67\\
6.60	0.67\\
6.65	0.67\\
6.70	0.67\\
6.75	0.67\\
6.80	0.67\\
6.85	0.67\\
6.90	0.69\\
6.95	0.67\\
7.00	0.67\\
7.05	0.67\\
7.10	0.67\\
7.15	0.67\\
7.20	0.67\\
7.25	0.67\\
7.30	0.67\\
7.35	0.67\\
7.40	0.67\\
7.45	0.67\\
7.50	0.67\\
7.55	0.67\\
7.60	0.67\\
7.65	0.67\\
7.70	0.67\\
7.75	0.67\\
7.80	0.67\\
7.85	0.67\\
7.90	0.67\\
7.95	0.67\\
8.00	0.67\\
8.05	0.67\\
8.10	0.67\\
8.15	0.67\\
8.20	0.67\\
8.25	0.67\\
8.30	0.67\\
8.35	0.67\\
8.40	0.67\\
8.45	0.67\\
8.50	0.67\\
8.55	0.67\\
8.60	0.67\\
8.65	0.67\\
8.70	0.67\\
8.75	0.67\\
8.80	0.67\\
8.85	0.67\\
8.90	0.67\\
8.95	0.67\\
9.00	0.67\\
};
\addlegendentry{$r_{\text{rear}_{1,2,3}}$}

\addplot [color=black, dotted, line width=1.0pt]
  table[row sep=crcr]{%
0.05	0.20\\
0.10	0.20\\
0.15	0.20\\
0.20	0.20\\
0.25	0.20\\
0.30	0.20\\
0.35	0.20\\
0.40	0.20\\
0.45	0.20\\
0.50	0.20\\
0.55	0.20\\
0.60	0.20\\
0.65	0.20\\
0.70	0.20\\
0.75	0.20\\
0.80	0.20\\
0.85	0.20\\
0.90	0.20\\
0.95	0.20\\
1.00	0.20\\
1.05	0.20\\
1.10	0.20\\
1.15	0.20\\
1.20	0.20\\
1.25	0.20\\
1.30	0.20\\
1.35	0.20\\
1.40	0.20\\
1.45	0.20\\
1.50	0.20\\
1.55	0.20\\
1.60	0.20\\
1.65	0.20\\
1.70	0.20\\
1.75	0.20\\
1.80	0.20\\
1.85	0.20\\
1.90	0.20\\
1.95	0.20\\
2.00	0.20\\
2.05	0.20\\
2.10	0.20\\
2.15	0.20\\
2.20	0.20\\
2.25	0.20\\
2.30	0.20\\
2.35	0.20\\
2.40	0.20\\
2.45	0.20\\
2.50	0.20\\
2.55	0.20\\
2.60	0.20\\
2.65	0.20\\
2.70	0.20\\
2.75	0.20\\
2.80	0.20\\
2.85	0.20\\
2.90	0.20\\
2.95	0.20\\
3.00	0.20\\
3.05	0.20\\
3.10	0.20\\
3.15	0.20\\
3.20	0.20\\
3.25	0.20\\
3.30	0.20\\
3.35	0.20\\
3.40	0.20\\
3.45	0.20\\
3.50	0.20\\
3.55	0.20\\
3.60	0.20\\
3.65	0.20\\
3.70	0.20\\
3.75	0.20\\
3.80	0.20\\
3.85	0.20\\
3.90	0.20\\
3.95	0.20\\
4.00	0.20\\
4.05	0.20\\
4.10	0.20\\
4.15	0.20\\
4.20	0.20\\
4.25	0.20\\
4.30	0.20\\
4.35	0.20\\
4.40	0.20\\
4.45	0.20\\
4.50	0.20\\
4.55	0.20\\
4.60	0.20\\
4.65	0.20\\
4.70	0.20\\
4.75	0.20\\
4.80	0.20\\
4.85	0.20\\
4.90	0.20\\
4.95	0.20\\
5.00	0.20\\
5.05	0.20\\
5.10	0.20\\
5.15	0.20\\
5.20	0.20\\
5.25	0.20\\
5.30	0.20\\
5.35	0.20\\
5.40	0.20\\
5.45	0.20\\
5.50	0.20\\
5.55	0.20\\
5.60	0.20\\
5.65	0.20\\
5.70	0.20\\
5.75	0.20\\
5.80	0.20\\
5.85	0.20\\
5.90	0.20\\
5.95	0.20\\
6.00	0.20\\
6.05	0.20\\
6.10	0.20\\
6.15	0.20\\
6.20	0.20\\
6.25	0.20\\
6.30	0.20\\
6.35	0.20\\
6.40	0.20\\
6.45	0.20\\
6.50	0.20\\
6.55	0.20\\
6.60	0.20\\
6.65	0.20\\
6.70	0.20\\
6.75	0.20\\
6.80	0.20\\
6.85	0.20\\
6.90	0.20\\
6.95	0.20\\
7.00	0.20\\
7.05	0.20\\
7.10	0.20\\
7.15	0.20\\
7.20	0.20\\
7.25	0.20\\
7.30	0.20\\
7.35	0.20\\
7.40	0.20\\
7.45	0.20\\
7.50	0.20\\
7.55	0.20\\
7.60	0.20\\
7.65	0.20\\
7.70	0.20\\
7.75	0.20\\
7.80	0.20\\
7.85	0.20\\
7.90	0.20\\
7.95	0.20\\
8.00	0.20\\
8.05	0.20\\
8.10	0.20\\
8.15	0.20\\
8.20	0.20\\
8.25	0.20\\
8.30	0.20\\
8.35	0.20\\
8.40	0.20\\
8.45	0.20\\
8.50	0.20\\
8.55	0.20\\
8.60	0.20\\
8.65	0.20\\
8.70	0.20\\
8.75	0.20\\
8.80	0.20\\
8.85	0.20\\
8.90	0.20\\
8.95	0.20\\
9.00	0.20\\
};
\addlegendentry{$r_{obs}$}
\end{axis}
\end{tikzpicture}%

%% file: Tikz/3Dplot_rect.tex
%
%
\definecolor{mycolor1}{rgb}{0.85000,0.32500,0.09800}%
\definecolor{mycolor2}{rgb}{0.92900,0.69400,0.12500}%
\definecolor{mycolor3}{rgb}{0.46600,0.67400,0.18800}%
\definecolor{mycolor4}{rgb}{0.30100,0.74500,0.93300}%
\begin{tikzpicture}

\begin{axis}[%
width=0.951\fwidth,
height=0.77\fwidth,
at={(0\fwidth,1.1\fwidth)},
scale only axis,
xmin=1.00,
xmax=5.00,
tick align=outside,
xlabel style={font=\color{white!15!black}},
xlabel={$x$ [m]},
ymin=-0.20,
ymax=0.80,
ylabel style={font=\color{white!15!black}},
ylabel={$y$ [m]},
zmin=0.50,
zmax=1.80,
zlabel style={font=\color{white!15!black}},
zlabel={$z$ [m]},
view={-22.80}{21.43},
axis background/.style={fill=white},
xmajorgrids,
ymajorgrids,
zmajorgrids,
legend style={at={(1.05,1.02)},legend cell align=left, align=left, legend columns=5, draw=white!15!black}
]
\addplot3 [color=black, line width=1.5pt]
 table[row sep=crcr] {%
2.18	0.48	1.41\\
2.22	0.49	1.41\\
2.26	0.49	1.42\\
2.30	0.49	1.43\\
2.33	0.50	1.43\\
2.37	0.50	1.44\\
2.40	0.50	1.44\\
2.43	0.50	1.44\\
2.46	0.50	1.45\\
2.49	0.50	1.45\\
2.51	0.50	1.46\\
2.54	0.50	1.46\\
2.56	0.50	1.46\\
2.58	0.50	1.46\\
2.60	0.50	1.47\\
2.62	0.50	1.47\\
2.64	0.50	1.47\\
2.66	0.50	1.47\\
2.67	0.50	1.47\\
2.69	0.50	1.47\\
2.71	0.50	1.47\\
2.72	0.50	1.47\\
2.73	0.50	1.47\\
2.75	0.50	1.48\\
2.76	0.50	1.48\\
2.77	0.50	1.48\\
2.78	0.50	1.48\\
2.79	0.50	1.48\\
2.80	0.50	1.48\\
2.81	0.50	1.48\\
2.82	0.50	1.48\\
2.83	0.50	1.48\\
2.84	0.50	1.48\\
2.84	0.50	1.48\\
2.85	0.50	1.47\\
2.86	0.50	1.47\\
2.86	0.50	1.47\\
2.87	0.50	1.47\\
2.88	0.50	1.47\\
2.88	0.50	1.47\\
2.89	0.50	1.47\\
2.89	0.50	1.47\\
2.90	0.50	1.47\\
2.90	0.50	1.47\\
2.91	0.50	1.47\\
2.91	0.50	1.47\\
2.92	0.50	1.47\\
2.92	0.50	1.47\\
2.93	0.50	1.47\\
2.93	0.50	1.47\\
2.93	0.50	1.47\\
2.94	0.50	1.47\\
2.94	0.50	1.47\\
2.94	0.50	1.47\\
2.94	0.50	1.47\\
2.95	0.50	1.47\\
2.95	0.50	1.47\\
2.95	0.50	1.47\\
2.95	0.50	1.47\\
2.96	0.50	1.47\\
2.96	0.50	1.47\\
2.96	0.50	1.47\\
2.96	0.50	1.47\\
2.96	0.50	1.47\\
2.97	0.50	1.47\\
2.97	0.50	1.47\\
2.97	0.50	1.47\\
2.97	0.50	1.46\\
2.97	0.50	1.46\\
2.97	0.50	1.46\\
2.97	0.50	1.46\\
2.98	0.50	1.46\\
2.98	0.50	1.46\\
2.98	0.50	1.46\\
2.98	0.50	1.46\\
2.98	0.50	1.46\\
2.98	0.50	1.46\\
2.98	0.50	1.46\\
2.98	0.50	1.46\\
2.98	0.50	1.46\\
2.98	0.50	1.46\\
2.99	0.50	1.45\\
2.99	0.50	1.44\\
2.99	0.49	1.43\\
3.00	0.48	1.42\\
3.01	0.48	1.41\\
3.02	0.47	1.39\\
3.04	0.45	1.38\\
3.05	0.44	1.36\\
3.07	0.42	1.35\\
3.10	0.40	1.33\\
3.13	0.37	1.31\\
3.16	0.35	1.30\\
3.19	0.33	1.28\\
3.23	0.30	1.26\\
3.27	0.28	1.25\\
3.32	0.25	1.23\\
3.37	0.23	1.22\\
3.42	0.21	1.20\\
3.47	0.18	1.18\\
4.10	0.03	1.04\\
4.15	0.02	1.03\\
4.19	0.02	1.03\\
4.24	0.01	1.02\\
4.28	0.01	1.01\\
4.32	0.01	1.01\\
4.35	0.00	1.00\\
4.39	0.00	1.00\\
4.42	0.00	0.99\\
4.45	-0.00	0.99\\
4.48	-0.00	0.98\\
4.51	-0.00	0.98\\
4.53	-0.00	0.98\\
4.56	-0.00	0.97\\
4.58	-0.00	0.97\\
4.60	-0.00	0.97\\
4.63	-0.00	0.97\\
4.65	-0.00	0.96\\
4.66	-0.00	0.96\\
4.68	-0.00	0.96\\
4.70	-0.00	0.96\\
4.71	-0.00	0.96\\
4.73	-0.00	0.96\\
4.74	-0.00	0.96\\
4.76	-0.00	0.96\\
4.77	-0.00	0.95\\
4.78	-0.00	0.95\\
4.79	-0.00	0.95\\
4.80	-0.00	0.95\\
4.81	-0.00	0.95\\
4.82	0.00	0.95\\
4.83	0.00	0.95\\
4.84	0.00	0.95\\
4.84	0.00	0.95\\
4.85	0.00	0.95\\
4.86	0.00	0.96\\
4.87	0.00	0.96\\
4.87	0.00	0.96\\
4.88	0.00	0.96\\
4.88	0.00	0.96\\
4.89	0.00	0.96\\
4.90	0.00	0.96\\
4.90	0.00	0.96\\
4.91	0.00	0.96\\
4.91	0.00	0.96\\
4.91	0.00	0.96\\
4.92	0.00	0.96\\
4.92	0.00	0.96\\
4.93	0.00	0.96\\
4.93	0.00	0.96\\
4.93	0.00	0.96\\
4.94	0.00	0.96\\
4.94	0.00	0.96\\
4.94	0.00	0.96\\
4.94	0.00	0.96\\
4.95	0.00	0.96\\
4.95	0.00	0.96\\
4.95	0.00	0.96\\
4.95	0.00	0.97\\
4.96	0.00	0.97\\
4.96	0.00	0.97\\
4.96	0.00	0.97\\
4.96	0.00	0.97\\
4.96	0.00	0.97\\
4.97	0.00	0.97\\
};
 \addlegendentry{Path}

\addplot3 [color=mycolor1, line width=2.0pt]
 table[row sep=crcr] {%
2.00	0.30	1.20\\
2.00	0.30	1.70\\
2.00	0.70	1.70\\
2.00	0.70	1.20\\
2.00	0.30	1.20\\
};
 \addlegendentry{Passage1}

\addplot3 [color=black, line width=1.5pt, forget plot]
 table[row sep=crcr] {%
1.00	0.00	1.00\\
1.00	0.00	1.00\\
1.00	0.00	1.01\\
1.00	0.00	1.01\\
1.01	0.01	1.02\\
1.01	0.01	1.03\\
1.02	0.02	1.04\\
1.03	0.03	1.06\\
1.04	0.04	1.07\\
1.06	0.05	1.09\\
1.07	0.07	1.10\\
1.10	0.09	1.12\\
1.12	0.11	1.13\\
1.15	0.14	1.15\\
1.18	0.16	1.17\\
1.22	0.19	1.18\\
1.26	0.21	1.20\\
1.30	0.23	1.21\\
1.35	0.26	1.23\\
1.40	0.28	1.24\\
1.45	0.31	1.26\\
1.50	0.33	1.27\\
1.56	0.35	1.28\\
1.61	0.37	1.30\\
1.67	0.38	1.31\\
1.72	0.40	1.32\\
1.78	0.41	1.33\\
1.83	0.43	1.34\\
1.89	0.44	1.35\\
1.94	0.45	1.36\\
1.99	0.46	1.37\\
2.04	0.47	1.38\\
2.09	0.47	1.39\\
2.13	0.48	1.40\\
2.18	0.48	1.41\\
};

\addplot3 [color=mycolor2, line width=2.0pt]
 table[row sep=crcr] {%
4.00	-0.15	0.70\\
4.00	-0.15	1.20\\
4.00	0.15	1.20\\
4.00	0.15	0.70\\
4.00	-0.15	0.70\\
};
 \addlegendentry{Passage2}

\addplot3 [color=black, line width=1.5pt, forget plot]
 table[row sep=crcr] {%
3.47	0.18	1.18\\
3.52	0.16	1.17\\
3.58	0.14	1.15\\
3.63	0.13	1.14\\
3.69	0.11	1.13\\
3.74	0.10	1.11\\
3.80	0.08	1.10\\
3.85	0.07	1.09\\
3.91	0.06	1.08\\
3.96	0.05	1.07\\
4.01	0.04	1.06\\
4.06	0.03	1.05\\
4.10	0.03	1.04\\
};

\addplot3 [color=white!30!black!, only marks, mark size=3.5pt, mark=*, mark options={solid, fill=white!30!black!}]
 table[row sep=crcr] {%
1.00	0.00	1.00\\
};
\addlegendentry{Start}
 \addplot3 [color=white!60!black!, mark size=2.5pt, mark=square*, mark options={solid, fill=white!60!black!}]
 table[row sep=crcr] {%
3.00	0.50	1.50\\
};
\addlegendentry{Goal}
 \addplot3 [color=white!60!black!, mark size=2.5pt, mark=square*, mark options={solid, fill=white!60!black!}]
 table[row sep=crcr] {%
5.00	0.00	1.00\\
};
 \end{axis}

\begin{axis}[%
width=0.2377\fwidth,
height=0.1925\fwidth,
at={(0.70\fwidth,1.58\fwidth)},
scale only axis,
xmin=1.00,
xmax=5.00,
xlabel style={font=\color{white!15!black},font=\tiny,shift={(0.0,0.1)}},
xlabel={$x$ [m]},
ymin=-0.20,
ymax=0.80,
ylabel style={font=\color{white!15!black},font=\tiny,shift={(0.0,-0.2)}},
ticklabel style = {font=\tiny},  
ylabel={$y$ [m]},
axis background/.style={fill=white},
xmajorgrids,
ymajorgrids
]
\addplot [color=black, line width=1.0pt, forget plot]
  table[row sep=crcr]{%
1.00	0.00\\
1.00	0.00\\
1.00	0.00\\
1.00	0.00\\
1.01	0.01\\
1.01	0.01\\
1.02	0.02\\
1.03	0.03\\
1.04	0.04\\
1.06	0.05\\
1.07	0.07\\
1.10	0.09\\
1.12	0.11\\
1.15	0.14\\
1.18	0.16\\
1.22	0.19\\
1.26	0.21\\
1.30	0.23\\
1.35	0.26\\
1.40	0.28\\
1.45	0.31\\
1.50	0.33\\
1.56	0.35\\
1.61	0.37\\
1.67	0.38\\
1.72	0.40\\
1.78	0.41\\
1.83	0.43\\
1.89	0.44\\
1.94	0.45\\
1.99	0.46\\
2.04	0.47\\
2.09	0.47\\
2.13	0.48\\
2.18	0.48\\
2.22	0.49\\
2.26	0.49\\
2.30	0.49\\
2.33	0.50\\
2.37	0.50\\
2.40	0.50\\
2.43	0.50\\
2.46	0.50\\
2.49	0.50\\
2.51	0.50\\
2.54	0.50\\
2.56	0.50\\
2.58	0.50\\
2.60	0.50\\
2.62	0.50\\
2.64	0.50\\
2.66	0.50\\
2.67	0.50\\
2.69	0.50\\
2.71	0.50\\
2.72	0.50\\
2.73	0.50\\
2.75	0.50\\
2.76	0.50\\
2.77	0.50\\
2.78	0.50\\
2.79	0.50\\
2.80	0.50\\
2.81	0.50\\
2.82	0.50\\
2.83	0.50\\
2.84	0.50\\
2.84	0.50\\
2.85	0.50\\
2.86	0.50\\
2.86	0.50\\
2.87	0.50\\
2.88	0.50\\
2.88	0.50\\
2.89	0.50\\
2.89	0.50\\
2.90	0.50\\
2.90	0.50\\
2.91	0.50\\
2.91	0.50\\
2.92	0.50\\
2.92	0.50\\
2.93	0.50\\
2.93	0.50\\
2.93	0.50\\
2.94	0.50\\
2.94	0.50\\
2.94	0.50\\
2.94	0.50\\
2.95	0.50\\
2.95	0.50\\
2.95	0.50\\
2.95	0.50\\
2.96	0.50\\
2.96	0.50\\
2.96	0.50\\
2.96	0.50\\
2.96	0.50\\
2.97	0.50\\
2.97	0.50\\
2.97	0.50\\
2.97	0.50\\
2.97	0.50\\
2.97	0.50\\
2.97	0.50\\
2.98	0.50\\
2.98	0.50\\
2.98	0.50\\
2.98	0.50\\
2.98	0.50\\
2.98	0.50\\
2.98	0.50\\
2.98	0.50\\
2.98	0.50\\
2.98	0.50\\
2.99	0.50\\
2.99	0.50\\
2.99	0.49\\
3.00	0.48\\
3.01	0.48\\
3.02	0.47\\
3.04	0.45\\
3.05	0.44\\
3.07	0.42\\
3.10	0.40\\
3.13	0.37\\
3.16	0.35\\
3.19	0.33\\
3.23	0.30\\
3.27	0.28\\
3.32	0.25\\
3.37	0.23\\
3.42	0.21\\
3.47	0.18\\
3.52	0.16\\
3.58	0.14\\
3.63	0.13\\
3.69	0.11\\
3.74	0.10\\
3.80	0.08\\
3.85	0.07\\
3.91	0.06\\
3.96	0.05\\
4.01	0.04\\
4.06	0.03\\
4.10	0.03\\
4.15	0.02\\
4.19	0.02\\
4.24	0.01\\
4.28	0.01\\
4.32	0.01\\
4.35	0.00\\
4.39	0.00\\
4.42	0.00\\
4.45	-0.00\\
4.48	-0.00\\
4.51	-0.00\\
4.53	-0.00\\
4.56	-0.00\\
4.58	-0.00\\
4.60	-0.00\\
4.63	-0.00\\
4.65	-0.00\\
4.66	-0.00\\
4.68	-0.00\\
4.70	-0.00\\
4.71	-0.00\\
4.73	-0.00\\
4.74	-0.00\\
4.76	-0.00\\
4.77	-0.00\\
4.78	-0.00\\
4.79	-0.00\\
4.80	-0.00\\
4.81	-0.00\\
4.82	0.00\\
4.83	0.00\\
4.84	0.00\\
4.84	0.00\\
4.85	0.00\\
4.86	0.00\\
4.87	0.00\\
4.87	0.00\\
4.88	0.00\\
4.88	0.00\\
4.89	0.00\\
4.90	0.00\\
4.90	0.00\\
4.91	0.00\\
4.91	0.00\\
4.91	0.00\\
4.92	0.00\\
4.92	0.00\\
4.93	0.00\\
4.93	0.00\\
4.93	0.00\\
4.94	0.00\\
4.94	0.00\\
4.94	0.00\\
4.94	0.00\\
4.95	0.00\\
4.95	0.00\\
4.95	0.00\\
4.95	0.00\\
4.96	0.00\\
4.96	0.00\\
4.96	0.00\\
4.96	0.00\\
4.96	0.00\\
4.97	0.00\\
};
\addplot [color=mycolor1, line width=2.0pt, forget plot]
  table[row sep=crcr]{%
2.00	0.30\\
2.00	0.70\\
};
\addplot [color=mycolor2, line width=2.0pt, forget plot]
  table[row sep=crcr]{%
4.00	-0.15\\
4.00	0.15\\
};
\addplot [color=white!30!black!, only marks, mark size=2.0pt, mark=*, mark options={solid, fill=white!30!black!}, forget plot]
  table[row sep=crcr]{%
1.00	0.00\\
};
\addplot [color=white!60!black!, mark size=1.5pt, mark=square*, mark options={solid, fill=white!60!black!}, forget plot]
  table[row sep=crcr]{%
3.00	0.50\\
};
\addplot [color=white!60!black!, mark size=1.5pt, mark=square*, mark options={solid, fill=white!60!black!}, forget plot]
  table[row sep=crcr]{%
5.00	0.00\\
};
\end{axis}

\end{tikzpicture}%

%% file: Tikz/config_rect.tex
%
%
\begin{tikzpicture}

\begin{axis}[%
width=1.05\fwidth,
height=0.38\fwidth,
at={(0\fwidth,1.1\fwidth)},
scale only axis,
xmin=0.00,
xmax=10.50,
ymin=0.00,
ymax=3.05,
ylabel style={font=\color{white!15!black}},
ylabel={$\unit[d]{[m]}$},
axis background/.style={fill=white},
xmajorgrids,
ymajorgrids,
legend style={at={(0.63,0.8)},legend cell align=left, align=left, legend columns=2, draw=white!15!black}
]
\addplot [color=black, dotted,line width=1.5pt]
  table[row sep=crcr]{%
0.05	1.21\\
0.10	1.20\\
0.15	1.20\\
0.20	1.20\\
0.25	1.19\\
0.30	1.18\\
0.35	1.17\\
0.40	1.15\\
0.45	1.13\\
0.50	1.11\\
0.55	1.08\\
0.60	1.05\\
0.65	1.01\\
0.70	0.97\\
0.75	0.93\\
0.80	0.88\\
0.85	0.83\\
0.90	0.78\\
0.95	0.73\\
1.00	0.67\\
1.05	0.61\\
1.10	0.56\\
1.15	0.50\\
1.20	0.44\\
1.25	0.38\\
1.30	0.32\\
1.35	0.27\\
1.40	0.21\\
1.45	0.16\\
1.50	0.12\\
1.55	0.09\\
1.60	0.08\\
1.65	0.11\\
1.70	0.14\\
1.75	0.18\\
1.80	0.22\\
1.85	0.26\\
1.90	0.30\\
1.95	0.33\\
2.00	0.37\\
2.05	0.40\\
2.10	0.43\\
2.15	0.46\\
2.20	0.49\\
2.25	0.51\\
2.30	0.54\\
2.35	0.56\\
2.40	0.58\\
2.45	0.60\\
2.50	0.62\\
2.55	0.64\\
2.60	0.66\\
2.65	0.68\\
2.70	0.69\\
2.75	0.71\\
2.80	0.72\\
2.85	0.73\\
2.90	0.75\\
2.95	0.76\\
3.00	0.77\\
3.05	0.78\\
3.10	0.79\\
3.15	0.80\\
3.20	0.81\\
3.25	0.82\\
3.30	0.83\\
3.35	0.84\\
3.40	0.84\\
3.45	0.85\\
3.50	0.86\\
3.55	0.87\\
3.60	0.87\\
3.65	0.88\\
3.70	0.88\\
3.75	0.89\\
3.80	0.89\\
3.85	0.90\\
3.90	0.90\\
3.95	0.91\\
4.00	0.91\\
4.05	0.92\\
4.10	0.92\\
4.15	0.93\\
4.20	0.93\\
4.25	0.93\\
4.30	0.94\\
4.35	0.94\\
4.40	0.94\\
4.45	0.94\\
4.50	0.95\\
4.55	0.95\\
4.60	0.95\\
4.65	0.95\\
4.70	0.96\\
4.75	0.96\\
4.80	0.96\\
4.85	0.96\\
4.90	0.96\\
4.95	0.97\\
5.00	0.97\\
5.05	0.97\\
5.10	0.97\\
5.15	0.97\\
5.20	0.97\\
5.25	0.97\\
5.30	0.98\\
5.35	0.98\\
5.40	0.98\\
5.45	0.98\\
5.50	0.98\\
5.55	0.98\\
5.60	0.98\\
5.65	0.98\\
5.70	0.98\\
5.75	0.98\\
5.80	0.99\\
5.85	0.99\\
5.90	0.99\\
5.95	1.00\\
6.00	1.01\\
6.05	1.02\\
6.10	1.04\\
6.15	1.06\\
6.20	1.08\\
6.25	1.11\\
6.30	1.14\\
6.35	1.18\\
6.40	1.22\\
6.45	1.26\\
6.50	1.31\\
6.55	1.36\\
6.60	1.41\\
6.65	1.47\\
6.70	1.53\\
6.75	1.58\\
6.80	1.64\\
6.85	1.70\\
6.90	1.76\\
6.95	1.82\\
7.00	1.88\\
7.05	1.94\\
7.10	1.99\\
7.15	2.04\\
7.20	2.10\\
7.25	2.15\\
7.30	2.20\\
7.35	2.24\\
7.40	2.29\\
7.45	2.33\\
7.50	2.37\\
7.55	2.41\\
7.60	2.45\\
7.65	2.48\\
7.70	2.51\\
7.75	2.54\\
7.80	2.57\\
7.85	2.60\\
7.90	2.63\\
7.95	2.65\\
8.00	2.67\\
8.05	2.70\\
8.10	2.72\\
8.15	2.74\\
8.20	2.75\\
8.25	2.77\\
8.30	2.79\\
8.35	2.80\\
8.40	2.82\\
8.45	2.83\\
8.50	2.84\\
8.55	2.86\\
8.60	2.87\\
8.65	2.88\\
8.70	2.89\\
8.75	2.90\\
8.80	2.91\\
8.85	2.92\\
8.90	2.92\\
8.95	2.93\\
9.00	2.94\\
9.05	2.94\\
9.10	2.95\\
9.15	2.96\\
9.20	2.96\\
9.25	2.97\\
9.30	2.97\\
9.35	2.98\\
9.40	2.98\\
9.45	2.99\\
9.50	2.99\\
9.55	3.00\\
9.60	3.00\\
9.65	3.00\\
9.70	3.01\\
9.75	3.01\\
9.80	3.01\\
9.85	3.02\\
9.90	3.02\\
9.95	3.02\\
10.00	3.03\\
10.05	3.03\\
10.10	3.03\\
10.15	3.03\\
10.20	3.04\\
10.25	3.04\\
10.30	3.04\\
10.35	3.04\\
10.40	3.04\\
10.45	3.05\\
10.50	3.05\\
};
\addlegendentry{$\text{d}_\text{1}$}

\addplot [color=black, dashed, line width=1.5pt]
  table[row sep=crcr]{%
0.05	3.02\\
0.10	3.02\\
0.15	3.02\\
0.20	3.02\\
0.25	3.01\\
0.30	3.01\\
0.35	3.00\\
0.40	2.99\\
0.45	2.97\\
0.50	2.95\\
0.55	2.93\\
0.60	2.91\\
0.65	2.88\\
0.70	2.85\\
0.75	2.82\\
0.80	2.78\\
0.85	2.74\\
0.90	2.70\\
0.95	2.65\\
1.00	2.60\\
1.05	2.55\\
1.10	2.50\\
1.15	2.45\\
1.20	2.39\\
1.25	2.34\\
1.30	2.28\\
1.35	2.23\\
1.40	2.18\\
1.45	2.13\\
1.50	2.08\\
1.55	2.03\\
1.60	1.98\\
1.65	1.93\\
1.70	1.89\\
1.75	1.85\\
1.80	1.81\\
1.85	1.77\\
1.90	1.73\\
1.95	1.70\\
2.00	1.67\\
2.05	1.63\\
2.10	1.61\\
2.15	1.58\\
2.20	1.55\\
2.25	1.53\\
2.30	1.50\\
2.35	1.48\\
2.40	1.46\\
2.45	1.44\\
2.50	1.42\\
2.55	1.40\\
2.60	1.39\\
2.65	1.37\\
2.70	1.36\\
2.75	1.34\\
2.80	1.33\\
2.85	1.32\\
2.90	1.30\\
2.95	1.29\\
3.00	1.28\\
3.05	1.27\\
3.10	1.26\\
3.15	1.25\\
3.20	1.24\\
3.25	1.23\\
3.30	1.22\\
3.35	1.22\\
3.40	1.21\\
3.45	1.20\\
3.50	1.20\\
3.55	1.19\\
3.60	1.18\\
3.65	1.18\\
3.70	1.17\\
3.75	1.17\\
3.80	1.16\\
3.85	1.16\\
3.90	1.15\\
3.95	1.15\\
4.00	1.14\\
4.05	1.14\\
4.10	1.13\\
4.15	1.13\\
4.20	1.13\\
4.25	1.12\\
4.30	1.12\\
4.35	1.12\\
4.40	1.11\\
4.45	1.11\\
4.50	1.11\\
4.55	1.11\\
4.60	1.10\\
4.65	1.10\\
4.70	1.10\\
4.75	1.10\\
4.80	1.10\\
4.85	1.09\\
4.90	1.09\\
4.95	1.09\\
5.00	1.09\\
5.05	1.09\\
5.10	1.09\\
5.15	1.09\\
5.20	1.08\\
5.25	1.08\\
5.30	1.08\\
5.35	1.08\\
5.40	1.08\\
5.45	1.08\\
5.50	1.08\\
5.55	1.08\\
5.60	1.08\\
5.65	1.08\\
5.70	1.07\\
5.75	1.07\\
5.80	1.07\\
5.85	1.06\\
5.90	1.05\\
5.95	1.04\\
6.00	1.03\\
6.05	1.01\\
6.10	1.00\\
6.15	0.97\\
6.20	0.95\\
6.25	0.92\\
6.30	0.89\\
6.35	0.85\\
6.40	0.81\\
6.45	0.77\\
6.50	0.73\\
6.55	0.68\\
6.60	0.64\\
6.65	0.59\\
6.70	0.54\\
6.75	0.49\\
6.80	0.44\\
6.85	0.40\\
6.90	0.36\\
6.95	0.32\\
7.00	0.30\\
7.05	0.28\\
7.10	0.26\\
7.15	0.26\\
7.20	0.27\\
7.25	0.29\\
7.30	0.31\\
7.35	0.34\\
7.40	0.37\\
7.45	0.40\\
7.50	0.43\\
7.55	0.46\\
7.60	0.49\\
7.65	0.52\\
7.70	0.54\\
7.75	0.57\\
7.80	0.60\\
7.85	0.62\\
7.90	0.64\\
7.95	0.66\\
8.00	0.69\\
8.05	0.70\\
8.10	0.72\\
8.15	0.74\\
8.20	0.76\\
8.25	0.77\\
8.30	0.79\\
8.35	0.80\\
8.40	0.82\\
8.45	0.83\\
8.50	0.84\\
8.55	0.85\\
8.60	0.86\\
8.65	0.87\\
8.70	0.88\\
8.75	0.89\\
8.80	0.90\\
8.85	0.91\\
8.90	0.91\\
8.95	0.92\\
9.00	0.93\\
9.05	0.94\\
9.10	0.94\\
9.15	0.95\\
9.20	0.95\\
9.25	0.96\\
9.30	0.96\\
9.35	0.97\\
9.40	0.97\\
9.45	0.98\\
9.50	0.98\\
9.55	0.98\\
9.60	0.99\\
9.65	0.99\\
9.70	0.99\\
9.75	1.00\\
9.80	1.00\\
9.85	1.00\\
9.90	1.01\\
9.95	1.01\\
10.00	1.01\\
10.05	1.01\\
10.10	1.02\\
10.15	1.02\\
10.20	1.02\\
10.25	1.02\\
10.30	1.02\\
10.35	1.03\\
10.40	1.03\\
10.45	1.03\\
10.50	1.03\\
};
\addlegendentry{$\text{d}_\text{2}$}

\end{axis}

\begin{axis}[%
width=1.05\fwidth,
height=0.38\fwidth,
at={(0\fwidth,0.57\fwidth)},
scale only axis,
xmin=0.00,
xmax=10.50,
xlabel style={font=\color{white!15!black}},
xlabel={Time [sec]},
ymin=0.25,
ymax=0.70,
ylabel style={font=\color{white!15!black}},
ylabel={$\unit[r]{[m]}$},
axis background/.style={fill=white},
xmajorgrids,
ymajorgrids,
legend style={at={(0.93,1.23)},legend cell align=left, align=left, legend columns=4, draw=white!15!black}
]
\addplot [color=black, line width=3.0pt]
  table[row sep=crcr]{%
0.05	0.67\\
0.10	0.67\\
0.15	0.67\\
0.20	0.67\\
0.25	0.67\\
0.30	0.66\\
0.35	0.67\\
0.40	0.66\\
0.45	0.66\\
0.50	0.65\\
0.55	0.66\\
0.60	0.64\\
0.65	0.64\\
0.70	0.63\\
0.75	0.63\\
0.80	0.61\\
0.85	0.60\\
0.90	0.58\\
0.95	0.57\\
1.00	0.54\\
1.05	0.53\\
1.10	0.49\\
1.15	0.48\\
1.20	0.43\\
1.25	0.43\\
1.30	0.40\\
1.35	0.40\\
1.40	0.40\\
1.45	0.40\\
1.50	0.40\\
1.55	0.40\\
1.60	0.40\\
1.65	0.40\\
1.70	0.40\\
1.75	0.40\\
1.80	0.40\\
1.85	0.66\\
1.90	0.66\\
1.95	0.66\\
2.00	0.66\\
2.05	0.66\\
2.10	0.66\\
2.15	0.66\\
2.20	0.66\\
2.25	0.66\\
2.30	0.66\\
2.35	0.66\\
2.40	0.66\\
2.45	0.67\\
2.50	0.67\\
2.55	0.67\\
2.60	0.67\\
2.65	0.67\\
2.70	0.67\\
2.75	0.67\\
2.80	0.67\\
2.85	0.67\\
2.90	0.67\\
2.95	0.67\\
3.00	0.67\\
3.05	0.67\\
3.10	0.67\\
3.15	0.67\\
3.20	0.67\\
3.25	0.67\\
3.30	0.67\\
3.35	0.67\\
3.40	0.67\\
3.45	0.67\\
3.50	0.67\\
3.55	0.67\\
3.60	0.67\\
3.65	0.67\\
3.70	0.67\\
3.75	0.67\\
3.80	0.67\\
3.85	0.67\\
3.90	0.67\\
3.95	0.67\\
4.00	0.67\\
4.05	0.67\\
4.10	0.67\\
4.15	0.67\\
4.20	0.67\\
4.25	0.67\\
4.30	0.67\\
4.35	0.67\\
4.40	0.67\\
4.45	0.67\\
4.50	0.67\\
4.55	0.67\\
4.60	0.67\\
4.65	0.67\\
4.70	0.67\\
4.75	0.67\\
4.80	0.67\\
4.85	0.67\\
4.90	0.67\\
4.95	0.67\\
5.00	0.67\\
5.05	0.67\\
5.10	0.67\\
5.15	0.67\\
5.20	0.67\\
5.25	0.67\\
5.30	0.67\\
5.35	0.67\\
5.40	0.67\\
5.45	0.67\\
5.50	0.67\\
5.55	0.67\\
5.60	0.67\\
5.65	0.67\\
5.70	0.67\\
5.75	0.67\\
5.80	0.67\\
5.85	0.67\\
5.90	0.67\\
5.95	0.67\\
6.00	0.66\\
6.05	0.66\\
6.10	0.65\\
6.15	0.64\\
6.20	0.64\\
6.25	0.62\\
6.30	0.64\\
6.35	0.61\\
6.40	0.60\\
6.45	0.57\\
6.50	0.57\\
6.55	0.53\\
6.60	0.52\\
6.65	0.47\\
6.70	0.46\\
6.75	0.42\\
6.80	0.40\\
6.85	0.33\\
6.90	0.34\\
6.95	0.30\\
7.00	0.30\\
7.05	0.30\\
7.10	0.30\\
7.15	0.30\\
7.20	0.30\\
7.25	0.30\\
7.30	0.30\\
7.35	0.30\\
7.40	0.30\\
7.45	0.32\\
7.50	0.65\\
7.55	0.65\\
7.60	0.66\\
7.65	0.66\\
7.70	0.66\\
7.75	0.66\\
7.80	0.66\\
7.85	0.66\\
7.90	0.66\\
7.95	0.66\\
8.00	0.66\\
8.05	0.66\\
8.10	0.66\\
8.15	0.66\\
8.20	0.66\\
8.25	0.66\\
8.30	0.66\\
8.35	0.66\\
8.40	0.66\\
8.45	0.66\\
8.50	0.66\\
8.55	0.66\\
8.60	0.66\\
8.65	0.66\\
8.70	0.66\\
8.75	0.66\\
8.80	0.66\\
8.85	0.66\\
8.90	0.66\\
8.95	0.66\\
9.00	0.66\\
9.05	0.66\\
9.10	0.66\\
9.15	0.66\\
9.20	0.66\\
9.25	0.66\\
9.30	0.66\\
9.35	0.66\\
9.40	0.66\\
9.45	0.66\\
9.50	0.66\\
9.55	0.66\\
9.60	0.66\\
9.65	0.66\\
9.70	0.66\\
9.75	0.66\\
9.80	0.66\\
9.85	0.66\\
9.90	0.66\\
9.95	0.66\\
10.00	0.66\\
10.05	0.66\\
10.10	0.66\\
10.15	0.66\\
10.20	0.66\\
10.25	0.66\\
10.30	0.66\\
10.35	0.66\\
10.40	0.66\\
10.45	0.66\\
10.50	0.66\\
};
\addlegendentry{$r_{\text{front}}$}

\addplot [color=white!80!black!, line width=1.0pt]
  table[row sep=crcr]{%
0.05	0.67\\
0.10	0.67\\
0.15	0.67\\
0.20	0.67\\
0.25	0.67\\
0.30	0.66\\
0.35	0.67\\
0.40	0.66\\
0.45	0.66\\
0.50	0.65\\
0.55	0.66\\
0.60	0.64\\
0.65	0.64\\
0.70	0.63\\
0.75	0.63\\
0.80	0.61\\
0.85	0.60\\
0.90	0.58\\
0.95	0.57\\
1.00	0.54\\
1.05	0.53\\
1.10	0.49\\
1.15	0.48\\
1.20	0.43\\
1.25	0.43\\
1.30	0.40\\
1.35	0.40\\
1.40	0.40\\
1.45	0.40\\
1.50	0.40\\
1.55	0.40\\
1.60	0.40\\
1.65	0.40\\
1.70	0.40\\
1.75	0.40\\
1.80	0.40\\
1.85	0.66\\
1.90	0.66\\
1.95	0.66\\
2.00	0.66\\
2.05	0.66\\
2.10	0.66\\
2.15	0.66\\
2.20	0.66\\
2.25	0.66\\
2.30	0.66\\
2.35	0.66\\
2.40	0.66\\
2.45	0.67\\
2.50	0.67\\
2.55	0.67\\
2.60	0.67\\
2.65	0.67\\
2.70	0.67\\
2.75	0.67\\
2.80	0.67\\
2.85	0.67\\
2.90	0.67\\
2.95	0.67\\
3.00	0.67\\
3.05	0.67\\
3.10	0.67\\
3.15	0.67\\
3.20	0.67\\
3.25	0.67\\
3.30	0.67\\
3.35	0.67\\
3.40	0.67\\
3.45	0.67\\
3.50	0.67\\
3.55	0.67\\
3.60	0.67\\
3.65	0.67\\
3.70	0.67\\
3.75	0.67\\
3.80	0.67\\
3.85	0.67\\
3.90	0.67\\
3.95	0.67\\
4.00	0.67\\
4.05	0.67\\
4.10	0.67\\
4.15	0.67\\
4.20	0.67\\
4.25	0.67\\
4.30	0.67\\
4.35	0.67\\
4.40	0.67\\
4.45	0.67\\
4.50	0.67\\
4.55	0.67\\
4.60	0.67\\
4.65	0.67\\
4.70	0.67\\
4.75	0.67\\
4.80	0.67\\
4.85	0.67\\
4.90	0.67\\
4.95	0.67\\
5.00	0.67\\
5.05	0.67\\
5.10	0.67\\
5.15	0.67\\
5.20	0.67\\
5.25	0.67\\
5.30	0.67\\
5.35	0.67\\
5.40	0.67\\
5.45	0.67\\
5.50	0.67\\
5.55	0.67\\
5.60	0.67\\
5.65	0.67\\
5.70	0.67\\
5.75	0.67\\
5.80	0.67\\
5.85	0.67\\
5.90	0.67\\
5.95	0.67\\
6.00	0.66\\
6.05	0.66\\
6.10	0.65\\
6.15	0.64\\
6.20	0.64\\
6.25	0.62\\
6.30	0.64\\
6.35	0.61\\
6.40	0.60\\
6.45	0.57\\
6.50	0.57\\
6.55	0.53\\
6.60	0.52\\
6.65	0.47\\
6.70	0.46\\
6.75	0.42\\
6.80	0.40\\
6.85	0.33\\
6.90	0.34\\
6.95	0.30\\
7.00	0.30\\
7.05	0.30\\
7.10	0.30\\
7.15	0.30\\
7.20	0.30\\
7.25	0.30\\
7.30	0.30\\
7.35	0.30\\
7.40	0.30\\
7.45	0.32\\
7.50	0.65\\
7.55	0.65\\
7.60	0.66\\
7.65	0.66\\
7.70	0.66\\
7.75	0.66\\
7.80	0.66\\
7.85	0.66\\
7.90	0.66\\
7.95	0.66\\
8.00	0.66\\
8.05	0.66\\
8.10	0.66\\
8.15	0.66\\
8.20	0.66\\
8.25	0.66\\
8.30	0.66\\
8.35	0.66\\
8.40	0.66\\
8.45	0.66\\
8.50	0.66\\
8.55	0.66\\
8.60	0.66\\
8.65	0.66\\
8.70	0.66\\
8.75	0.66\\
8.80	0.66\\
8.85	0.66\\
8.90	0.66\\
8.95	0.66\\
9.00	0.66\\
9.05	0.66\\
9.10	0.66\\
9.15	0.66\\
9.20	0.66\\
9.25	0.66\\
9.30	0.66\\
9.35	0.66\\
9.40	0.66\\
9.45	0.66\\
9.50	0.66\\
9.55	0.66\\
9.60	0.66\\
9.65	0.66\\
9.70	0.66\\
9.75	0.66\\
9.80	0.66\\
9.85	0.66\\
9.90	0.66\\
9.95	0.66\\
10.00	0.66\\
10.05	0.66\\
10.10	0.66\\
10.15	0.66\\
10.20	0.66\\
10.25	0.66\\
10.30	0.66\\
10.35	0.66\\
10.40	0.66\\
10.45	0.66\\
10.50	0.66\\
};
\addlegendentry{$r_{\text{rear}}$}

\addplot [color=black,dotted, line width=1.0pt]
  table[row sep=crcr]{%
0.05	0.40\\
0.10	0.40\\
0.15	0.40\\
0.20	0.40\\
0.25	0.40\\
0.30	0.40\\
0.35	0.40\\
0.40	0.40\\
0.45	0.40\\
0.50	0.40\\
0.55	0.40\\
0.60	0.40\\
0.65	0.40\\
0.70	0.40\\
0.75	0.40\\
0.80	0.40\\
0.85	0.40\\
0.90	0.40\\
0.95	0.40\\
1.00	0.40\\
1.05	0.40\\
1.10	0.40\\
1.15	0.40\\
1.20	0.40\\
1.25	0.40\\
1.30	0.40\\
1.35	0.40\\
1.40	0.40\\
1.45	0.40\\
1.50	0.40\\
1.55	0.40\\
1.60	0.40\\
1.65	0.40\\
1.70	0.40\\
1.75	0.40\\
1.80	0.40\\
1.85	0.40\\
1.90	0.40\\
1.95	0.40\\
2.00	0.40\\
2.05	0.40\\
2.10	0.40\\
2.15	0.40\\
2.20	0.40\\
2.25	0.40\\
2.30	0.40\\
2.35	0.40\\
2.40	0.40\\
2.45	0.40\\
2.50	0.40\\
2.55	0.40\\
2.60	0.40\\
2.65	0.40\\
2.70	0.40\\
2.75	0.40\\
2.80	0.40\\
2.85	0.40\\
2.90	0.40\\
2.95	0.40\\
3.00	0.40\\
3.05	0.40\\
3.10	0.40\\
3.15	0.40\\
3.20	0.40\\
3.25	0.40\\
3.30	0.40\\
3.35	0.40\\
3.40	0.40\\
3.45	0.40\\
3.50	0.40\\
3.55	0.40\\
3.60	0.40\\
3.65	0.40\\
3.70	0.40\\
3.75	0.40\\
3.80	0.40\\
3.85	0.40\\
3.90	0.40\\
3.95	0.40\\
4.00	0.40\\
4.05	0.40\\
4.10	0.40\\
4.15	0.40\\
4.20	0.40\\
4.25	0.40\\
4.30	0.40\\
4.35	0.40\\
4.40	0.40\\
4.45	0.40\\
4.50	0.40\\
4.55	0.40\\
4.60	0.40\\
4.65	0.40\\
4.70	0.40\\
4.75	0.40\\
4.80	0.40\\
4.85	0.40\\
4.90	0.40\\
4.95	0.40\\
5.00	0.40\\
5.05	0.40\\
5.10	0.40\\
5.15	0.40\\
5.20	0.40\\
5.25	0.40\\
5.30	0.40\\
5.35	0.40\\
5.40	0.40\\
5.45	0.40\\
5.50	0.40\\
5.55	0.40\\
5.60	0.40\\
5.65	0.40\\
5.70	0.40\\
5.75	0.40\\
5.80	0.40\\
5.85	0.40\\
5.90	0.40\\
5.95	0.40\\
6.00	0.40\\
6.05	0.40\\
6.10	0.40\\
6.15	0.40\\
6.20	0.40\\
6.25	0.40\\
6.30	0.40\\
6.35	0.40\\
6.40	0.40\\
6.45	0.40\\
6.50	0.40\\
6.55	0.40\\
6.60	0.40\\
6.65	0.40\\
6.70	0.40\\
6.75	0.40\\
6.80	0.40\\
6.85	0.40\\
6.90	0.40\\
6.95	0.40\\
7.00	0.40\\
7.05	0.40\\
7.10	0.40\\
7.15	0.40\\
7.20	0.40\\
7.25	0.40\\
7.30	0.40\\
7.35	0.40\\
7.40	0.40\\
7.45	0.40\\
7.50	0.40\\
7.55	0.40\\
7.60	0.40\\
7.65	0.40\\
7.70	0.40\\
7.75	0.40\\
7.80	0.40\\
7.85	0.40\\
7.90	0.40\\
7.95	0.40\\
8.00	0.40\\
8.05	0.40\\
8.10	0.40\\
8.15	0.40\\
8.20	0.40\\
8.25	0.40\\
8.30	0.40\\
8.35	0.40\\
8.40	0.40\\
8.45	0.40\\
8.50	0.40\\
8.55	0.40\\
8.60	0.40\\
8.65	0.40\\
8.70	0.40\\
8.75	0.40\\
8.80	0.40\\
8.85	0.40\\
8.90	0.40\\
8.95	0.40\\
9.00	0.40\\
9.05	0.40\\
9.10	0.40\\
9.15	0.40\\
9.20	0.40\\
9.25	0.40\\
9.30	0.40\\
9.35	0.40\\
9.40	0.40\\
9.45	0.40\\
9.50	0.40\\
9.55	0.40\\
9.60	0.40\\
9.65	0.40\\
9.70	0.40\\
9.75	0.40\\
9.80	0.40\\
9.85	0.40\\
9.90	0.40\\
9.95	0.40\\
10.00	0.40\\
10.05	0.40\\
10.10	0.40\\
10.15	0.40\\
10.20	0.40\\
10.25	0.40\\
10.30	0.40\\
10.35	0.40\\
10.40	0.40\\
10.45	0.40\\
10.50	0.40\\
};
\addlegendentry{$w_{obs1}$}

\addplot [color=black, dashed, line width=1.0pt]
  table[row sep=crcr]{%
0.05	0.30\\
0.10	0.30\\
0.15	0.30\\
0.20	0.30\\
0.25	0.30\\
0.30	0.30\\
0.35	0.30\\
0.40	0.30\\
0.45	0.30\\
0.50	0.30\\
0.55	0.30\\
0.60	0.30\\
0.65	0.30\\
0.70	0.30\\
0.75	0.30\\
0.80	0.30\\
0.85	0.30\\
0.90	0.30\\
0.95	0.30\\
1.00	0.30\\
1.05	0.30\\
1.10	0.30\\
1.15	0.30\\
1.20	0.30\\
1.25	0.30\\
1.30	0.30\\
1.35	0.30\\
1.40	0.30\\
1.45	0.30\\
1.50	0.30\\
1.55	0.30\\
1.60	0.30\\
1.65	0.30\\
1.70	0.30\\
1.75	0.30\\
1.80	0.30\\
1.85	0.30\\
1.90	0.30\\
1.95	0.30\\
2.00	0.30\\
2.05	0.30\\
2.10	0.30\\
2.15	0.30\\
2.20	0.30\\
2.25	0.30\\
2.30	0.30\\
2.35	0.30\\
2.40	0.30\\
2.45	0.30\\
2.50	0.30\\
2.55	0.30\\
2.60	0.30\\
2.65	0.30\\
2.70	0.30\\
2.75	0.30\\
2.80	0.30\\
2.85	0.30\\
2.90	0.30\\
2.95	0.30\\
3.00	0.30\\
3.05	0.30\\
3.10	0.30\\
3.15	0.30\\
3.20	0.30\\
3.25	0.30\\
3.30	0.30\\
3.35	0.30\\
3.40	0.30\\
3.45	0.30\\
3.50	0.30\\
3.55	0.30\\
3.60	0.30\\
3.65	0.30\\
3.70	0.30\\
3.75	0.30\\
3.80	0.30\\
3.85	0.30\\
3.90	0.30\\
3.95	0.30\\
4.00	0.30\\
4.05	0.30\\
4.10	0.30\\
4.15	0.30\\
4.20	0.30\\
4.25	0.30\\
4.30	0.30\\
4.35	0.30\\
4.40	0.30\\
4.45	0.30\\
4.50	0.30\\
4.55	0.30\\
4.60	0.30\\
4.65	0.30\\
4.70	0.30\\
4.75	0.30\\
4.80	0.30\\
4.85	0.30\\
4.90	0.30\\
4.95	0.30\\
5.00	0.30\\
5.05	0.30\\
5.10	0.30\\
5.15	0.30\\
5.20	0.30\\
5.25	0.30\\
5.30	0.30\\
5.35	0.30\\
5.40	0.30\\
5.45	0.30\\
5.50	0.30\\
5.55	0.30\\
5.60	0.30\\
5.65	0.30\\
5.70	0.30\\
5.75	0.30\\
5.80	0.30\\
5.85	0.30\\
5.90	0.30\\
5.95	0.30\\
6.00	0.30\\
6.05	0.30\\
6.10	0.30\\
6.15	0.30\\
6.20	0.30\\
6.25	0.30\\
6.30	0.30\\
6.35	0.30\\
6.40	0.30\\
6.45	0.30\\
6.50	0.30\\
6.55	0.30\\
6.60	0.30\\
6.65	0.30\\
6.70	0.30\\
6.75	0.30\\
6.80	0.30\\
6.85	0.30\\
6.90	0.30\\
6.95	0.30\\
7.00	0.30\\
7.05	0.30\\
7.10	0.30\\
7.15	0.30\\
7.20	0.30\\
7.25	0.30\\
7.30	0.30\\
7.35	0.30\\
7.40	0.30\\
7.45	0.30\\
7.50	0.30\\
7.55	0.30\\
7.60	0.30\\
7.65	0.30\\
7.70	0.30\\
7.75	0.30\\
7.80	0.30\\
7.85	0.30\\
7.90	0.30\\
7.95	0.30\\
8.00	0.30\\
8.05	0.30\\
8.10	0.30\\
8.15	0.30\\
8.20	0.30\\
8.25	0.30\\
8.30	0.30\\
8.35	0.30\\
8.40	0.30\\
8.45	0.30\\
8.50	0.30\\
8.55	0.30\\
8.60	0.30\\
8.65	0.30\\
8.70	0.30\\
8.75	0.30\\
8.80	0.30\\
8.85	0.30\\
8.90	0.30\\
8.95	0.30\\
9.00	0.30\\
9.05	0.30\\
9.10	0.30\\
9.15	0.30\\
9.20	0.30\\
9.25	0.30\\
9.30	0.30\\
9.35	0.30\\
9.40	0.30\\
9.45	0.30\\
9.50	0.30\\
9.55	0.30\\
9.60	0.30\\
9.65	0.30\\
9.70	0.30\\
9.75	0.30\\
9.80	0.30\\
9.85	0.30\\
9.90	0.30\\
9.95	0.30\\
10.00	0.30\\
10.05	0.30\\
10.10	0.30\\
10.15	0.30\\
10.20	0.30\\
10.25	0.30\\
10.30	0.30\\
10.35	0.30\\
10.40	0.30\\
10.45	0.30\\
10.50	0.30\\
};
\addlegendentry{$w_{obs2}$}
\end{axis}
\end{tikzpicture}%